\newif\ifTechRep
\TechReptrue
\documentclass[sigconf]{acmart}
\usepackage{graphicx}
\usepackage{balance} 
\usepackage{enumitem}
\usepackage[hang,flushmargin]{footmisc}

\usepackage{amsthm, bm}
\usepackage[ruled,vlined]{algorithm2e}
\usepackage[labelfont=bf,font=small]{caption}
\usepackage[labelformat=simple]{subcaption}
\usepackage{xspace}
\usepackage{booktabs}
\usepackage{url}
\usepackage{multicol}
\usepackage{multirow}
\usepackage[normalem]{ulem}
\usepackage{color}
\usepackage{placeins}
\usepackage{pifont}
\usepackage{makecell}
\usepackage{rotating}
\usepackage{tablefootnote}
\usepackage{colortbl}
\usepackage{hyperref}
\usepackage{stmaryrd}
\usepackage{framed} 
\usepackage[noabbrev]{cleveref}
\usepackage{comment}
\usepackage{xcolor}

\AtBeginDocument{%
  \providecommand\BibTeX{{%
    \normalfont B\kern-0.5em{\scshape i\kern-0.25em b}\kern-0.8em\TeX}}}

\copyrightyear{2022}
\acmYear{2022}
\setcopyright{acmlicensed}\acmConference[SIGMOD '22]{Proceedings of the
2022 International Conference on Management of Data}{June 12--17,
2022}{Philadelphia, PA, USA}
\acmBooktitle{Proceedings of the 2022 International Conference on
Management of Data (SIGMOD '22), June 12--17, 2022, Philadelphia, PA, USA}
\acmPrice{15.00}
\acmDOI{10.1145/3514221.3517841}
\acmISBN{978-1-4503-9249-5/22/06}

\hypersetup{
  bookmarksdepth = {5},
  pdftitle = {},
  pdfkeywords = {},
  pdfauthor = {},
  pdfstartview={FitH},
  urlcolor=black,
  linkcolor=black,
  citecolor=black,
  colorlinks=true,
}

\setlist{nolistsep,leftmargin=*}
\newtheorem{example}{Example}

\DeclareMathOperator*{\argmax}{argmax}
\let\paragraph\relax 
\newcommand{\paragraph}[1]{\noindent\textbf{#1}}

\renewcommand{\footnotesize}{\scriptsize}
\newcommand{\ra}[1]{\renewcommand{\arraystretch}{#1}}
\newcommand{\cmark}{\multicolumn{1}{c}{\ding{51}}}

\newcommand{\turnr}{\rotatebox{90}}
\newcommand{\cc}{\cellcolor{white}}
 
\definecolor{darkgreen}{rgb}{0.0, 0.5, 0.0}

\usepackage{dcolumn}
\newcolumntype{d}[1]{D{.}{.}{#1}}  
\newcolumntype{C}{>{\footnotesize}l}

\newcommand{\citeTechRep}{\cite{technicalReport}}

\newcommand{\appOrTechRep}{\ifTechRep the Appendix\xspace\else our technical report~\citeTechRep\xspace\fi}

\newcommand{\orig}{\textsc{LR}\xspace}

\newcommand{\kam}{$\text{\textsc{Kam-Cal}}$\xspace}
\newcommand{\kamt}{$\text{\textsc{Kam-Cal\textsuperscript{dp}}}$\xspace}

\newcommand{\feld}{\textsc{Feld}\xspace}
\newcommand{\feldt}{$\text{\textsc{Feld}}^{\text{\textsc{dp}}}$\xspace}

\newcommand{\calm}{\textsc{Calmon}\xspace}
\newcommand{\calmt}{$\text{\textsc{Calmon\textsuperscript{dp}}}$\xspace}

\newcommand{\zhangpcf}{$\text{\textsc{Zha-Wu}}$\xspace}
\newcommand{\zhangwu}{$\text{\textsc{Zha-Wu}}$\xspace}
\newcommand{\zhangpcft}{$\text{\textsc{Zha-Wu}}^{\text{\textsc{psf}}}$\xspace}
\newcommand{\zhangdcet}{$\text{\textsc{Zha-Wu}}^{\text{\textsc{dce}}}$\xspace}

\newcommand{\sal}{$\text{\textsc{Salimi}}$\xspace}
\newcommand{\salt}{$\text{\textsc{Salimi}}^{\text{\textsc{jf}}}$\xspace}
\newcommand{\salmst}{$\text{\textsc{Salimi}}^{\text{\textsc{jf}}}_{\text{\textsc{MaxSAT}}}$\xspace}
\newcommand{\salmft}{$\text{\textsc{Salimi}}^{\text{\textsc{jf}}}_{\text{\textsc{MatFac}}}$\xspace}

\newcommand{\zaf}{\textsc{Zafar}\xspace}
\newcommand{\zdi}{$\text{\textsc{Zafar}}^{\text{\textsc{dp}}}$\xspace}
\newcommand{\zdmt}{$\text{\textsc{Zafar}}^{\text{\textsc{eo}}}_{\text{\textsc{Fair}}}$\xspace}
\newcommand{\zdifairt}{$\text{\textsc{Zafar}}^{\text{\textsc{dp}}}_{\text{\textsc{Fair}}}$\xspace}
\newcommand{\zdiacct}{$\text{\textsc{Zafar}}^{\text{\textsc{dp}}}_{\text{\textsc{Acc}}}$\xspace}

\newcommand{\zhang}{$\text{\textsc{Zha-Le}}$\xspace}
\newcommand{\zhangt}{$\text{\textsc{Zha-Le}}^{\text{\textsc{eo}}}$\xspace}

\newcommand{\kearns}{\textsc{Kearns}\xspace}
\newcommand{\kearnst}{$\text{\textsc{Kearns\textsuperscript{pe}}}$\xspace}

\newcommand{\celis}{\textsc{Celis}\xspace}
\newcommand{\celist}{\textsc{Celis\textsuperscript{pp}}\xspace}

\newcommand{\thomas}{\textsc{Thomas}\xspace}
\newcommand{\thomasdpt}{\textsc{Thomas\textsuperscript{dp}}\xspace}
\newcommand{\thomaseot}{\textsc{Thomas\textsuperscript{eo}}\xspace}

\newcommand{\kamdp}{\textsc{Kam-Kar}\xspace}
\newcommand{\kamdpt}{\textsc{Kam-Kar\textsuperscript{dp}}\xspace}

\newcommand{\hardt}{\textsc{Hardt}\xspace}
\newcommand{\hardtt}{$\text{\textsc{Hardt\textsuperscript{eo}}}$\xspace}

\newcommand{\pleiss}{\textsc{Pleiss}\xspace}
\newcommand{\pleisst}{$\text{\textsc{Pleiss\textsuperscript{eop}}}$\xspace}

\newcommand{\madrast}{$\text{\textsc{Madras\textsuperscript{dp}}}$\xspace}

\newcommand{\agarwaldpt}{$\text{\textsc{Agarwal\textsuperscript{dp}}}$\xspace}
\newcommand{\agarwaleot}{$\text{\textsc{Agarwal\textsuperscript{eo}}}$\xspace}

\newcommand{\prot}{0\xspace}
\newcommand{\priv}{1\xspace}
\newcommand{\discone}{\textsc{Discri\-mi\-nation-1}\xspace}
\newcommand{\disctwo}{\textsc{Discri\-mi\-nation-2}\xspace}
\newcommand{\countClassifier}{13\xspace}

\newcommand{\di}{$\mathit{DI}$\xspace}
\newcommand{\tprb}{$\mathit{TPRB}$\xspace}
\newcommand{\tnrb}{$\mathit{TNRB}$\xspace}

\newcommand{\id}{$\mathit{ID}$\xspace}
\newcommand{\te}{$\mathit{TE}$\xspace}
\newcommand{\nde}{$\mathit{NDE}$\xspace}
\newcommand{\nie}{$\mathit{NIE}$\xspace}

\newcounter{commentCounter}

\begin{document}


\title[Through the Data Management Lens: Experimental Analysis and Evaluation of Fair Classification]{Through the Data Management Lens:\\ Experimental Analysis and Evaluation of Fair Classification}


\author{Maliha Tashfia Islam}
\affiliation{%
  \institution{University of Massachusetts Amherst}
  \country{}}
\email{mtislam@cs.umass.edu}

\author{Anna Fariha}
\affiliation{%
  \institution{Microsoft}
  \country{}}
\email{annafariha@microsoft.com}

\author{Alexandra Meliou}
\affiliation{%
  \institution{University of Massachusetts Amherst}
  \country{}}
\email{ameli@cs.umass.edu}

\author{Babak Salimi}
\affiliation{%
  \institution{University of California, San Diego}
  \country{}}
\email{bsalimi@ucsd.edu}


\begin{abstract}
%

\looseness-1 Classification, a heavily studied data-driven machine learning
task, drives a large number of prediction systems involving critical decisions
such as loan approval and criminal risk assessment. However, classifiers often
demonstrate discriminatory behavior, especially when presented with biased data.
Consequently, fairness in classification has emerged as a high-priority research
area. Data management research is showing an increasing presence and interest in
topics related to data and algorithmic fairness, including the topic of fair
classification. The interdisciplinary efforts in fair classification, with
machine learning research having the largest presence, have resulted in a large
number of fairness notions and a wide range of approaches that have not been
systematically evaluated and compared. In this paper, we contribute a broad
analysis of \countClassifier fair classification approaches and additional
variants, over their correctness, fairness, efficiency, scalability, robustness
to data errors, sensitivity to underlying ML model, data efficiency, and
stability using a variety of metrics and real-world datasets. Our analysis
highlights novel insights on the impact of different metrics and high-level
approach characteristics on different aspects of performance. We also discuss
general principles for choosing approaches suitable for different practical
settings, and identify areas where data-management-centric solutions are likely
to have the most impact.
\end{abstract}




\maketitle

\section{Introduction}\label{sec:introduction}
%
Virtually every aspect of human activity relies on automated systems that use
prediction models learned from data: from routine everyday tasks, such as search
results and product recommendations~\cite{grbovic2015}, all the way to
high-stakes decisions such as mortgage approval~\cite{mortgage}, job applicant
filtering~\cite{jobapp}, and pre-trial risk assessment of criminal
defendants~\cite{compas}. However, automated predictions are only as good
as the data that drives them. As inherent biases are
common in data~\cite{barocas2016}, data-driven systems commonly demonstrate
unfair and discriminatory behavior~\cite{compas, berk, Salimi, valentino}.

\looseness-1
Data management research has shown growing interest in the topic of fairness
over applications related to ranking, data synthesis, result diversification,
and others~\cite{kuhlman2020, stoyanovich2018, asudeh2019, yan2019, asudeh2020,
DBLP:journals/corr/abs-2006-06053, DBLP:journals/corr/abs-2002-03508}. However,
much of this work does not target prediction systems directly. In fact, a
relatively small portion of the fairness literature within the data management
community has directly targeted \emph{classification}~\cite{feldman, lahoti2019,
Salimi, zhang2021, salazar2021}, one of the most important and heavily studied
supervised ML tasks that drives many broadly used prediction systems. In
contrast, machine learning research has rapidly produced a large body of work on
the problem of improving fairness in classification.

In this paper, we closely study and empirically evaluate existing work on fair
classification, across different research communities, with two primary
objectives: (1)~to highlight data management aspects of existing work, such as
scalability, robustness to data errors, stability wrt to partitions of training
data, and data efficiency, which are important practical considerations often
overlooked in other communities, and (2)~to produce a deeper understanding of
tradeoffs that may exist across various approaches, creating guidelines for
where data management solutions are more likely to have impact. We proceed to
provide a brief background on the problem of fair classification and existing
approaches, we state the scope of our work and contrast with prior evaluation
and analysis research, and, finally, we list our contributions.

\smallskip 

\paragraph{Background on fair classification.}\looseness-1
Classifiers typically focus on maximizing \emph{correctness}, i.e., how
well predictions match the ground truth. To that end, a trained classifier
naturally prioritizes the minimization of prediction error over the majority
groups within the data, and, thus, performs better for entities belonging to
those groups. However, this may result in poor prediction performance over
minority groups. Moreover, as all data-driven approaches, classifiers also
suffer from the general phenomenon of ``garbage-in, garbage-out'': if the data
contains inherent biases, the model will reflect or even exacerbate them. Thus,
traditional learning may discriminate in two ways: (1)~models make more
incorrect predictions over the minority than the majority groups, and
(2)~they replicate training data biases. We highlight this with a
real-world example.

\begin{example} \label{ex:discrimination} Consider COMPAS, a risk-assessment
    system that can predict recidivism (the tendency to reoffense) in convicted
    criminals. It is used by the U.S. courts to classify defendants as high- or
    low-risk according to their likelihood of recidivating within 2 years of
    initial assessment~\cite{flores2016}, and achieves nearly $70\%$
    accuracy~\cite{risktool}. In 2014, a detailed analysis of COMPAS revealed
    some very troubling findings: black defendants are twice more likely than
    white defendants to be \emph{incorrectly} predicted as high-risk, while
    white reoffenders are \emph{incorrectly} predicted as low-risk almost twice
    as often as black reoffenders~\cite{compas}. While COMPAS' overall accuracy
    was similar over both groups ($67\%$ for black and $69\%$ for white), its
    mistakes affected the two groups disproportionately. COMPAS was further
    criticized for exacerbating societal bias due to utilizing historical arrest
    data in the training set, despite certain populations being proven to be
    more policed than others~\cite{arrestdata}. 
\end{example}

\vspace{-1.5mm}
Example~\ref{ex:discrimination} is not an isolated incident;  other
cases of classifier discrimination have pointed towards racial~\cite{berk},
gender~\cite{Salimi}, and other forms of bias and
unfairness~\cite{valentino}. The pervasiveness of discriminatory
behavior in prediction systems indicates that \emph{fairness} should be an
important objective in classification. In recent years, study of fair
classification has garnered significant interest across multiple
disciplines~\cite{feldman, Salimi, zafar, celis, hardt}, and a multitude of
approaches and notions of fairness have emerged~\cite{fairdef, narayanan}. We
consider two principal dimensions in characterizing the work in this domain:
(1)~the targeted notion of fairness, and (2)~the stage---before, during, or
after training---when fairness-enforcing mechanisms are applied.

\looseness-1 \emph{Fairness notions and mechanisms.} Fairness is subjective and
specifying what is fair is non-trivial: definitions of fairness are often driven
by application-specific and even legal considerations. Existing literature has
proposed a large number of notions to capture different fairness
objectives~\cite{fairdef, narayanan}, and new ones continue to emerge. A
principled comparison of these notions is non-trivial, due to the high diversity
in their mechanisms. Some fairness notions measure discrimination through
\emph{causal} association among attributes of interest (e.g., race and
prediction), while others study non-causal associations. Further, some notions
capture if \emph{individuals} are treated fairly, while others quantify fair
treatment of a \emph{group} (e.g., people of certain race or gender). The demand
for domain knowledge also varies: some rely on \emph{observational} data, while
others require \emph{interventions} or \emph{counterfactuals}. To add further
complexity, multiple recent studies~\cite{csp, kleinberg, majumder2021} prove
that most fairness notions tend to be incompatible with each other and cannot be
enforced simultaneously. \label{pg:r1o2} \label{pg:r3o3_1}

\emph{Fairness-enforcing stage.}
Existing methods in fair classification operate in one of the three possible
stages. \emph{Pre-processing} approaches attempt to repair biases in the data
\emph{before} the data is used to train a classifier~\cite{kamiran, feldman,
calmon, zhang2017, zhang2017kdd, Salimi}. Data management research in fair
classification has typically focused on the pre-processing stage. In contrast,
the machine learning community largely explored \emph{in-processing} approaches,
which alter the learning procedure used by the classifier~\cite{zafar, zafarDM,
zhang, kearns, celis, thomas2019}, and \emph{post-processing} approaches, which
alter the classifier predictions to ensure fairness ~\cite{kamiran2012, hardt,
pleiss}. Similar to fairness notions, the wide variety of mechanisms applied by
fair approaches present a significant challenge in understanding them. Further,
there is a clear lack of literature that empirically evaluate these approaches,
making it difficult to compare the tradeoffs that approaches may make while
enforcing fairness.

\smallskip

\paragraph{Scope of our work.}\label{pg:r1o5} We present a systematic and
thorough empirical evaluation of \countClassifier fair classification approaches
and some of their variants, resulting in~18 different approaches, along axes
that the data management community cares about: \emph{correctness},
\emph{fairness}, \emph{scalability}, \emph{robustness to data errors},
\emph{sensitivity to ML model}, \emph{data efficiency}, and \emph{stability}. We
selected approaches that target a representative variety of fairness definitions
and span all three (pre, in, and post) fairness-enforcing stages. In general,
there is no one-size-fits-all solution when it comes to choosing the best fair
approach and the choice is application-specific. However, our evaluation has two
main objectives: (1)~to highlight practical concerns such as scalability,
robustness to data errors, etc., that are relevant to many real-world
applications but have been overlooked in fairness literature, and (2)~to produce
a deeper understanding of tradeoffs and challenges across various approaches,
creating guidelines for where data management solutions are more likely to have
impact. For example, our findings suggest that pre-processing approaches, while
a natural fit for data-focused solutions, tend to face scalability issues with
high-dimensional data. The contributions of our work lie both in the breadth of
our evaluation, as well as in the unique perspective of data-management
considerations, which have not been previously explored in this context. To the
best of our knowledge, this is the first study and evaluation of fair
classification approaches through a data management lens.

\emph{Other evaluation and analysis work on fair classification.}
Our focus on the empirical evaluation of methods in fair classification
distinguishes our work from existing surveys that review the broad area but do
not include experimental results and analysis~\cite{fairdef, makhlouf2020,
mehrabi2019, caton2020}. Moreover, prior work on the evaluation of fair
classifiers had a narrower scope than ours. Friedler et al.~\cite{friedler}
carry out experimental analysis similar to ours by evaluating variations of 4
fair approaches over 5 fairness metrics, while Jones et al.~\cite{jones2020}
evaluate variations of 6 fair approaches over 3 fairness metrics. However, they
overlook performance aspects (e.g., runtime, scalability, data-efficiency) and
robustness to data-quality issues (e.g., errors), which are critical in
practice. Further, their analysis excludes post-processing approaches and
individual fairness metrics. AI Fairness 360~\cite{aif} is an extensible toolkit
that offers mechanisms to empirically evaluate fair approaches over different
fairness metrics. However, it does not offer any insight highlighting the
tradeoffs among fair approaches, and cannot compare other aspects such as
efficiency, scalability, robustness to data errors, stability, etc. Lastly, a
few general frameworks~\cite{fairtest, themis} evaluate fair approaches on a
specific fairness metric, but are not designed to offer insights based on
comparative analysis.

\vspace{3mm}
\noindent\textbf{Contributions.} 
In this paper, we make the following contributions:

\begin{itemize}
	
    \item We provide a new and informative categorization of 34 existing
    fairness notions, based on the high-level aspects of association,
    granularity, causal hierarchy, and requirements. We discuss their
    implications, tradeoffs, and limitations, and justify the choices of metrics
    for our evaluation. (Section~\ref{sec:background})
	
    \item We provide an overview of \countClassifier fair classification
    approaches and several variants. We select 5
    \emph{pre-processing}~\cite{kamiran, feldman, calmon, zhang2017,
    zhang2017kdd, Salimi}, 5 \emph{in-processing}~\cite{zafar, zafarDM, zhang,
    kearns, celis, thomas2019}, and 3 \emph{post-processing}
    approaches~\cite{kamiran2012, hardt, pleiss} for our evaluation.
    (Section~\ref{sec:algorithm})

    \item We evaluate a total of 18 variants of fair classification techniques
    with respect to 4 correctness and 5 fairness metrics over 3 real-world
    datasets including Adult~\cite{adult} and COMPAS~\cite{compas}. Our
    evaluation provides interesting insights regarding the trends in
    fairness-correctness tradeoffs. (Section~\ref{subsec:correct-fair})
    
    \item Our runtime evaluation indicates that post-processing approaches are
    generally most efficient and scalable. However, their efficiency and
    scalability are due to the simplicity of their mechanism, which limits their
    capacity of balancing correctness-fairness tradeoffs. In contrast, pre- and
    in-processing approaches generally incur higher runtimes, but offer more
    flexibility in controlling correct\-ness-fairness tradeoffs.
    (Section~\ref{subsec:runtime})

    \item We investigate the robustness of all approaches to quality issues
    (e.g., errors) in training data, shedding light on their feasibility in
    practical settings. Our results indicate that pre- and in-processing exhibit
    poor generalizability and often fail to achieve their target fairness, while
    post-processing is more robust. (Section~\ref{sec:robustness})

    \item To evaluate the sensitivity of pre- and post-processing approaches to
    the choice of ML model, we pair each approach with 5 different ML models and
    compare their correctness-fairness balance. Our findings show that
    pre-processing approaches can produce noticeably varied results on different
    models, while post-processing is not sensitive to the choice of ML model.
    (Section~\ref{sec:resilience})

    \item We summarize further results on the data efficiency (dependence on
    training set size) and stability (variance over different partitions of the
    training data) of all approaches. Our results suggest that most approaches
    are data-efficient and stable, and there is no significant trend.
    (Section~\ref{sec:others})

    \item Finally, based on the insights from our evaluation, we discuss
    general guidelines towards selecting suitable fair classification
    approaches in different settings, and highlight possible areas where
    data management solutions can be most impactful.
    (Section~\ref{sec:discussion}) 
    
\end{itemize}

\section{Evaluation Metrics}\label{sec:background}
%
\def\X{\mathbb{X}} \def\Dom{\mathbf{Dom}} \def\D{\mathcal{D}}

In this section, we introduce the metrics that we use to measure the correctness
and fairness of the evaluated techniques. We start with some basic notations
related to the concepts of binary classification and then proceed to describe
the two types of evaluation metrics and the rationale behind our choices.

\subsubsection*{Basic notations.} Let $\D$ be an annotated dataset with the
schema $(\X, S; Y)$, where $\X$ denotes a set of attributes that describe each
tuple or individual in the dataset $\D$, $S$ denotes a sensitive attribute, and
$Y$ denotes the annotation (ground-truth class label). Without loss of
generality, we assume that $S$ is binary, i.e., $\Dom(S) = \{\prot, \priv\}$,
where \priv indicates a \emph{privileged} and \prot indicates an
\emph{unprivileged} group. We use $S_t$ to denote the particular sensitive
attribute assignment of a tuple $t \in \D$.
We denote a binary classification task $f: f(\X) \rightarrow \hat{Y}$, where
$\hat{Y}$ denotes the \emph{predicted} class label ($\Dom(Y) = \Dom(\hat Y) =
\{0, 1\}$). Without loss of generality, we interpret $1$ as a favorable
(positive) prediction and $0$ as an unfavorable (negative) prediction. We use
$Y_t$ and $\hat{Y}_t$ to denote the ground-truth and predicted class label for
$t$, respectively. We summarize the notations in Figure~\ref{table:notation}.

\begin{figure}[t]
    \setlength{\tabcolsep}{2.5pt}
    \renewcommand{\arraystretch}{0.8}
	\centering
	\resizebox{0.4\textwidth}{!}{\small
    \begin{tabular}{ll}
	\toprule
	\textbf{Notation}             & \textbf{Description} \\ 
	\midrule 
    $\X$                          &  A set of attributes \\ 
	$X$,  $\Dom(X)$				  &  A single attribute $X$ and its value domain\\
    $S$                           &  A sensitive attribute\\ 
	$Y$, $\hat{Y}$                &  Attribute denoting ground-truth and predicted class label\\  
	$\D$                		  &  An annotated dataset with the schema $(\X, S; Y)$\\ 
    $f(\X)\rightarrow\hat{Y}$ 	  &  A binary classifier\\ 
    $S_t$ 			 			  &  Value of the sensitive attribute $S$ for tuple $t\in \D$\\ 
	$Y_t$, $\hat{Y}_t$			  &  Ground-truth and predicted class labels for tuple $t\in \D$\\
    \bottomrule
	\end{tabular}}
	\vspace{-3mm}
    \caption{Summary of notations.}
    \vspace{2mm}
    \label{table:notation}

	\setlength{\tabcolsep}{1.5pt}
    \renewcommand{\arraystretch}{-2}
	\centering
	\resizebox{0.4\textwidth}{!}{\small
		\begin{tabular}{lcl<{\hspace{0.7mm}}l}
			
			\toprule	
			\bf{Metric}	& \bf{Definition} & \bf{Range} & \bf{Interpretation}\\
			\midrule
			Accuracy & $\frac{\mathit{TP} + \mathit{TN}}{\mathit{TP} + \mathit{TN} + \mathit{FP} + \mathit{FN}}$ 		& [0, 1] & \makecell[l]{Accuracy = 1 $\rightarrow$ completely correct\\Accuracy = 0 $\rightarrow$ completely incorrect} \\\hline
			\rule{0pt}{12pt}Precision & $\frac{\mathit{TP}}{\mathit{TP} + \mathit{FP}}$ 	& [0, 1] & \makecell[l]{Precision = 1 $\rightarrow$ completely correct\\Precision = 0 $\rightarrow$ completely incorrect} \\\hline
			\rule{0pt}{12pt}Recall & $\frac{\mathit{TP}}{\mathit{TP} + \mathit{FN}}$ 		& [0, 1] & \makecell[l]{Recall = 1 $\rightarrow$ completely correct\\Recall = 0 $\rightarrow$ completely incorrect}\\\hline
			\rule{0pt}{12pt}F$_1$-score & $\frac{2 \cdot \mathit{Precision} \cdot 
			\mathit{Recall}}{\mathit{Precision} + \mathit{Recall}}$						    & [0, 1] & \makecell[l]{F$_1$-score = 1 $\rightarrow$ completely correct\\F$_1$-score = 0 $\rightarrow$ completely incorrect}\\
			\bottomrule
		\end{tabular}
	}
	\vspace{-3mm}
	\caption{List of correctness metrics used in our evaluation.}
	\vspace{-4mm}
	\label{fig:correctness}
\end{figure} 

\subsection{Correctness} 
\looseness-1 Correctness of a binary classifier measures how well its
predictions match the ground truth. Given a dataset $\D$ and a binary classifier
$f$, we profile $f$'s predictions on $\D$ using $\mathit{TP}$, $\mathit{TN}$,
$\mathit{FP}$, and $\mathit{FN}$, which denote the numbers of true positives,
true negatives, false positives, and false negatives, respectively. Further,
$\mathit{TPR}$, $\mathit{TNR}$, $\mathit{FPR}$, and $\mathit{FNR}$ denote the
rate of true positives, true negatives, false positives, and false negatives,
respectively.

\smallskip
\paragraph{Metrics.}
We measure correctness through well-studied metrics
in literature~\cite{accuracy} (Figure~\ref{fig:correctness}). Intuitively,
\emph{accuracy} captures the overall correctness of the predictions made by a
classifier; \emph{precision} captures ``preciseness'', i.e., the fraction of
positive predictions that are correctly predicted as positive; and \emph{recall}
captures ``coverage'', i.e., the fraction of positive tuples that are correctly
predicted as positive. The \emph{F$_1$-score} is the harmonic mean of precision
and recall. While accuracy is an effective correctness metric when datasets have
a balanced class distribution, it can be misleading for imbalanced datasets,
which is found frequently in real-world scenarios. In such cases, precision,
recall, and F$_1$-score, together, are more insightful.

\subsection{Fairness}\label{subsec:fairnessmetrics}
Fairness in classifier predictions typically targets sensitive attributes, such
as gender, race, etc. Example~\ref{ex:discrimination} highlights how a
classifier can discriminate despite being reasonably accurate.

\begingroup
\setlength{\tabcolsep}{1.2pt}
\begin{figure*}[t]
\centering\resizebox{\textwidth}{!}{{
\begin{tabular}{p{0.4cm}llllllllllllllllll}
\toprule

&\multirow[b]{2}{*}{\bf{Fairness notion}} & \multirow[b]{2}{*}{\bf{Metric}} & \multicolumn{3}{c}{\bf{Granularity}} && \multicolumn{3}{c}{\bf{Causal hierarchy}} &&\multicolumn{4}{c}{\bf{Additional requirements}}\\
\cline{4-6} \cline{8-10} \cline{12-15}
&&&\multicolumn{2}{c}{\footnotesize group} &\footnotesize \: individual
&&\footnotesize observation &\footnotesize intervention &\footnotesize counterfactual&  
&\footnotesize {\makecell[l]{prediction\\probability} }
&\footnotesize {\makecell[l]{causality\\model}}
&\footnotesize {\makecell[l]{resolving\\attribute}}
&\footnotesize {\makecell[l]{similarity\\metric}}\\
\cline{4-5}
&&&\footnotesize {\makecell[l]{demography-\:\: \\aware}} &\footnotesize {\makecell[l]{error-\\aware} } \\
\midrule

\multirow{20}{*}{\makecell{\turnr{non-causal}}}
&conditional statistical parity~\cite{csp}   		& conditional statistical parity										&\cmark &		&		&&\cmark &       &			&& 		& 		& 		&   		\\
\rowcolor{gray!25}\cc
&demographic parity$^\dagger$~\cite{dwork}    		& disparate impact~\cite{zafar}, CV score~\cite{calders2009} 			&\cmark &		& 		&&\cmark &		 &			&& 		& 		& 		&&   		\\
&intersectional fairness~\cite{foulds2020}   		& differential fairness													&\cmark &		&		&&\cmark &       &			&& 		& 		& 		&   		\\
&conditional accuracy equality~\cite{berk}   		& false discovery/omission rate parity									&		&\cmark &		&&\cmark &       &			&& 		& 		& 		&   		\\ 
&predictive parity~\cite{chouldechova}       		& false discovery rate parity											&		&\cmark &	  	&&\cmark &       &			&& 		& 	    & 		&   		\\ 
&overall accuracy equality~\cite{berk}       		& balanced classification rate~\cite{friedler}                          &		&\cmark &		&&\cmark &       &			&& 		& 		& 		&   		\\ 
&treatment equality~\cite{berk}              		& ratio of false negative and false positive                            &		&\cmark & 	    &&\cmark &       &			&& 		& 		& 		&   		\\ 
\rowcolor{gray!25}\cc
&equalized odds~\cite{hardt}                  		& true positive/negative rate balance                                   &		&\cmark & 	    &&\cmark &       &			&& 		& 	    & 		&&   		\\ 
&equal opportunity$^\ddagger$~\cite{hardt} 			& true negative rate balance 										    &		&\cmark & 	    &&\cmark &       &			&& 		& 	    & 	    &   		\\ 
&resilience to random bias~\cite{fish}       		& resilience to random bias											    &		&\cmark & 		&&\cmark &       &			&& 		& 		& 		&   		\\ 
&preference-based fairness~\cite{zafarnips}  		& group benefit   													    &		&\cmark & 	    &&\cmark &       &			&& 		& 	    & 	    &   		\\ 
&calibration~\cite{chouldechova}             		& calibration  														    &		&\cmark & 	    &&\cmark &       &			&&\cmark& 	    & 	    &   		\\ 
&calibration within groups~\cite{kleinberg}  		& well calibration  													&		&\cmark & 	    &&\cmark &       &			&&\cmark& 	    & 	    &   		\\ 
&positive class balance~\cite{kleinberg}     		& fairness to positive class 										    &		&\cmark &		&&\cmark &       &			&&\cmark& 		& 		&   		\\ 
&negative class balance~\cite{kleinberg}     		& fairness to negative class 											&		&\cmark &		&&\cmark &       &			&&\cmark& 		& 		&   		\\ 
\rowcolor{gray!25}\cc
&individual discrimination$^{\dagger\dagger}$~\cite{themis}& individual discrimination 										&		&		&\cmark	&&\cmark &		 &			&&		&		&		&& 			\\
&metric multifairness~\cite{kim}             		& metric multifairness    											    & 	    &		&\cmark &&\cmark & 		 &			&&      &		& 	    & \cmark    \\
&fairness through awareness~\cite{dwork}     		& fairness through awareness 											&		&		&\cmark &&\cmark & 		 &			&&      &		& 	    & \cmark    \\
&fairness through unawareness~\cite{kusner}  		& Kusner et al.~\cite{kusner} 											& 	    &		&\cmark &&\cmark & 		 &			&&      &		&  	    &			\\ \cline{1-17}

\multirow{15}{*}{\makecell{\turnr{causal}}}
&proxy fairness~\cite{proxy}                 		& proxy fairness   													    &\cmark	&		&		&&		 &\cmark&			&&		 &\cmark &		&			\\
\rowcolor{gray!25}\cc
&total causal effect~\cite{pearl}					& total effect															&\cmark &		&		&&		 &\cmark&			&&		 &\cmark &		&&			\\
&direct causal effect~\cite{pearl}					& natural direct effect													&\cmark &		&		&&		 &\cmark&			&&		 &\cmark &		&			\\
&indirect causal effect~\cite{pearl}				& natural indirect effect												&\cmark &		&		&&		 &\cmark&			&&		 &\cmark &		&			\\
&path-specific fairness~\cite{zhang2017}          	& path specific effect										 		    &\cmark & 		&		&&		 &\cmark&			&&	     &\cmark &		&   		\\
&unresolved discrimination~\cite{proxy}      		& causal risk difference~\cite{rdc} 									&\cmark	&		&		&&		 &\cmark&			&& 		 & 		 &\cmark&   		\\
&interventional/justifiable fairness~\cite{Salimi}  & ratio of observable discrimination									&\cmark &	    &		&&		 &\cmark&			&&       & 		 &\cmark&   		\\
&fair on average causal effect~\cite{khademi2019}	& fair on average causal effect											&\cmark	&		&		&&		 &\cmark&			&&		 &\cmark &		&   		\\
&non-discrimination criterion~\cite{zhang2017kdd}	& non-discrimination criterion											&\cmark	&		&		&&		 &\cmark&			&&		 &\cmark &		&	   		\\
&equality of effort~\cite{huan2020}					& equality of effort													&\cmark	&		&		&&       &\cmark&			&&		 &\cmark &		&	   		\\
&counterfactual effects~\cite{zhang2018}			& counterfactual direct/indirect effect									&\cmark	&		&		&&		 &		&\cmark		&&		 &\cmark &					\\
&counterfactual error rates~\cite{zhang2018nips}    & counterfactual error rates		                             		&		&\cmark	&		&&		 &		&\cmark		&&		 &\cmark &		&	  		\\
&counterfactual fairness~\cite{kusner}       		& counterfactual effect~\cite{pcf} 									    &	    &		&\cmark &&		 &		&\cmark 	&&	     &\cmark &		&    		\\
&path-specific counterfactuals~\cite{pcf}    		& counterfactual effect                       	                        & 		&		&\cmark &&		 &		&\cmark 	&&		 &\cmark &		&    		\\
&individual direct discrimination~\cite{zhang2016}	& individual direct discrimination										&		&		&\cmark	&&		 &\cmark& 			&& 		 &\cmark &		&			&&\\

\bottomrule

\end{tabular}}} 
\vspace{-3mm}
\caption{We categorize fairness notions and metrics in the literature according
to the type of association considered between attributes (causal or non-causal)
and list other properties: granularity (group or individual), and position in
the causal hierarchy based on required domain knowledge (observation,
intervention, or counterfactual). Here, we use intervention in the context of
causal inference that requires interventions to adhere to the causal model of
the data. We further partition group-level notions based on their strategy to
measure discrimination (demography- or error-aware). All notions require
knowledge of the sensitive attributes and the classifier predictions. Some
definitions place additional requirements, shown in the rightmost four columns.
For our evaluation, we choose five metrics (Figure~\ref{fig:fairness}) that
cover the highlighted definitions. ($^\dagger$also known as statistical parity;
$^\ddagger$also known as predictive equality; $^{\dagger\dagger}$also known as
causal discrimination).}
\vspace{-3mm}
\label{table:definitions}	
\end{figure*}
\endgroup

\subsubsection{Fairness notions.} \label{pg:r3o3_2}
\looseness-1 
Fairness is not entirely objective, and societal
requirements and legal principles often demand different characterizations.
Fairness is also a relatively new concern within the research community.
Consequently, a large number of different fairness definitions have emerged,
along with a variety of quantifying metrics. Figure~\ref{table:definitions}
presents a list of 34 fairness notions and corresponding metrics that have
been studied in the literature. We primarily categorize these notions based
on the association considered between the sensitive attribute and the
prediction: some notions analyze the source of discrimination through
\emph{causal} relationships among the attributes, while others compute
\emph{non-causal} associations through observed statistical correlations. We
highlight further distinction among the notions based on their granularity,
position in the causal hierarchy, and additional requirements they impose:

\vspace{1mm}
\textbf{Granularity.} We classify fairness notions based on the granularity of
their target: \emph{group} fairness characterizes if any demographic group,
collectively, is being discriminated against; \emph{individual} fairness
determines if similar individuals are treated similarly, regardless of the
values of the sensitive attribute. Group-based notions can further be
categorized as \emph{demography-aware}, which consider the distribution of
outcomes among groups to measure fairness, and \emph{error-aware}, which
compare the error rates for each group.

\looseness-1 \textbf{Causal hierarchy.} A key feature of the fairness notions is
their position in the causal hierarchy that is determined by the extent of
domain knowledge they require. We highlight this distinction using Pearl's
ladder of causation~\cite{pearl}, a hierarchy of three levels of increasing
complexity: (1)~observation (2)~intervention, and (3)~counterfactual. Notions at
the \emph{observation} level can be computed entirely from observational data.
Notions at the \emph{intervention} level use both observational data and the
underling causal structure, i.e., an abstract model that shows whether any
causal relationship exists between attributes. Lastly, notions at the
\emph{counterfactual} level demand observational data, and full specification of
the underlying causal model denoting the exact functional relationships between
attributes.

\looseness-1
\textbf{Additional requirements.} 
All notions require information on the sensitive attribute and the classifier
predictions. Some notions impose additional requirements, such as causality
models or causal structure~\cite{pearl}, resolving attributes that mediate the
relationship between the sensitive attribute and the outcome in
non-discriminatory ways~\cite{rdc}, similarity metric between
individuals~\cite{dwork}, etc.

\vspace{-1mm}
\subsubsection{Fairness metrics.} 
\looseness-1 While Figure~\ref{table:definitions} highlights a wide range of
proposed fairness notions, Prior works~\cite{friedler, majumder2021} have shown
that a large number of metrics (and their notions) strongly correlate with one
another, and, thus, are highly redundant. For our evaluation, we carefully
selected five fairness metrics (Figure~\ref{fig:fairness}) that are most
prevalent in the literature and capture commonly occurring discriminations in
binary classification~\cite{chouldechova}. We briefly review these metrics and
refer to \appOrTechRep for a detailed discussion.

\smallskip
\noindent\textbf{Non-causal metrics} depend entirely on empirical
data and look for statistical relationships between the sensitive attribute and
the prediction. We experiment with the following non-causal metrics:

\looseness-1
\emph{Disparate Impact (DI)} compares the distribution of predictions among
sensitive groups and captures if they are independent of the
sensitive attribute. Specifically, $\mathit{DI}$ computes the ratio of empirical
probabilities of receiving positive predictions between the unprivileged and the
privileged groups (Figure~\ref{fig:fairness}, row~1). \di is also commonly known
by its corresponding notion, demographic parity~\cite{dwork}.

\emph{True Positive Rate Balance (\tprb) and True Negative Rate Balance (\tnrb)}
measure discrimination as the difference in $\mathit{TPR}$ and $\mathit{TNR}$,
respectively, between the privileged and the unprivileged groups
(Figure~\ref{fig:fairness}, rows~2--3). These metrics are also known as
equalized odds~\cite{hardt}, the notion they jointly measure.

\emph{Individual Discrimination (\id)~\cite{themis}} checks whether assigning
different values to the sensitive attribute changes the prediction for
an individual. Specifically, $\mathit{ID}$ is computed as the fraction of tuples
for which changing the sensitive attribute causes a change in the prediction for
otherwise identical data points (Figure~\ref{fig:fairness}, row~4).

\begingroup
\setlength{\tabcolsep}{2.5pt}
\renewcommand{\arraystretch}{0.0}
\begin{figure*}[t]
    \centering 
		\resizebox{0.95\textwidth}{!}{\small
		\begin{tabular}{llll<{\hspace{0.7mm}}l}
				\toprule
				\bf{Metric}		& \bf{Definition}	 & \bf{Fairness notion}  &\bf{Range} &\bf{Interpretation}\\ 
				\midrule
				Disparate Impact ($\mathit{DI}$)~\cite{feldman} & \makecell[l]{\footnotesize
				$\displaystyle\frac{Pr(\hat{Y} {=} 1 \mid S {=} \prot)}{ Pr(\hat{Y} {=} 1
				\mid S {=} \priv)}$} & demographic parity & [0, $\infty$) & 
				\makecell[l]{$\mathit{DI} = 1 $~~~$\rightarrow$ completely fair\\
							 $\mathit{DI} = 0 $~~~$\rightarrow$ completely unfair\\
							 $\mathit{DI} = \infty\rightarrow$ completely unfair}\\
				\hline
				True Positive Rate Balance ($\mathit{TPRB}$)~\cite{hardt} &
				{\footnotesize$Pr(\hat{Y} {=} 1 \mid Y {=} 1, S {=} \priv) -
				Pr(\hat{Y} {=} 1 \mid Y {=} 1, S {=} \prot)$} & equalized odds & [-1, 1] &
				\makecell[l]{$\lvert \mathit{TPRB} \rvert$ = 0 $\rightarrow$ completely fair\\
							 $\lvert \mathit{TPRB} \rvert$ = 1 $\rightarrow$ completely unfair}  
				
				\\
				\hline
				True Negative Rate Balance ($\mathit{TNRB}$)~\cite{hardt} & {\footnotesize
				$Pr(\hat{Y} {=} 0 \mid Y {=} 0, S {=} \priv) - Pr(\hat{Y} {=} 0
				\mid Y {=} 0, S {=} \prot)$} & equalized odds & [-1, 1] & 
				\makecell[l]{$\lvert \mathit{TNRB} \rvert$ = 0 $\rightarrow$ completely fair\\
							 $\lvert \mathit{TNRB} \rvert$ = 1 $\rightarrow$ completely unfair}  
				    
				\\
				\hline
				Individual Discrimination ($\mathit{ID}$)~\cite{themis} & {\footnotesize
				$\frac{| Q |}{| \D |},$ given $Q = \{a \in \D \mid \exists b :
				\X_a {=} \X_b \wedge S_a {\neq} S_b\ \wedge \hat{Y_a} {\neq}
				\hat{Y_b}\}$} & individual discrimination & [0, 1] & 
				\makecell[l]{$\mathit{ID}$ = 0 $\rightarrow$ completely fair\\
							 $\mathit{ID}$ = 1 $\rightarrow$ completely unfair}     \\
				\\
				\hline
				Total Effect ($\mathit{TE}$)~\cite{pearl} & {\footnotesize
				$Pr(\hat{Y}_{S=\priv} = 1) - Pr(\hat{Y}_{S=\prot} = 1)  $} 
				& total causal effect & [-1, 1] & 
				\makecell[l]{$\lvert \mathit{TE} \rvert$ = 0 $\rightarrow$ completely fair\\
							$\lvert \mathit{TE} \rvert$ = 1 $\rightarrow$ completely unfair}     \\

				\bottomrule
			\end{tabular}
		} 
	\vspace{-3mm}
	\caption{List of fairness metrics we use to evaluate fair classification
	approaches. These metrics effectively contrast between causal and non-causal
	associations; and cover group- and individual-level discrimination,
	observation- and intervention-level techniques.}
	\vspace{-2mm}
	\label{fig:fairness}

\end{figure*}
\endgroup

\vspace{1mm}
\noindent\textbf{Causal metrics} determine discrimination by considering the
causal relationship between the sensitive attribute and the prediction, as
opposed to their statistical dependencies. As non-causal metrics cannot reason
about whether a sensitive attribute is the true cause of discrimination, causal
metrics address this limitation through additional domain knowledge. We
experiment with \emph{Total Effect (\te)~\cite{pearl}}, a causal metric that
measures discrimination as the causal influence of the sensitive attribute on
prediction. It measures the effect of interventions to the sensitive attribute
on the prediction, to determine the extent of causal influence
(Figure~\ref{fig:fairness}, row 5). $\mathit{TE}$ is often decomposed into
indirect (causal influence mediated by other attributes) and direct (influence
that is not mediated) effects, or path-specific effects (influence through
particular causal pathways) that are needed in many real-world
situations~\cite{texas2015,zhang2018nips}.

\vspace{1mm} \looseness-1\noindent\emph{Discussion on metric choices.} We
select metrics to cover a variety of categories in our classification,
including causal and non-causal associations, group- and individual-level
fairness, and observational and interventional techniques (highlighted rows in
Figure~\ref{table:definitions}). Other causal notions can also address the
limitations of non-causal metrics, but they often require additional
information (e.g., structural equations for counterfactuals) to be computed
from observational data. We choose metrics that are feasible within the scope
of our experiments, and exclude ones that make strong and impractical
assumptions about the problem setting~\cite{pearl}. For similar reasons, we do
not include individual-level metrics that depend on counterfactuals or
similarity measures between individuals.

\vspace{-2.5mm}
\section{Fair Classification Approaches}\label{sec:algorithm}
%

Fair classification techniques vary in the fairness notions they target and the
mechanisms they employ. We categorize approaches based on the stage when
fairness-enforcing mechanisms are applied. (1)~\emph{Pre-processing} approaches
attempt to repair biases in the data before training; (2)~\emph{in-processing}
approaches modify the learning procedure to include fairness considerations;
(3)~\emph{post-processing} approaches modify the classifier predictions. For
our evaluation, we select approaches that span all three stages and target a
representative variety of fairness notions, including causal and non-causal
associations, observation- and intervention-level techniques.
Figure~\ref{table:methods} overviews our chosen approaches. We proceed to
provide a high-level description of the approaches in each category,
underscoring their similarities and differences. (Details are
in~\cite{technicalReport}).

\vspace{-1mm}
%

\def\A{\mathbf{A}} \def\I{\mathbf{I}}

\smallskip
\subsection{Pre-processing}    
\looseness-1 Pre-processing approaches are motivated from the fact that ML
techniques are data-driven and the predictions of a classifier reflect trends
and biases in the training data. Data management research most naturally fits in
this category. These approaches modify the data before training to remove
biases, which subsequently ensures that the predictions made by a learned
classifier satisfy the target fairness notion. The main advantage of
pre-processing is that it is model-agnostic, allowing flexibility in choosing
the classifiers based on the application requirements. However, since
pre-processing happens \emph{before} training and does not have access to the
predictions, these approaches are limited in the number of notions they can
support and do not always come with provable guarantees of fairness.

In our evaluation, we include three pre-processing approaches that enforce
non-causal fairness notions and two approaches that target causal notions. We
briefly discuss these approaches here.

\emph{\kam}~\cite{kamiran} is a pre-processing approach that enforces
demographic parity, a notion that ensures model prediction $\hat Y$ is
independent of the sensitive attribute $S$. Assuming that $\hat{Y}$ reasonably
approximates the ground truth $Y$, \kam argues that $\hat{Y}$ is likely to be
independent of $S$ when the classifier is deployed, if there is no dependency
between $Y$ and $S$ in the training data. To this end, \kam resamples the
training data $\D$ with a weighted sampling technique to ensure that $S$ and $Y$
are statistically independent.

\emph{\feld}~\cite{feldman} is another approach that enforces demographic
parity. It argues that demographic parity can be ensured if the marginal
distribution of each attribute $X \in \X$ is similar across the sensitive groups
in training data $\D$. Intuitively, if a model learns from such data, it is
likely to predict based on attributes that are independent of $S$, which in turn
satisfies demographic parity. To that end, \feld modifies the values for each
attribute $X$ until the marginal distributions are similar for the privileged
and unprivileged group. Unlike \kam that only resamples the tuples, \feld
modifies the training data. Further, \kam enforces demographic parity through
independence between $S$ and $Y$, while \feld reformulates it as an independence
condition between $\X$ and $S$.

\emph{\calm}~\cite{calmon} is one more approach targeting demographic parity.
The goal of this approach is to reduce the dependency between $S$ and $Y$ by
minimally perturbing the attribute values of $\X$ and $Y$ and without
significantly distorting the underlying data distribution. It utilizes the joint
distribution associated with $\D$ and a set of pre-defined distortion functions
to define the corresponding optimization problem for minimal repair. \calm uses
convex optimization techniques to solve this optimization problem and minimally
modifies $\X$ and $Y$ to achieve the target fairness goal. Among \kam, \feld,
and \calm, it is the only approach that modifies both training and test data.

\begin{sloppypar}
\emph{\zhangwu}~\cite{zhang2017,zhang2017kdd} proposes two methods that target
causal notions: path-specific fairness (\zhangpcft) and direct causal effect
(\zhangdcet). \zhangpcft enforces path-specific fairness by modifying $Y$ such
that all causal influences of $S$ over $Y$ are removed. It learns a causal graph
over $\D$ to discover (direct and indirect) causal associations between $Y$ and
$S$, and translates the minimal repair of $Y$ to a quadratic programming
problem. On the other hand, \zhangdcet aims to minimize the direct causal effect
of $S$ on $Y$. It determines a set of parents ($Q$) of $Y$ that blocks all
indirect causal paths from $S$ to $Y$ and uses $Q$ to compute the causal
effect on the direct path. Then, it modifies the distribution of $Y$ such that
the direct causal effect is below a user-defined threshold. \zhangwu is
different from all the aforementioned approaches as it enforces causal notions
using additional domain knowledge.
\end{sloppypar}

\sal~\cite{Salimi} enforces justifiable fairness, which prohibits causal
dependency between $S$ and $\hat Y$, except through a set of admissible
attributes $\A \in \X$. $\A$ is pre-defined such that the effect of $S$ on $\hat
Y$ through $\A$ is deemed non-discriminatory. All other attributes are
considered discriminatory and constitute the inadmissible set ($\I$). Like other
approaches, \sal assumes that $\hat Y$ is likely to be fair if a classifier is
trained on data $\D$ where $Y$ satisfies the target fairness notion. It enforces
justifiable fairness as a conditional independence and minimally modifies the
underlying data distribution such that $Y$ is conditionally independent of $\I$
given $\A$. \sal solves the optimization problem corresponding to minimal repair
using weighted maximum satisfiability (\salmst) and matrix factorization
(\salmft). Unlike \zhangwu, \sal does not require the entire causal graph and
repairs $\D$ by inserting or deleting tuples.

\def\ci{\perp\!\!\!\perp}

\begingroup
\renewcommand{\arraystretch}{1.2}
\setlength{\tabcolsep}{2.5pt}

  \begin{figure*}[t]
    \centering\resizebox{1\textwidth}{!}
	{
    	\begin{tabular}{@{}cllp{11cm}l<{\hspace{0.7mm}}l@{}}
    		\toprule	
			\textbf{{Stage}}  & \textbf{Approach} 		& \multicolumn{1}{l}{\textbf{Fairness notion(s)}} &\multicolumn{1}{l}{\textbf{Key mechanism}} && \multicolumn{1}{l}{\bf{Evaluated version(s)}} \\
    		\midrule		    		

			\multirow{9}{*}{\makecell{pre}}
			&\kam\cite{kamiran}    		& demographic parity 				   			& 	Apply weighted resampling over tuples in $\D$ to remove dependency between $S$ and $Y$.
																						&&	\begin{tabular}{@{}lp{4.4cm}@{}}
																								$\bullet$ \kamt & 
																							\end{tabular} 	 \\ \cline{2-6}
			&\feld\cite{feldman}    	& demographic parity 		 			   		& 	\begin{tabular}{@{}p{11cm}@{}}
																								Repair each $X \in \X$ independently s.t. $X$'s marginal distribution is indistinguishable across sensitive groups. Training and test data are both modified.
																						  	\end{tabular}		
																						&&	\begin{tabular}{@{}l p{4.4cm}@{}}
																								$\bullet$ \feldt &
																							\end{tabular} 	 \\ \cline{2-6}
			&\calm\cite{calmon}     	& demographic parity 				    		& 	\begin{tabular}{@{}p{11cm}@{}}
																								Modify $\X$ and $Y$ to reduce dependency between $Y$ and $S$, while preventing major distortion of the joint data distribution and significant change of the attribute values. Training and test data are both modified.
																						   	\end{tabular}	 
																						&&	\begin{tabular}{@{}lp{4.4cm}@{}}
																								$\bullet$ \calmt & 
																							\end{tabular} \\ \cline{2-6}
			&\multirow{2}*{\makecell[l]{\vspace{2mm}\\\zhangpcf\cite{
			zhang2017, zhang2017kdd}}}  & path-specific fairness						& 	\begin{tabular}{@{}p{11cm}@{}}
																								Exploit a (learned) causal model over the attributes to discover (direct and indirect) causal association between $Y$ and $S$. Modify $Y$ to remove such causal association.
			  																				\end{tabular} 
																						&&	\begin{tabular}{@{}lp{4.4cm}@{}}
																							$\bullet$ \zhangpcft & 
																							\end{tabular} \\ 
										&& direct causal effect 						& 	\begin{tabular}{@{}p{11cm}@{}}
																								Given a causal graph, identify the set of parents ($Q$) of $Y$ that blocks all indirect paths from $S$ to $Y$. Use $Q$ to compute the causal effect of $S$ on $Y$ through the direct path and modify $Y$ s.t. this effect is within allowable threshold.
																							\end{tabular} 
																						&&	\begin{tabular}{@{}lp{4.4cm}@{}}
																							$\bullet$ \zhangdcet & 
																							\end{tabular} \\ \cline{2-6}																				
			
			&\sal\cite{Salimi}     		& justifiable fairness  			    		& 	\begin{tabular}{@{}p{11cm}@{}}
																								Mark attributes as \emph{admissible} ($A$)---allowed to have causal association---or	\emph{inadmissible} ($I$)---prohibited to have causal association---with $Y$; repair $\D$ to ensure that $Y$ is conditionally independent of $I$, given $A$.
																								Reduce the repair problem to known problems.
																						  	\end{tabular}
																						&&	\begin{tabular}{@{}lp{4.4cm}@{}}
																								$\bullet$ \salmst & (Weighted maximum satisfiability)\\ 
																								$\bullet$ \salmft & (Matrix factorization)
																							\end{tabular} \\																						 																						
			\Xhline{3\arrayrulewidth}				
			\multirow{13}{*}{\makecell{in}}			
			&\zaf\cite{zafar, zafarDM}  &\makecell[l]{demographic parity\\ 
										equalized odds} 								& 	\begin{tabular}{@{}p{11cm}@{}}
																								Use tuple $t$'s distance from the decision boundary as a proxy of $\hat Y_t$. Model fairness violation by the correlation between this distance and $S$ over all tuples in $\D$. 
																								Solve variations of constrained optimization problem that either maximizes prediction accuracy under constraint on maximum fairness violation, or minimizes fairness violation under constraint on maximum allowable accuracy compromise.
																						   	\end{tabular} 
																						&&	\begin{tabular}{@{}lp{4.4cm}@{}}
																								$\bullet$ \zdifairt & (Maximize accuracy under constraint on demographic parity) \\
																								$\bullet$ \zdiacct  & (Maximize demographic parity under constraint on accuracy) \\
																								$\bullet$ \zdmt	  	& (Same as \zdifairt, but use misclassified tuples only)
																							\end{tabular}\\ \cline{2-6}
			&\zhang\cite{zhang}      	& equalized odds								& 	\begin{tabular}{@{}p{11cm}@{}}
																								Utilize adversarial learning to train classifier $f:f(\X, S) \rightarrow \hat Y$ and adversary $\alpha: \alpha(Y, \hat Y) \rightarrow \hat S$ together. Enforce fairness by ensuring that $\alpha$ cannot infer $S$ from $Y$ and $\hat Y$.
																						  	\end{tabular}	
																						&&	\begin{tabular}{@{}lp{4.4cm}@{}}
																								$\bullet$ \zhangt &
																							\end{tabular} \\ \cline{2-6}
			&\kearns\cite{kearns}     	& \makecell[l]{demographic parity\\ 
										predictive equality}							& 	\begin{tabular}{@{}p{11cm}@{}}
																								Use sensitive attribute(s) to construct a set of subgroups. Define fairness constraint s.t. the probability of positive outcomes (demographic parity) or $\mathit{FPR}$ (predictive equality) of each subgroup matches that of the overall population. 
																						 	\end{tabular}	
																						&&	\begin{tabular}{@{}lp{4.4cm}@{}}
																								$\bullet$ \kearnst & (For subgroups $\{\D_1, \D_2, \dots\}$ where each $\D_i \subset \D$, ensure that $\forall \D_i$, $\mathit{FPR} (\D_i) \approx \mathit{FPR} (\D)$) 
																							\end{tabular} \\ \cline{2-6}
			&\celis\cite{celis}      	&  \makecell[l]{equalized odds\\demographic parity\\predictive parity\\cond. acc. equality}					
																						& 	\begin{tabular}{@{}p{11cm}@{}}
																								Unify multiple fairness notions in a general framework by converting the fairness constraints to a linear form. 
																								Solve the corresponding linear constrained optimization problem s.t. prediction error is minimized under fairness constraints.	
																							\end{tabular}
																						&&	\begin{tabular}{@{}lp{4.4cm}@{}}
																								$\bullet$ \celist & (Enforce {}\small$Pr(Y {=} 0 \mid \hat Y {=} 1, S {=} 0) \approx Pr(Y {=} 0 \mid \hat Y {=} 1, S {=} 1))$
																							\end{tabular} \\ \cline{2-6}
			
			&\thomas\cite{thomas2019} 	& \makecell[l]{demographic parity\\ equalized odds\\  equal opportunity\\predictive equality}		
																						& 	\begin{tabular}{@{}p{11cm}@{}}
																								Compute worst possible fairness violation a classifier can incur for a set of parameters and pick parameters for which this worst possible violation is within an allowable threshold.																				
																						  	\end{tabular}
																						&&	\begin{tabular}{@{}lp{4.4cm}@{}}
																								$\bullet$ \thomasdpt & (Enforce demographic parity)	\\
																								$\bullet$ \thomaseot & (Enforce equalized odds)
																							\end{tabular}\\
																																						
			\Xhline{3\arrayrulewidth}
			
			\multirow{5}{*}{\makecell{post}}
			
			&\kamdp\cite{kamiran2012}	& demographic parity							& 	\begin{tabular}{@{}p{11cm}@{}}
																								Modify $\hat Y$ for tuples close to the decision boundary (i.e., subject to low prediction confidence) s.t. the probability of positive outcome is similar across sensitive groups.
																						  	\end{tabular}	
																						&&	\begin{tabular}{@{}lp{4.4cm}@{}}
																								$\bullet$ \kamdpt & 
																							\end{tabular} \\ \cline{2-6}
																						
			&\hardt\cite{hardt} 		& equalized odds		    					& 	\begin{tabular}{@{}p{11cm}@{}}
																								Derive new predictor based on $\hat Y$ and $S$ s.t. $\mathit{TPR}$ and $\mathit{TNR}$ are similar across sensitive groups.
																						  	\end{tabular}
																						&&	\begin{tabular}{@{}lp{4.4cm}@{}}
																								$\bullet$ \hardtt & 
																							\end{tabular} \\ \cline{2-6}
																						
    		&\pleiss\cite{pleiss}  		& \makecell[l]{equal opportunity\\ predictive equality} 
																						& 	\begin{tabular}{@{}p{11cm}@{}}
																								Modify $\hat Y$ for random tuples to equalize $\mathit{TPR}$ (or $\mathit{FPR}$) across sensitive groups.
																							\end{tabular}
																						&&	\begin{tabular}{@{}lp{4.4cm}@{}}
																								$\bullet$ \pleisst & (Equalize $\mathit{TPR}$)
																							\end{tabular} \\
    		\bottomrule
    	\end{tabular}
	} 
	 \vspace{-3mm}
	 \caption{List of fair approaches, fairness notions they support, and
	 high-level descriptions of the mechanisms they apply to ensure fairness.
	 According to the stage of the classifier pipeline where fairness-enhancing
	 mechanism is applied, these approaches are divided into three groups:
	 (1)~pre-processing, 
	 (2)~in-processing, and 
	 (3)~post-processing. 
	 In the rightmost column, we list the variations
	 of each approach that we consider in our evaluation. We denote in the
	 superscript the fairness notion that a specific variation is designed to
	 support.}
	 \vspace{-3mm}
    \label{table:methods} 
  \end{figure*}
\endgroup

\vspace{-1.5mm}
%
\vspace{-3mm}
\subsection{In-processing}

In-processing approaches are most favored by the machine learning
community~\cite{zafar, celis, kearns, zhang} and the majority of the fair
classification approaches fall under this category. In-processing takes place
within the training stage and fairness is typically added as a constraint to the
classifier's objective function (that maximizes correctness). The advantage of
in-processing lies precisely in the ability to adjust the classification
objective to address fairness requirements directly, and, thus has the potential
to provide guarantees. However, in-processing techniques are model-specific and
require re-implementation of the learning algorithms to include the fairness
constraints. This hinges on the assumption that the model is replaceable or
modifiable, which may not always be the case. We choose and discuss five
different in-processing approaches and their variants, to best highlight the
variety of existing techniques.

\looseness-1 \emph{\zaf}~\cite{zafar,zafarDM} proposes two methods to enforce
demographic parity (\zdi) and equalized odds (\zdmt). Both utilize tuples'
distance from the decision boundary as a proxy of $\hat Y$ to model fairness
violations, translate their corresponding fairness notion to a convex function
of the classifier parameters. \zdi solves the resulting constrained optimization
problem to compute optimal classifier parameters that either maximizes
prediction accuracy under fairness constraints (\zdiacct), or minimizes fairness
violation under constraints on accuracy compromise (\zdifairt). \zdmt only
computes parameters that maximize prediction accuracy under fairness
constraint~\cite{dccp}.

\emph{\zhang}~\cite{zhang} enforces the notion of equalized odds. It leverages
\emph{adversarial learning}, a technique where a classifier and an adversary
with mutually competing goals are trained together. Given $\D$, the goal of the
classifier is to maximize the accuracy of $\hat Y$, while the adversary attempts
to correctly predict $S$ using $\hat Y$ and $Y$. \zhang utilizes gradient
descent techniques~\cite{sgd} to compute the classifier's optimal parameters
such that $\hat Y$ does not contain any information about $S$ that the adversary
can exploit. 

\emph{\kearns}~\cite{kearns} is an in-processing approach that either enforces
demographic parity, or predictive equality (i.e., equal $\mathit{FPR}$ for the
sensitive groups). \kearns aims to approximately enforce the target fairness
notion within a set of subgroups defined using one or more (user-specified)
sensitive attributes. To that end, \kearns solves a constrained optimization
problem to obtain optimal classifier parameters such that the proportion of
positive outcomes (demographic parity) or FPR (predictive equality) is
approximately equal to that of the entire population.

\emph{\celis}~\cite{celis} accommodates a wide range of notions: predictive
parity, demographic parity, equalized odds, and conditional accuracy equality.
It reduces all fairness notions to linear forms and solves the corresponding
convex optimization problem using Lagrange multipliers~\cite{lagrange1976} to
minimize prediction error under fairness constraints. Unlike prior approaches
that only enforce specific fairness notions, \celis is designed to support a
wide variety of fairness notion within a single framework.

\emph{\thomas}~\cite{thomas2019} is another approach that provides a general
framework to accommodate a large number of notions. It supports demographic
parity, equalized odds, equal opportunity, and predictive equality. Given $\D$
and a target fairness notion, \thomas ensures that a classifier $f$ trained on
$\D$ only picks solutions that satisfy the fairness notion with high
probability. \thomas computes an upper bound (with high confidence) of the
maximum possible fairness violation that a classifier can incur at test time,
and returns optimal classifier parameters for which this worst possible
violation is within an allowable threshold.
\vspace{-1mm}
%
\subsection{Post-processing}
Post-processing approaches are model-agnostic and enforce fairness by
manipulating predictions made by an already-trained classifier. Their
benefit is that they do not require classifier retraining. However, since
post-processing is applied in a late stage of the learning process, it offers
less flexibility than pre- and in-processing. We briefly describe the techniques
behind the three post-processing approaches we evaluate.
 
\emph{\kamdp}~\cite{kamiran2012} targets demographic parity based
on the intuition that discriminatory decisions are most often made for tuples
close to the decision boundary, because the prediction confidence (i.e., the
probability of belonging to the predicted class) is low for those tuples. Given
a classifier, \kamdp derives a critical region around the decision boundary and
randomly modifies $\hat Y$ for tuples in that region until the probability of
positive outcome is similar across sensitive groups, i.e., demographic parity is
achieved.

\emph{\hardt}~\cite{hardt} enforces equalized odds through modifying the
predictions $\hat Y$. Given access to $Y$ and $S$ from the training data $\D$,
\hardt learns the parameters of a new mapping $g:g(\hat Y, S) \rightarrow \tilde
Y$ to replace $\hat Y$ such that $\mathit{TPR}$ and $\mathit{TNR}$ are equalized
across the sensitive groups. The new mapping is learned by solving a linear
program.

\emph{\pleiss}~\cite{pleiss} enforces equal opportunity (equal $\mathit{TPR}$
across the sensitive groups) or predictive equality (equal $\mathit{FPR}$ across
the sensitive groups), while maintaining the consistency (i.e., calibration)
between the classifier's prediction probability for a class with the expected
frequency of that class. To that end, \pleiss modifies $\hat Y$ for a random
subset of tuples within the group with higher $\mathit{TPR}$ (or lower
$\mathit{FPR}$) until $\mathit{TPR}$ (or $\mathit{FPR}$) is equalized.
 
%

\vspace{3mm}
\emph{Other approaches.} \looseness-1 We evaluate and discuss more fair
approaches (not in Figure~\ref{table:methods}) in \appOrTechRep. While other
fair approaches exist, some are incorporated in the ones we
evaluate~\cite{calders2009, calders2010, calders2010three}, while others are
empirically inferior~\cite{kamishima2012}, offer weaker
guarantees~\cite{agarwal2018, quadrianto2017}, do not offer a practical
solution~\cite{wood2017, noriega2019}, or do not apply to the classification
setting~\cite{samadi2018,louizos2015, goh2016,lahoti2020}. Some make strong
assumptions about the problem setting~\cite{kusner, kilbertus2017, nabi,
chiappa, russell2017, zhang2018, zhang2018nips, mishler2021}, or require
additional information~\cite{zemel2013, dwork, lahoti2019, ruoss2020}, which are
dataset-specific and hinge on domain knowledge.

\section{Evaluation and Analysis}\label{sec:evaluation}
%

In this section, we present results of our comparative evaluation over 18
variations of fair classification approaches as listed in
Figure~\ref{table:methods}. The objectives of our performance evaluation are:
(1)~to contrast the effectiveness of all approaches in enforcing fairness and
observe correctness-fairness tradeoffs, i.e., the compromise in correctness to
achieve fairness (Section~\ref{subsec:correct-fair}), (2)~to contrast their
efficiency and scalability with varying dataset size and dimensionality
(Section~\ref{subsec:runtime}), (3)~to compare robustness against errors in
training data (Section~\ref{sec:robustness}), (4)~to compare the sensitivity of
pre- and post-processing approaches to the choice of ML models
(Section~\ref{sec:resilience}), and (5)~to contrast stability (lack of
variability) over different partitions of training data and to contrast data
efficiency (dependence on dataset size) of all approaches
(Section~\ref{sec:others}). Our results affirm and extend previous results
reported by the evaluated approaches.

\looseness-1 Additionally, we present a comparative analysis, focusing on the
stage dimension (pre, in, and post). Our analysis highlights findings that
explain the behavior of fair approaches in different settings. For example, we
find that the impact of enforcing a specific fairness notion can be explained
through the score of a fairness-unaware classifier for that notion: larger
discrimination by the fairness-unaware classifier indicates that a fair approach
that targets that notion will likely incur higher drop in accuracy. Further, we
provide novel insights that underscore the strengths and weaknesses across pre-,
in-, and post-processing approaches. We find that all approaches behave
unpredictably in the presence of corrupt data; however, post-processing is
generally more robust than pre-processing and in-processing.

Next we provide details on our experimental settings: evaluated approaches,
their implementation details, evaluation metrics, and the datasets. Then we
present our empirical findings.

\subsection{Experimental Settings}
\paragraph{Approaches.}\looseness-1 We evaluated 18 variants of \countClassifier
fair classification approaches (Figure~\ref{table:methods}). Pre- and
post-processing approaches require a classifier to complete the model pipeline
and we used logistic regression (LR) as the classifier. This is in line with the
evaluations of the original papers as they all use LR. Moreover, to contrast all
fair approaches against a fairness-unaware approach, we trained an unconstrained
LR classifier over each dataset. Hyper-parameter settings of all approaches are
detailed in \appOrTechRep.

\vspace{1mm}
\paragraph{System and implementation.} We conducted the experiments on a machine
equipped with Intel(R) Core(TM) i5-7200U CPU (2.71 GHz, Quad-Core) and 8 GB RAM,
running on Windows 10 (version 1903) operating system. We collected some of the
source code from the authors' public repositories, some by contacting the
authors, and the rest from the open source library AI Fairness 360~\cite{aif}
(additional details are in \appOrTechRep). All approaches are implemented in
Python. We implemented the fairness-unaware classifier \orig using Scikit-learn
(version 0.22.1) in Python 3.6. Implementations of all these approaches use a
single-threaded environment, i.e., only one of the available processor cores is
used. We used the open source library DoWhy~\cite{sharma2020} to compute causal
quantities. We implemented the evaluation scripts in 
Python 3.6~\cite{coderepo}.

\vspace{1mm}

\paragraph{Metrics.} \looseness-1 We evaluated all approaches using
four correctness metrics (Figure~\ref{fig:correctness}) and five fairness
metrics (Figure~\ref{fig:fairness}). We normalize fairness metrics to share
the same range, scale, and interpretation. We report $\mathit{DI}^* =
\min(\mathit{DI}, \frac{1}{\mathit{DI}})$, which ensures that low fairness with
respect to $\mathit{DI}$ ($\mathit{DI}\rightarrow 0$ and $\mathit{DI}\rightarrow
\infty$) is mapped to low values for $\mathit{DI}^*$. Further, we report $1 -
\lvert \mathit{TPRB}\rvert$, $1 - \lvert \mathit{TNRB} \rvert$, $1 -
\mathit{ID}$, $1 - \lvert\mathit{TE} \rvert$; this way, high discrimination with
respect to, say, $\mathit{TPRB}$, maps to low fairness value in $1 -
\lvert\mathit{TPRB}\rvert$. Moreover, $\mathit{ID}$ requires two parameters: a
confidence fraction and an error-bound. We choose a confidence of 99\% and an
error-bound of 1\%, which implies that discrimination computed using
$\mathit{ID}$ is within 1\% error margin of the actual discrimination with 99\%
confidence.

\vspace{1mm}

\paragraph{Datasets.} 
Our evaluation includes 3 real-world datasets, summarized in
Figure~\ref{table:dataset}. Each dataset contains varied degrees of real-world
biases, allowing for the evaluation of the fair classification approaches
against different scenarios. Furthermore, these datasets are well-studied in the
fairness literature and are frequently used as benchmarks to evaluate fair
classification approaches~\cite{friedler, jones2020, mehrabi2019}.

\emph{Adult}~\cite{adult} is extracted from the 1994 US census and contains
information about individuals over demographic and occupational attributes such
as race, sex, education level, occupation, etc. Adult reflects historical
gender-based income inequality: 11\% of the females report high income ($Y =
1$), compared to 32\% of the males. Hence, we choose \texttt{sex} as the
sensitive attribute with \texttt{female} as the unprivileged and \texttt{male}
as the privileged group. 

\emph{COMPAS}~\cite{compas} contains background information---such as age, sex,
prior convictions, etc.---of defendants arrested in 2013--2014 and their
subsequent assessment scores by the COMPAS recidivism tool~\cite{risktool}. The
data contains racial bias: 51\% African-Americans re-offend within two years
($Y = 0$), compared to 39\% in other races. We select \texttt{race} as the
sensitive attribute with \texttt{African-American} as the unprivileged and all
other races as the privileged group.

\emph{German}~\cite{Kaggle} contains records of individuals applying for credit
or loan to a bank, with attributes age, sex, credit history, savings, etc. 70\%
of the entire population are of low credit risk ($Y = 1$), with this percentage
being slightly lower for females than males: 65\% vs 71\%. Hence, we choose
\texttt{sex} as the sensitive attribute with \texttt{female} as the unprivileged
and \texttt{male} as the privileged group.

\begingroup
\renewcommand{\arraystretch}{1.2}
\setlength{\tabcolsep}{1.5pt}
  \begin{figure}[t]
    \centering
    \resizebox{0.47\textwidth}{!}{\small
    \begin{tabular}{@{}llrrlllll@{}}
		\toprule
     \multirow{2}{*}{\textbf{Dataset}}  & \multirow{2}{*}{\makecell[c]{\textbf{Size}\\\textbf{(MB)}}} & \multicolumn{1}{c}{\multirow{2}{*}{$|\D|$}} & 
	 \multicolumn{1}{c}{\multirow{2}{*}{$|\X|$}} & \multicolumn{1}{c}{\multirow{2}{*}{$S$}}
     &\multicolumn{2}{c}{\textbf{Sensitive groups}} & \multicolumn{1}{c}{\multirow{2}{*}{\textbf{Target task}}} \\
 	\cline{6-7}
 	&&&&&\textbf{Unprivileged} & \textbf{Privileged} &\\
		\midrule
		Adult    & 3.70  & 45,222    &9   & Sex  & Female & Male                 & ~~Income $\geq \$$50K     \\
		COMPAS   & 0.18  & 7,214     &3   & Race & \makecell[l]{African-American}~~ & Others   & ~~Risk of recidivism     \\
		German   & 0.04  & 1,000     &9   & Sex  & Female & Male                 & ~~Credit risk            \\\bottomrule
    \end{tabular}} 
    \vspace{-2mm}
    \caption{Summary of the datasets. We choose our datasets to be varied in
    size, number of data points, number of attributes, and different instances
    of sensitive-attribute-based discrimination. We provide the target
    prediction tasks in the rightmost column.}
    \vspace{-3mm}	
  \label{table:dataset}
  \end{figure}
\endgroup

\smallskip

\paragraph{Train-validation-test setting.} The train-test split for each
dataset was 70\%-30\% (using random selection) and we validated each classifier
using 5-fold cross validation.

%

\subsection{Correctness and Fairness} \label{subsec:correct-fair}
\begin{figure*}[t]
  \centering
  \begin{subfigure}[t]{0.99\textwidth}
  \includegraphics[width=1\textwidth]{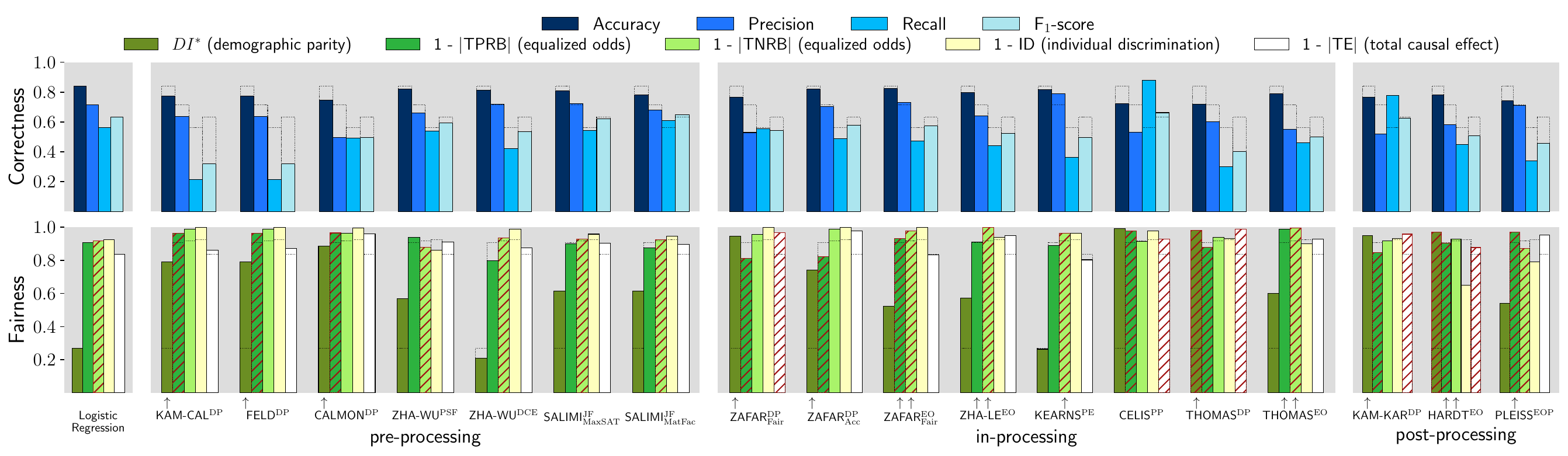}
  \vspace{-6mm}
  \caption{Adult}
  \label{fig:experiment_adult}
  \end{subfigure}

  \begin{subfigure}[t]{0.99\textwidth}
  \includegraphics[width=1\textwidth]{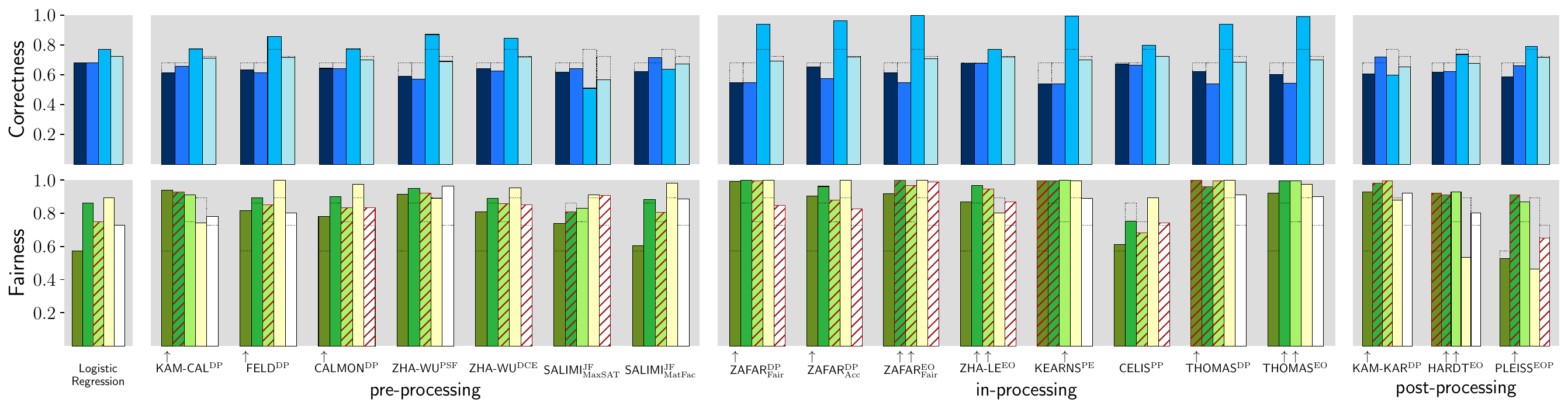}
  \vspace{-6mm}
  \caption{COMPAS}
  \label{fig:experiment_compas}
  \end{subfigure}
  
  \begin{subfigure}[t]{0.99\textwidth}
  \includegraphics[width=1\textwidth]{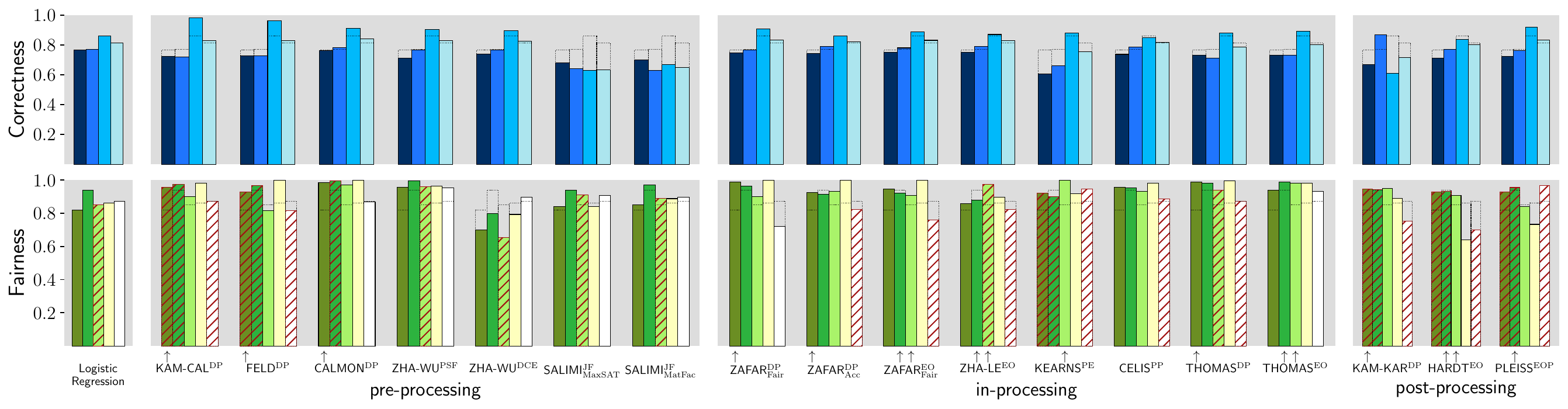}
  \vspace{-6mm}
  \caption{German}
  \label{fig:experiment_german}
  \end{subfigure}
  
  \vspace{-4mm} 
  
  \caption{Correctness and fairness scores of the 18 fair classification
  approaches over (a)~Adult, (b)~COMPAS, and (c)~German datasets. Higher scores
  for correctness (fairness) metrics correspond to more correct (fair) outcomes.
  The bars highlighted in red denote the reverse direction of the remaining
  discrimination---favoring the unprivileged group more than the privileged
  group. The arrows ($\uparrow$) denote the fairness metric(s) each approach is
  optimized for. The bar plots for \orig are overlaid for aiding visual
  comparison. } \vspace{-2mm}
  \label{fig:experiment_adult_compas_german}
  
\end{figure*}

Figure~\ref{fig:experiment_adult_compas_german} presents our correctness
and fairness results over all approaches and metrics across the 3 datasets.
Below, we discuss the key findings of this evaluation.

\smallskip

\noindent \emph{The fairness performance of fairness-\underline{unaware}
approaches influences the relative accuracy of fair approaches.} Classifiers
typically target accuracy as their optimization objective. Fair approaches,
directly or indirectly, modify this objective to target both fairness and
accuracy. When a fairness-unaware technique displays significantly different
performance across different fairness metrics (e.g., low fairness wrt \di and
high fairness wrt \tprb), this appears to translate to a significant difference
in the accuracy of fair approaches that target these fairness metrics (higher
accuracy drop for approaches that target \di, and lower drop for those that
target \tprb).

Figure~\ref{fig:experiment_adult} demonstrates this scenario for Adult. \orig
trained on this dataset achieves high fairness in terms of \tprb and \tnrb, but
exhibits very low fairness in terms of \di. We observe that the approaches that
optimize \di (such as \kamt and \calmt) demonstrate a much larger accuracy drop
than the approaches that target equalized odds (such as \zdmt,
\zhangt, and \kearnst). \zdiacct is an exception as it explicitly controls the
allowable accuracy drop. We hypothesize that in an effort to enforce fairness in
terms of \di, the corresponding approaches shift the decision boundary
significantly compared to \orig. In contrast, approaches that target \tprb and
\tnrb do not need a significant boundary shift as \orig's performance on these
metrics is already high. The post-processing approaches, \hardtt and \pleisst,
appear to be outliers in this observation, but as we discuss later, their
accuracy drop is indicative of the poor correctness-fairness balance that is
typical in post-processing. In the other two datasets, \orig does not display
such differences across these fairness metrics, and we do not observe
significant differences in the accuracy performance of fair approaches that
target demographic parity vs equalized odds.

\vspace{2mm}
\noindent
\fbox{
\parbox{0.96\columnwidth}{
\emph{Key takeaway:} 
  Fair approaches generally trade accuracy for fairness. The compromise in
  accuracy is bigger when fairness-unaware approaches achieve low fairness wrt
  the fairness metric that a fair approach optimizes for, relative to other
  metrics. The tradeoff is less interpretable for correctness metrics other
  than accuracy, as classifiers typically do not optimize for them.
}}

\begin{figure*}[t]
  \centering
  \hspace{5mm}\includegraphics[width=0.8\textwidth, height=0.03\textwidth]{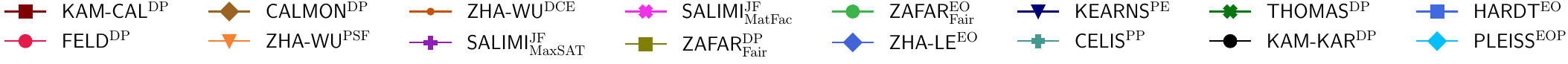}\vspace{-1mm}
  \begin{subfigure}[b]{0.188\textwidth}
    \includegraphics[width=1\textwidth]{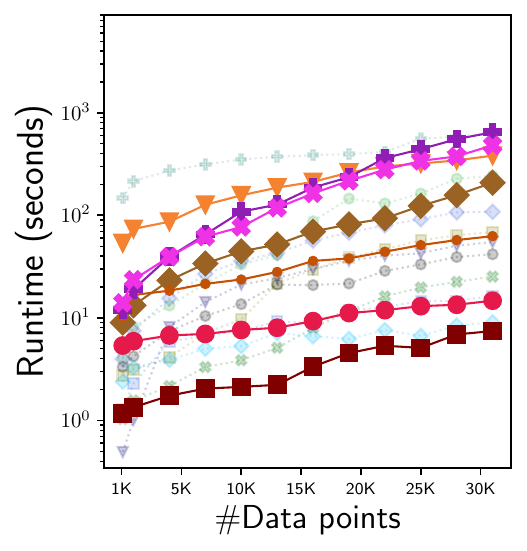}
	\vspace{-6mm}
    \caption{pre-processing}
    \label{fig:run_prep} 
  \end{subfigure} \hspace{-1em}
  \hspace{0.5mm}
  \begin{subfigure}[b]{0.16\textwidth}
    \includegraphics[width=1\textwidth]{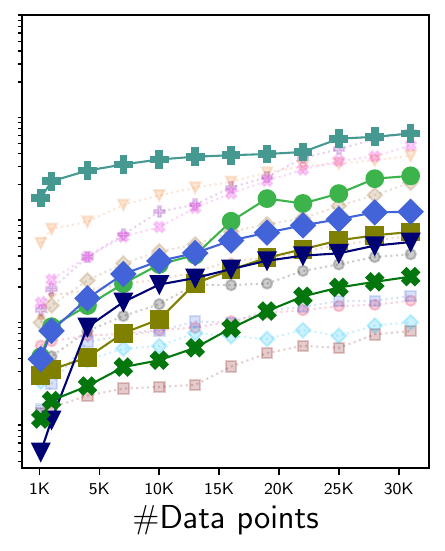}
	\vspace{-6mm}
    \caption{in-processing}
    \label{fig:run_in}
  \end{subfigure} \hspace{-1em}
  \hspace{0.5mm}
  \begin{subfigure}[b]{0.16\textwidth}
    \includegraphics[width=1\textwidth]{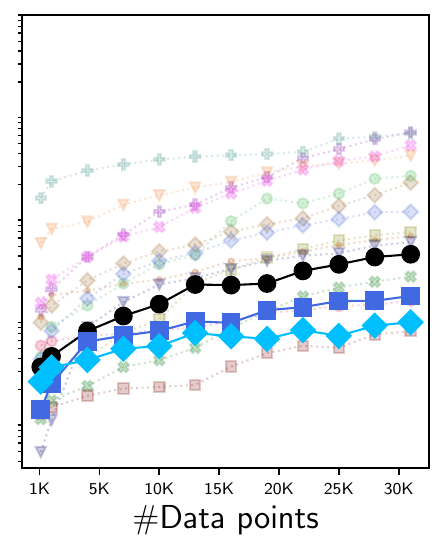}
	\vspace{-6mm}
    \caption{post-processing}
    \label{fig:run_post}
  \end{subfigure}
  \hspace{0.5mm}
  \begin{subfigure}[b]{0.16\textwidth}
    \includegraphics[width=1\textwidth]{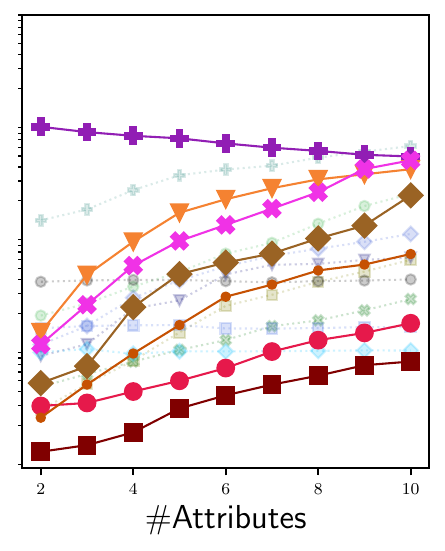}
    \vspace{-6mm}
	\caption{pre-processing}
    \label{fig:scale_prep} 
  \end{subfigure} \hspace{-1em}
  \hspace{0.5mm}
  \begin{subfigure}[b]{0.16\textwidth}
    \includegraphics[width=1\textwidth]{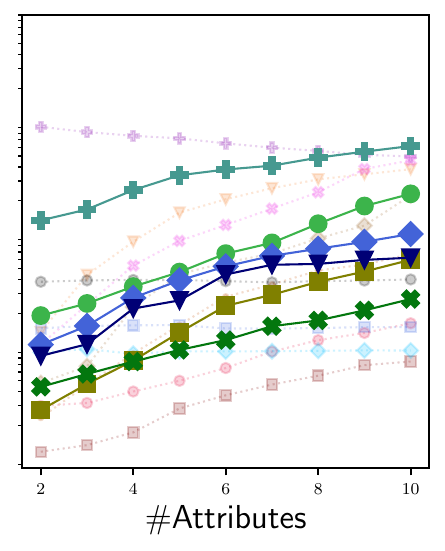}
	\vspace{-6mm}
	\caption{in-processing}
    \label{fig:scale_in}
  \end{subfigure} \hspace{-1em}
  \hspace{0.5mm}
  \begin{subfigure}[b]{0.16\textwidth}
    \includegraphics[width=1\textwidth]{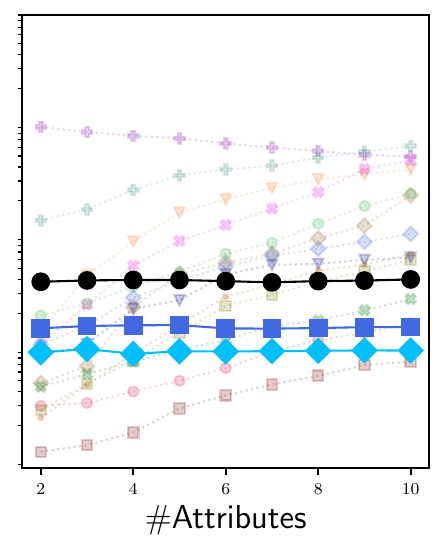}
	\vspace{-6mm}
    \caption{post-processing}
    \label{fig:scale_post}
  \end{subfigure}
  \vspace{-3mm}
  \caption{Results of efficiency and scalability experiments on the fair
  approaches. (a) -- (c) show runtime overhead with varying data size and (d) -- (f)
  show runtime overhead with varying number of attributes in Adult dataset. Note that the
  y-axis is in log scale.}
  \label{fig:runtime_experiment} 
  \vspace{-2mm}
\end{figure*}

\vspace{1mm}

\noindent\emph{There is no single winner.} All approaches succeed in improving
fairness wrt the metric (and notion) they target. However, they cannot guarantee
fairness wrt other notions: their performance wrt those notions is generally
unpredictable. This is in line with the impossibility theorem, which states that
enforcing multiple notions of fairness is impossible in the general
case~\cite{chouldechova}. While we observe that approaches frequently improve on
fairness metrics they do not explicitly target, this can depend on the dataset
and on correlations across metrics. No approach achieves perfect fairness across
all metrics. \thomaseot comes close in the German dataset, but this dataset
contains low gender bias as even \orig achieves reasonable fairness scores on
all metrics, especially compared to Adult and COMPAS. Further, many techniques
exhibit ``reverse'' discrimination (the red stripes indicate discrimination
against the privileged group), but these effects are generally small (a high
striped bar indicates high fairness, and, thus, low discrimination in the
opposite direction).\label{pg:r4o2}

\smallskip
\noindent
\fbox{
\parbox{0.96\columnwidth}{
\emph{Key takeaway:} 
  Approaches improve fairness on the metric they target, but their performance
  on other metrics is unpredictable.
}}
\smallskip

\noindent\emph{Causal fairness metrics explain some of the apparent
discrimination.} We noted a significant difference in the proportions of \te
that transmit through the direct and indirect paths on Adult (detailed in
\appOrTechRep), which signifies that the causal influence of gender on outcome
mostly goes through indirect paths. Specifically, attributes such as education,
occupation, working hours/week, etc.\ mediate this causal influence and
partially explain why there is an income gap between genders. We observe that
all causal approaches, \zhangpcft, \zhangdcet, and \salt, consistently improve
the fairness scores in \te over all datasets. In contrast, non-causal approaches
behave unpredictably and often decrease the scores in \te, particularly the
component that controls direct (and discriminatory) causal influence of the
sensitive attribute.

\medskip
\noindent
\fbox{
\parbox{0.96\columnwidth}{
\emph{Key takeaway:} 
  Reasoning about the causal structure is important, as it provides useful
  clues in understanding and explaining discrimination. Non-causal approaches
  establish statistical balance at the cost of exacerbating causal biases.
  We are not arguing that \te alone can resolve
  biases; arguably, the fact that women earn less due to their education and
  occupation may in itself be a bias we want to eliminate. More fine-grained causal 
  notions are needed to capture the nuances of fairness in a particular setting.
}}

\smallskip
\noindent \emph{Post-processing approaches tend to violate individual level
fairness}. We note that the fairness scores in \id are generally lower for
post-processing than pre- and in-processing. This is because post-processing
operates on less information than pre- and in-processing and does not assume
knowledge of the attributes in the training data. Thus, it does not take
similarity of individuals into account and tends to produce different outcomes
based on the sensitive attribute. However, some pre- and in-processing
approaches---e.g., \feldt, \zdi, and \zdmt---trivially satisfy \id by
discarding the sensitive attribute while training, even though they do not
target individual fairness. This indicates that \id is too rigid to fully
capture individual discrimination as it only compares identical (except for the
sensitive attribute) rather than similar individuals.

\vspace{1mm}
\noindent
\fbox{
\parbox{0.96\columnwidth}{
\emph{Key takeaway:} 
  Post-processing approaches can significantly violate individual level
  fairness. This is an inherent limitation of post processing, as it has no
  knowledge of the attributes in the training data and cannot take individual
  similarity into account. However, \id is too rigid in practice and higher
  fairness scores in \id among pre- and in-processing approaches do not
  necessarily translate to higher individual fairness.
}}
\vspace{1mm}

\noindent\emph{Pre- and in-processing achieve better correctness-fairness
balance than post-processing}. Post-processing operates at a late stage of the
learning process and does not have access to all of the data attributes by design.
As a result, it has less flexibility than pre- and in-processing. Given the fact
that post-hoc correction of predictions are sub-optimal with finite training
data~\cite{wood2017}, post-processing approaches typically achieve inferior
correctness-fairness balance compared to other approaches. In all the datasets,
post-processing achieves on average 2-5\% lower accuracy compared to pre- and
in-processing that target the same fairness metrics. There is no significant
difference in performance among the pre- and in-processing approaches. However,
we note that the correctness-fairness balance of pre-processing approaches
varies depending on the downstream ML model (Section~\ref{sec:resilience}), and,
thus, we cannot conclude if pre-processing is always comparable in performance
with in-processing.

\vspace{1mm}
\noindent
\fbox{
\parbox{0.96\columnwidth}{
\emph{Key takeaway:}
  Pre- and in-processing achieve better correctness and fairness compared to
  post-processing. The performance of pre- and in-processing approaches is not
  always comparable as the former varies depending on the choice of ML model. 
}}

\begin{figure*}[t]
  \centering
  {\includegraphics[width=0.99\textwidth, keepaspectratio]{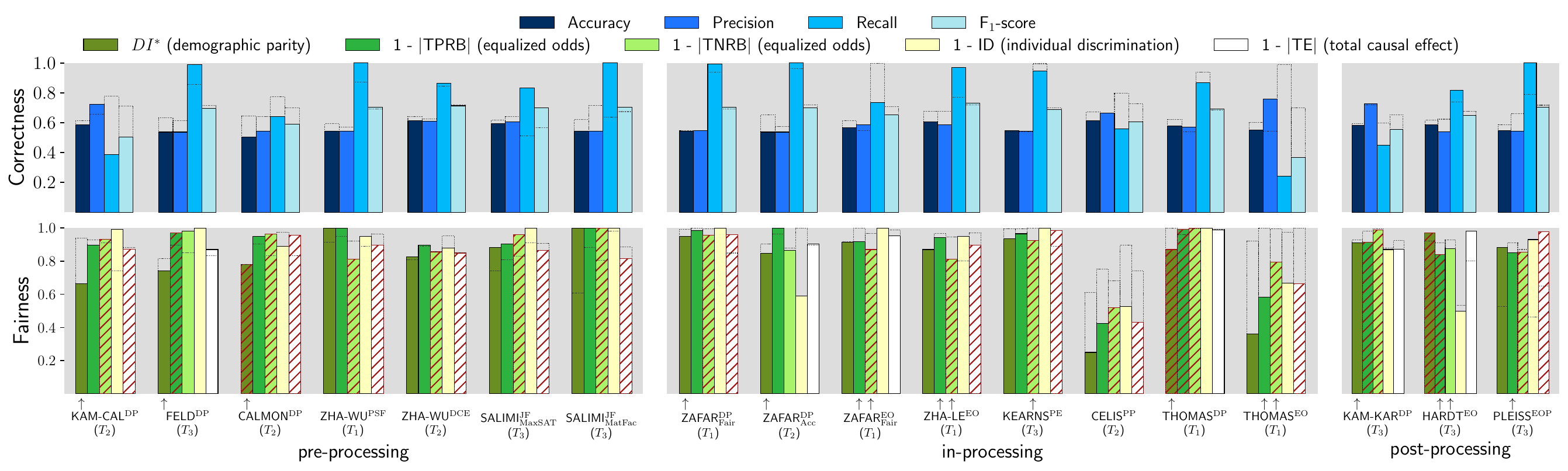}} 	
  \vspace{-4mm} 
  \caption{We examine the robustness of fair approaches to data errors.
  Higher scores for correctness (fairness) metrics correspond to more correct
  (fair) outcomes. The bars highlighted in red denote the reverse direction of
  the remaining discrimination---favoring the unprivileged group more than the
  privileged group. The arrows ($\uparrow$) denote the fairness metric(s) each
  approach is optimized for. The bar plots for each approach on the error-free
  dataset are overlaid for aiding visual comparison.} 
  \vspace{-3mm}
  \label{fig:robust_compas}
\end{figure*}

\begin{figure*}[t]
  \centering
  \includegraphics[ width=1\textwidth]{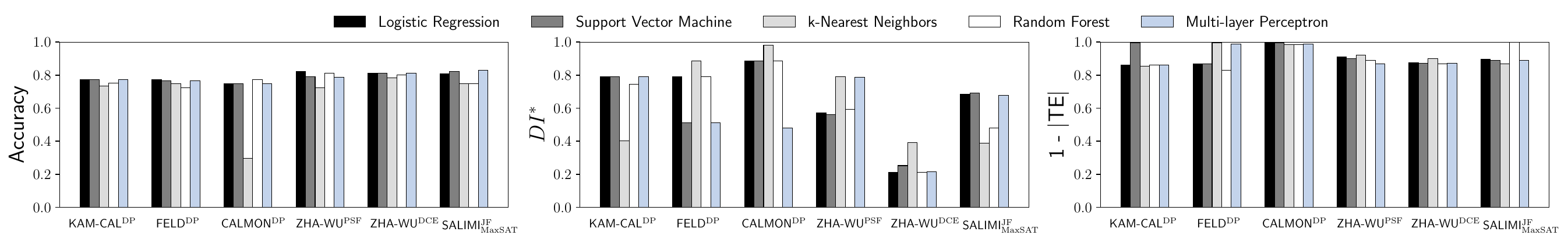} 
  \vspace{-8mm}
  \caption{The sensitivity of pre-processing approaches to the choice of ML
    model in terms of accuracy, DI$^*$, and TE on the Adult dataset. 
    }
  \label{fig:resilience} 
  \vspace{-1mm}
\end{figure*}

\vspace{-2mm}
\subsection{Efficiency and Scalability} \label{subsec:runtime} 

We now study the runtime behavior of all approaches, to investigate their
efficiency gap and highlight the need for scalability considerations. While some
approaches can benefit from optimizations, such as the use of GPUs, producing
these optimizations is beyond our scope. We do not present separate variants of
the same approach unless they differ significantly in behavior. We compute the
total runtime of each approach as pre-processing time (if any) + training time +
post-processing time (if any). We subtract from all methods the runtime of
\orig, so that what we report is the overhead each approach introduces over the
fairness-unaware method.

Our first experiment investigates the efficiency and scalability of the
approaches as the number of data points increases. We used the Adult dataset,
as it contains the highest number of data points, and executed new instances of
each approach with different numbers of data points (from 0.1K to 31K) sampled
from the dataset. Our second experiment explores the runtime behavior of the
approaches as the number of attributes increases. We used the Adult dataset
here as well, as it contains the highest number of attributes. We executed new
instances of each approach with different number of attributes (from 2 to 10).
We present the results in Figure~\ref{fig:runtime_experiment}.

\noindent\emph{Post-processing approaches are generally most efficient and
scalable}. Post-processing approaches tend to be very efficient, as their
mechanisms are less complex compared to pre- and in-processing approaches. As a
result, they scale well wrt increasing data sizes and they are not affected by
increase in the number of attributes. A few pre- and in-processing techniques
like \kamt and \thomasdpt do perform better than post-processing, but this does
not hold for most other techniques in their categories.

\vspace{1mm}
\noindent
\fbox{
\parbox{0.96\columnwidth}{
\emph{Key takeaway:} 
  Post-processing approaches are more efficient and scalable than pre- and
  in-processing approaches. Pre- and in-processing approaches generally incur
  higher runtimes, which depend on their computational complexities.
}}
\smallskip

\noindent\emph{Causal computations incur sharper runtime penalties}. An
important observation from \Cref{fig:run_prep} is that causal mechanisms---such
as \zhangpcft, \zhangdcet, and \salt---incur significantly higher runtimes
compared to other pre-processing approaches. In fact, both variations of \salt
are NP-hard in nature. Simply, discovering causal associations from data is more
complex than non-causal associations. \calmt also demonstrates high runtimes, in
its case due to relying on solving convex optimization problems, and very poor
scalability with increasing attributes (Figure~\ref{fig:runtime_experiment}(d)).


\noindent
\fbox{
\parbox{0.96\columnwidth}{
\emph{Key takeaway:} 
  Causality-based mechanisms incur higher runtimes. Other complex mechanisms also
  lead to efficiency and scalability challenges.
}}
\vspace{0.7mm}

\noindent\emph{Pre-processing scales better with increasing data sizes than with
increasing number of attributes.} We note a clear separation between the
inherently more complex pre-processing methods (\zhangpcft, \zhangdcet, \salt,
and \calmt) and the rest (\kamt and \feldt). In fact, \kamt and \feldt perform
on par with or better than post-processing in terms of efficiency, and generally
better than most in-processing approaches. Generally, pre-processing
demonstrates more robust scaling behavior wrt data size than the number of
attributes. In fact, the runtime of several pre-processing approaches appears to
grow exponentially with the number of attributes (\Cref{fig:scale_prep}).
Causality-based approaches display similar challenges. The behavior of \salmst
is of note: in contrast with other techniques, its performance improves as the
number of attributes grows. This is because the number of constraints in \salmst
increases rapidly with fewer attributes, resulting in higher runtimes in those
settings.

\vspace{1mm}
\noindent\emph{In-processing approaches are more affected by the data size than
by the number of attributes, but the difference is less distinct than
pre-processing.} In-processing techniques show a slightly sharper rise in
runtime when the data size increases compared to pre-processing approaches
(\Cref{fig:run_in}) and scale more gracefully than pre-processing ones with the
number of attributes. Their runtime does increase, since the higher number of
attributes increases the complexity of the decision boundary in optimization
problems, but it is generally lower than pre-processing, which typically
performs data modification on a per-attribute basis.

\smallskip
\noindent
\fbox{
\parbox{0.96\columnwidth}{
\emph{Key takeaway:} 
  Pre-processing approaches are generally more affected by the number of
  attributes than the data size. In-processing approaches appear to scale better
  with the number of attributes than with the data size, but this distinction
  is less clear than pre-processing.
}}

%

\subsection{Robustness to Data Errors}\label{sec:robustness} Fair ML approaches
typically assume (explicitly or implicitly) that training and testing data are
drawn from the same target distribution; thus, they can only address
discrimination that is reflected in the data generative process. However,
training data is susceptible to data quality issues such as selection bias,
misclassification, technical errors, etc., which are introduced during data
collection and preparation, and distort the underlying distribution in a way
that data no longer represents the target
population~\cite{schelter2018,schelter2021}. Furthermore, data quality issues
are highly correlated with sensitive attributes in many domains like healthcare
and immigration~\cite{chen2020, nobles2000}. For example, African-American
patients are more likely to be seen in clinics where documentation is less
accurate or systematically different than other higher-end healthcare
services~\cite{francesco2018}. 

In this section, we investigate the robustness of fair ML approaches to data
quality issues. For this experiment, we injected COMPAS with various
combinations of common data errors; we present our findings on three training
datasets that contain the following errors: ($T_1$)~swapped values between
\texttt{Prior\_convictions} and \texttt{Age}; ($T_2$)~scaled values of
\texttt{Prior\_convictions} and noisy values of \texttt{Age}; ($T_3$)~missing
values of \texttt{Race} and \texttt{Risk\_of\_recidivism} that are imputed using
standard Scikit-learn imputers. All errors were randomly and disproportionately
introduced, affecting 50\% of African-Americans and 10\% of other races. The
main purpose of our experiments is to highlight situations where classifiers may
perform unexpectedly, not to exhaustively evaluate over all possible scenarios.
We present the results in Figure~\ref{fig:robust_compas}; for each approach, we
only report our findings on the set that most affected the correctness-fairness
balance and refer to \appOrTechRep for full results.

\emph{Post-processing approaches are more robust against data errors than pre-
and in-processing}. Post-processing is designed to manipulate the predictions of
a learned classifier and does not access the data attributes. Hence, our
experiments with $T_1$ and $T_2$ did not significantly affect the fairness
scores of post-processing approaches. We find that post-processing approaches
are most affected when trained on $T_3$, as they rely on the sensitive attribute
and labels in the training data. We notice 2-5\% drop in accuracy and
F$_1$-score, and 5-10\% decrease in the fairness metrics the approaches
optimize. 

Pre- and in-processing are affected by all types of data errors. In most cases,
we see a sharp decline in accuracy ranging from 5 to 10\%. \kamt, \zdifairt,
and \kearnst are exceptions: they usually incur high accuracy penalties for
enforcing fairness (Figure~\ref{fig:experiment_compas}) and errors only further
reduce accuracy by 2--4\%. We note an interesting distinction between
approaches that enforce demography- and error-aware fairness notions.
Approaches targeting demography-aware notions cope better and their target
fairness scores are typically within 5\% of what they achieve in the absence of
errors. These approaches repair the training data (or constrain the classifier)
to meet some target demography and we hypothesize that their robustness is due
to the fact that the target demography holds regardless of data errors. Thus,
the drop in fairness is less severe even though accuracy is affected by corrupt
data. In contrast, approaches enforcing error-aware notions are more severely
impacted and we observe drops in their target fairness metric ranging from 8 to
20\%. These approaches equalize error rates between the sensitive groups and
heavily depend on the correctness of predictions. For instance, we observe that
\calmt and \kamdpt pay the least penalty in their target fairness metric even
when presented with corrupt data, while \zhangt, \kearnst, and \thomaseot all
report significant drops. Finally, the changes are unpredictable for the
metrics not optimized by each approach.

\smallskip
\noindent
\fbox{
\parbox{0.96\columnwidth}{
\emph{Key takeaway:}
  Pre- and in-processing exhibit poor generalizability in the presence of data
  quality issues in the training data and fail to build models that are fair on the
  target population. Post-processing is more robust by design.
}}

\subsection{Sensitivity to the Underlying ML Model}\label{sec:resilience} All
pre- and post-processing approaches need to be combined with a classifier to
complete the ML pipeline. In this section, we study the sensitivity of pre- and
post-processing approaches to the choice of ML model used for classification. We
executed a new instance of each approach on the Adult dataset after pairing them
with each of the following models: Logistic Regression (LR), Support Vector
Machine (SVM), Random Forest (RF), k-Nearest Neighbors (k-NN), and Multi-layer
Perceptron (MLP). We implemented each classifier using Scikit-learn (version
0.22.1) and chose hyper-parameters that maximize correctness in the
fairness-unaware setting (detailed in \appOrTechRep).
Figure~\ref{fig:resilience} presents our results on the pre-processing
approaches; we detail the rest in \appOrTechRep.

\begin{sloppypar}
\emph{The choice of model affects pre-processing approaches, while
post-processing ones are generally less impacted}. By design, post-processing
approaches do not make any assumptions about the classifier that produced the
predictions. Our experiments showed that their accuracy and fairness only vary
slightly across different models, likely due to variation in prediction
probabilities generated by each classifier. In contrast, the
correctness-fairness balance of pre-processing approaches varies significantly
with the choice of downstream ML model. This indicates that off-the-shelf
classifier models are not always suitable for pre-processing, and
hyper-parameter settings should be specific to the repaired data produced by
each approach.
\end{sloppypar}

\vspace{1mm}
\noindent
\fbox{
\parbox{0.96\columnwidth}{
\emph{Key takeaway:}
  Pre-processing is sensitive to the choice to ML model; the approaches require the
  hyper-parameters to be tuned separately per classifier model and in accordance
  to the repaired data. In contrast, post-processing is resilient to the choice of
  ML model and behaves similarly regardless of the model.  
}}

\vspace{-2mm}
\subsection{Other Results}\label{sec:others} In our evaluation, we further
explored \emph{data efficiency} and \emph{stability} of all the approaches. Due
to space limitations, we summarize the results here, and refer the reader to
\appOrTechRep for a full analysis. Our findings suggest that most approaches are
data-efficient, and the size of the training set does not impact their accuracy
and fairness significantly. Further, we find that approaches show low variance
over different choices of training sets, with only a small number of outliers. 

\vspace{-2mm}

\section{Lessons and Discussion}\label{sec:discussion}
%

The goal of our work has been to bring some clarity to the vast and diverse
landscape of fair classification research. Work on this topic has spanned
multiple disciplines with different priorities and focus, resulting in a wide
range of approaches and diverging evaluation goals. Data management research
has started making important contributions to this area, and we believe that
there are a lot of opportunities for impact and synergy. Through our
evaluation, we aimed in particular to identify areas and opportunities where
data management contributions appear better-suited to be successful. We discuss
these general guidelines here.

\vspace{1mm}
\noindent\emph{Pre-processing approaches are a natural fit but exhibit
scalability challenges.} Data management research has primarily focused on the
pre-processing stage, as data manipulations create a natural fit. However, our
evaluation shows that pre-processing tends to not scale robustly with the number
of attributes. Research in pre-processing methods should be mindful of problem
settings where the high data dimensionality may lead to a poor fit. This
observation also points to an opportunity that plays squarely into the strengths
of the data management community, as efforts can focus on attacking this
scalability challenge. Some contributions already exist in this direction (e.g.,
\salt has a parallel implementation), and improvements are likely to lead to
more impact. Notably, causality-based approaches produce sophisticated repairs,
but impose a significant runtime penalty. \kamt and \feldt use simpler repairs,
resulting in orders of magnitude better runtime performance, but tend to produce
poorer fairness wrt the causal metrics.\label{pg:r1o3}

\vspace{1mm}
\noindent 
\emph{Synergy with data cleaning and repairs.}
Our evaluation highlights the impact of data quality issues on the performance
of pre- and in-processing techniques. Considerations of data quality are a
particularly good fit for pre-processing methods, as they already focus on data
repairs. Investigating repairs that combine both cleaning and fairness
objectives has the potential to lead to increased robustness, which may give
pre-processing approaches an edge against in-processing in practical settings.

\vspace{1mm}
\noindent 
\emph{Synergy with ML research.} \looseness-1 Our analysis notes that some
in-processing techniques scale poorly with increasing data size compared to
pre-processing approaches. Generally, runtime performance is often overlooked in
ML research, and data management contributions can likely have
impact in improving in-processing approaches in that regard. Further,
pre-processing approaches vary in performance depending on the downstream ML
model. These approaches have the potential to improve their resilience, and
further investigation can explain on how to best pair these approaches with ML
models. 

\vspace{1mm}
\noindent
\label{pg:r4o1} \emph{Applicability of fairness notions and
approaches.} Due to the variety of notions and approaches in literature, the
task of choosing the most suitable fair classification approach can be daunting.
As we saw in our evaluation, performance of different approaches as measured by
different metrics can diverge, and it is important to follow the application
requirements before attacking a problem setting with a particular method. It is
similarly important to consider what fairness notions capture the nuances of and
context required by the specific application.

Non-causal notions typically present the fewest computational challenges, and
can be enforced efficiently. However, they aim at statistical balance, often at
the cost of exacerbating causal biases.  Causal notions provide stronger
guarantees and are generally a good fit when adequate domain knowledge and
computational resources are available. Enforcing multiple notions is not
typically recommended, as prior literature has proved that different fairness
constraints cannot be satisfied simultaneously and combining several constraints
leads to a vacuous classifier~\cite{kleinberg,narayanan}.
\label{pg:r1o2_2}

There are also considerable tradeoffs across the different stages of fairness
enforcing mechanisms. Pre-processing presents the flexibility of being model
agnostic, but there can be practical constraints to modifying training data as
this may violate anti-discrimination laws~\cite{barocas2016}. Additionally,
pre-processing repairs data on the assumption that model predictions will follow
the ground truth. However, it cannot enforce fairness notions that balance the
correctness of predictions across sensitive groups, as it cannot make
assumptions on the correctness of predictions before model training. This means
that notions such as equalized odds and predictive parity cannot be easily
handled in the pre-processing stage. Our findings also suggest that
pre-processing can pose scalability challenges with high dimensional data, and
can vary in performance if the downstream model is not fixed. In contrast,
in-processing directly modifies the learning objective, enforces a wider variety
of notions, and provides better fairness guarantees. However, it is
model-specific and works under the assumption that the model is replaceable,
which may not be practically feasible. Similar to pre-processing, in-processing
also encounters scalability issues in our experiments and their fairness
guarantees may not hold if the training data contains errors. On the other hand,
post-processing works on top of a trained classifier, which generally makes it
more efficient and robust than pre- and in-processing. However, it often
achieves poorer correctness-fairness balance, a critical component in any
application. Lastly, combining multiple approaches is possible, but faces
practical hurdles such as substantial penalties in correctness, runtime
overhead, and required access to the entire ML pipeline. \label{pg:r4o3}

We hope that our analysis will be helpful to outline useful perspectives and
directions to data management research in fair classification. To the best of
our knowledge, ours is the broadest evaluation and analysis of work in this
area, and can contribute to a useful roadmap for the research community.

\small{\noindent\textbf{Acknowledgements:} This work was supported by the
NSF under grants CCF-1763423, IIS-1943971, and 2112606. We thank Sainyam
Galhotra for useful discussions.}

\newpage
\bibliographystyle{ACM-Reference-Format}
\bibliography{paper}

\ifTechRep
\appendix
\section*{Appendix}
%

\section{Description of Fairness Metrics} 
In this section, we provide detailed discussion of the fairness metrics in
Figure~\ref{fig:fairness} that we choose for evaluating the fair approaches. We
begin with the non-causal metrics and then continue to the causal ones in order
to best highlight their rationale and differences. 

%

\subsection{Non-causal Fairness Metrics} The non-causal metrics depend entirely
on empirical data and aim to establish statistical relationships between the
sensitive attribute and the predictions. We start with an example that
highlights two common types of discrimination typically determined from
empirical data and proceed to describe how our chosen metrics operate.

\begin{example} \label{ex:stat} 
\looseness-1 Consider a model of university admissions that aims to offer
admission to highly-qualified students. The admissions committee automates the
admission process by training a binary classifier over historical admissions
data. Female students are historically underrepresented at this university,
making up 40\% of the student body; so, we designate males as the privileged
group ($S = 1$), and females as the unprivileged group ($S = 0$). After
training, the classifier achieves $87\%$ accuracy and $78\%$ F$_1$-score over
the training data. Figure~\ref{fig:sample} summarizes the prediction-related
statistics for both groups. Although the classifier is satisfactory in terms of
correctness, it is not fair across gender. Specifically, we observe two ways
females are being discriminated:

\begin{itemize}
    \item (\discone) The fraction of females predicted as highly-qualified
    (positive) is $\frac{7+2}{40} \approx$ 23\%, which is significantly lower
    than the fraction of males predicted as highly-qualified ($\frac{14+6}{60}
    \approx $ 33\%). This highlights how a group can receive an unfair advantage
    (or disadvantage) if the proportion of positive and negative predictions
    differs across groups.

    \item (\disctwo) The true positive rate for females is
    $\frac{\mathit{TP}}{\mathit{TP}+\mathit{FN}} = \frac{7}{7+3}= 70\%$, which
    is significantly lower than that of the males
    ($\frac{\mathit{TP}}{\mathit{TP}+\mathit{FN}} = \frac{14}{14+2} \approx
    88\%$). This indicates how predictions can disadvantage a group if the
    correctness of predictions differs across groups.
\end{itemize}
\end{example}

\begin{figure}
	\centering
	{\includegraphics[width=0.43\textwidth, keepaspectratio]{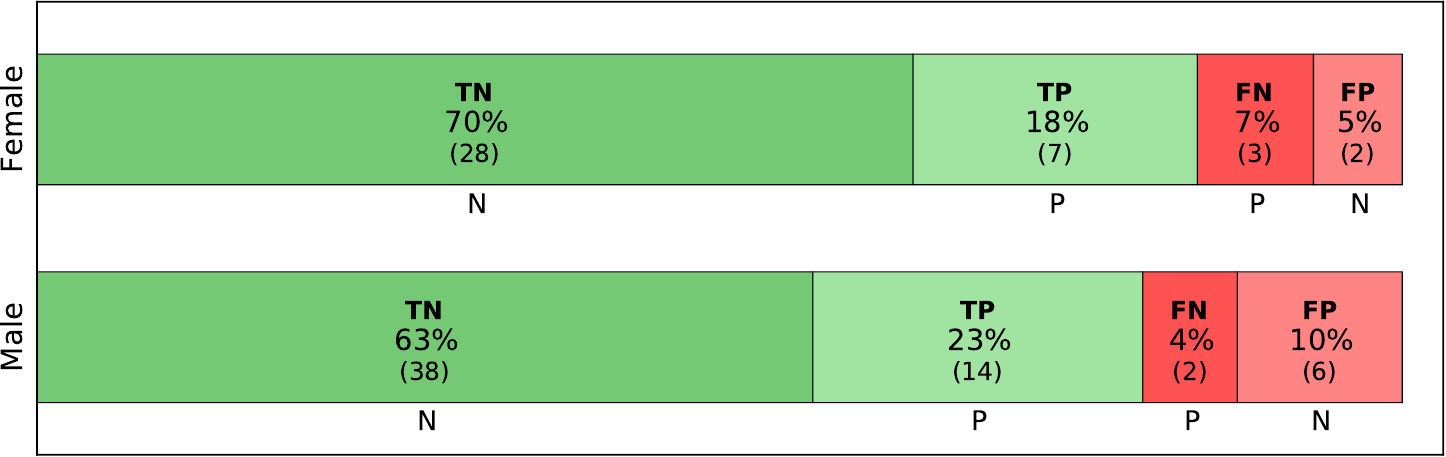}} 
	 \caption{Prediction statistics over 100 applicants,
	 grouped by gender: 60 male (bottom) and 40 female
	 (top). The ground truth (positives as $P$ and negatives as $N$) is
	 indicated below each segment.}
	\label{fig:sample}
\end{figure}

\smallskip\noindent\textbf{Disparate Impact ($\mathbf{DI}$)} is a group,
non-causal, and observation level metric. It quantifies demographic
parity~\cite{dwork}, a fairness notion that states that positive predictions
should be independent of the sensitive attribute. To measure demographic parity,
$\mathit{DI}$ computes the ratio of empirical probabilities of receiving
positive predictions between the unprivileged and the privileged groups.

\begin{align*}
    \mathit{DI} = \frac{ Pr(\hat{Y} = 1 \mid S = \prot)}{ Pr(\hat{Y} = 1 \mid S = \priv)}\\
\end{align*}

\looseness-1 $\mathit{DI}$ lies in the range $[0, \infty)$. $\mathit{DI} = 1$
denotes perfect demographic parity. $\mathit{DI} < 1$ indicates that the
classifier favors the privileged group and $\mathit{DI} > 1$ means the opposite.
In Example~\ref{ex:stat}, $\mathit{DI}$ $= \frac{9/40}{20/60} = 0.67$, which
suggests that positive predictions are not independent of gender as males have
higher probability to receive positive predictions than females. This is
indicative of \discone: the fraction of females being granted admission is much
lower than males.

\smallskip\noindent\textbf{True Positive Rate Balance ($\mathbf{TPRB}$) and True
Negative Rate Balance ($\mathbf{TNRB}$)} are two group, non-causal, and
observation level metrics. They measure discrimination as the difference in
$\mathit{TPR}$ and $\mathit{TNR}$, respectively, between the privileged and
unprivileged groups.

\begin{align*}
\mathit{TPRB} = Pr(\hat{Y} {=} 1 \mid Y {=} 1, S {=} \priv) - Pr(\hat{Y} {=} 1 \mid Y {=} 1, S {=} \prot)
\\
\mathit{TNRB} = Pr(\hat{Y} {=} 0 \mid Y {=} 0, S {=} \priv) - Pr(\hat{Y} {=} 0 \mid Y {=} 0, S {=} \prot) \\
\end{align*}

Both $\mathit{TPRB}$ and $\mathit{TNRB}$ lie in the range $[-1, 1]$. These two
metrics, together, measure equalized odds~\cite{hardt}, which states that
prediction statistics (e.g., $\mathit{TPR}$ and $\mathit{TNR}$ ) should be
similar across the privileged and the unprivileged groups. Perfect equalized
odds is achieved when $\mathit{TPRB}$ and $\mathit{TNRB}$ are $0$, as the
classifier performs equally well for both groups. A positive value in either of
the two metrics indicates that the classifier tends to misclassify the
unprivileged group more. In Example~\ref{ex:stat}, $\mathit{TPRB}$ =
$\frac{14}{16} - \frac{7}{10} = 0.18$ and $\mathit{TNRB}$ = $\frac{38}{44} -
\frac{28}{30} = -0.07$. The high positive value of $\mathit{TPRB}$ indicates
\disctwo: the $\mathit{TPR}$ of females is much lower than males.

\smallskip\noindent\textbf{Individual Discrimination
($\mathbf{ID}$)~\cite{themis}} is an individual, non-causal, and observation
level metric. It allows us to determine both the classifier's discrimination
with respect to individuals and the influence of the sensitive attribute.
Specifically, $\mathit{ID}$ is the fraction of tuples for which, changing the
sensitive attribute causes a change in the prediction, compared to otherwise
identical data points. Suppose that $Q$ is the set of such tuples, defined as $Q
= \{a \in \D \mid \exists b : \X_a {=} \X_b \wedge S_a {\neq} S_b \wedge
\hat{Y_a} {\neq} \hat{Y_b}\}$; then $\mathit{ID}$ = $\frac{|Q|}{|\D|}$.
$\mathit{ID}$ lies in the range $[0, 1]$ and $\mathit{ID}$ $= 0$ corresponds to
perfect individual fairness as there exists no data point for which changing
sensitive attribute results in a different prediction.

\begin{example} 
Consider 12 university applicants shown in Figure~\ref{table:sampledata}. To
measure $\mathit{ID}$, we alter the sensitive attribute (\texttt{gender}) of
each tuple while keeping rest of the attributes intact, and re-evaluate the
classifier on the altered tuples. Suppose that the prediction for $t_7$ changes
from 0 to 1 when $t_7$'s \texttt{gender} is changed from \texttt{Female} to
\texttt{Male}, and that predictions do not change for any other tuples. Then,
$\mathit{ID}$ $= \frac{1}{12} = 0.08$, indicating that $8\%$ of the applicants
are discriminated because of their gender.
\end{example}

The formal definition of $\mathit{ID}$ requires computation over all possible
data points in the domain of attributes, but practical heuristics limit
interventions to smaller datasets of interest~\cite{themis}.

\setlength{\tabcolsep}{2.5pt}
\renewcommand{\arraystretch}{0}
    \begin{figure}[t]
    \centering\ra{1} {\small 
    {\small
    \begin{tabular}{@{}llllllcc@{}}
    \toprule
    &\multicolumn{2}{c}{$\X$} &&\multicolumn{1}{c}{$S$}
    &&\multicolumn{1}{c}{$\hat Y$} \\
    \cmidrule{2-3}  \cmidrule{5-5} \cmidrule{7-7} 
    id &\texttt{SAT}
    &\makecell[l]{\texttt{dept\_choice}}  
    &&\texttt{gender}
    &&\makecell[l]{\texttt{admitted}} \\ 
    \midrule 
    t$_1$       &High       &Physics      &&Male    &&1 \\
    t$_2$       &High       &Mathematics  &&Male    &&0  \\
    t$_3$       &Average    &Physics      &&Male    &&1  \\
    t$_4$       &High       &Mathematics  &&Male    &&1  \\
    t$_5$       &High       &Physics	  &&Male    &&1  \\
    t$_6$       &Average    &Mathematics  &&Male    &&0 \\
    t$_7$       &high       &Mathematics  &&Female  &&0  \\
    t$_8$       &Average    &Mathematics  &&Female  &&0 \\
    t$_9$       &high       &Mathematics  &&Female  &&1  \\
    t$_{10}$    &high       &Physics      &&Female  &&1  \\
    t$_{11}$    &Average    &Mathematics  &&Female  &&0  \\
    t$_{12}$    &Average    &Physics      &&Female  &&1  \\
    \bottomrule
    \end{tabular}}} 
    \vspace{-2mm}
    \caption{Sample data for 12 university applicants.} 
    \vspace{-2mm}
    \label{table:sampledata}
    \end{figure}

\begin{figure}
    \centering
    \vspace{3mm}
    {\includegraphics[width=0.35\textwidth, keepaspectratio]{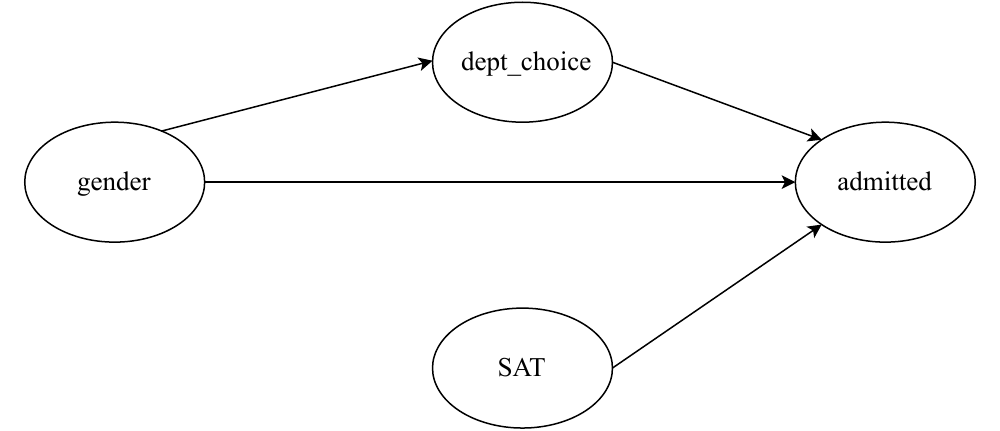}} 
    \vspace{-2mm}
        \caption{The graphical causal model corresponding to
        Example~\ref{ex:stat}.}
    \label{fig:dag}
    \vspace{-2mm}
\end{figure}

\subsection{Causal Fairness Metrics} 
Causal fairness metrics differ from non-causal metrics in that they consider
additional domain knowledge to reason about the underlying data generation
process and the way changes in attributes propagate to the prediction. We first
briefly discuss the structural causal model, the basic framework of causal
fairness. 

A structural causal model is a semantic framework that unifies the concepts of
causal graphs representing causal relationships among attributes and structural
equations denoting how each attribute is determined. The causal graph is
typically a directed acyclic graph, where vertices represent attributes and
edges represent functional relationships between these attributes. A structural
equation describes the process by which an attribute changes in response to
other attributes. An intervention on an attribute $X \in \X$, denoted by
$do(X=x)$, modifies the causal model by replacing the structural equations
associated with $X$ with a constant $x \in Dom(X)$. In the causal graph, this is
akin to removing all incoming edges into $X$ and replacing with $X \leftarrow
x$. The result of an intervention is a counterfactual world, where a different
probability distribution is induced over the other attributes. We use
$\hat{Y}_{X=x}$ to denote the potential outcome resulting from the intervention
and $Pr(\hat{Y}_{X=x} = y)$ as the counterfactual distribution of predictions.
These quantities can often be uniquely identified from empirical data, we refer
to prior literature for a detailed discussion~\cite{pearl}. Most causal fairness
metrics are defined in terms of interventions and counterfactuals; they can
account for confounding effects present in observational data and explain the
apparent discrimination using the intermediate attributes on the path from the
sensitive attribute to the prediction. 

\smallskip \noindent\textbf{Total Effect ($\mathbf{TE}$)~\cite{pearl}} is a
group, causal, and intervention level metric. It intervenes on the sensitive
attribute and measures discrimination as the effect of the intervention on the
prediction through all causal paths.
\begin{align*}
    \mathit{TE} {=} Pr(\hat{Y}_{S=\priv} = 1) - Pr(\hat{Y}_{S=\prot} = 1)\\
\end{align*}

$\mathit{TE} \in [-1, 1]$ and $\mathit{TE} = 0$ indicates complete fairness as
the causal effect of the sensitive attribute on the prediction is zero.  

\begin{example} 
    Suppose, the causal graph shown in Figure~\ref{fig:dag} represents the data
    generation process for Figure~\ref{table:sampledata}. Since the sensitive
    attribute (\texttt{gender}) and the prediction (\texttt{admitted}) do not
    share any confounders (i.e., common causes), we can easily compute
    $\mathit{TE}$ from observational data~\cite{pearl}. $\mathit{TE} {=}
    Pr(\hat{Y}_{S=\priv} = 1) - Pr(\hat{Y}_{S=\prot} = 1)$ $= Pr(\hat{Y} \mid
    S=\priv) - Pr(\hat{Y} \mid S=\prot)$ $= \frac{4}{6} - \frac{3}{6}$ $=
    0.16\%$. This confirms that the sensitive attribute causally influences the
    prediction and men are overall more likely to receive to a favorable
    outcome.
\end{example} 

\smallskip \noindent\textbf{Natural Direct Effect ($\mathbf{NDE}$)~\cite{pearl}}
is a group, causal, and intervention level metric. It computes discrimination as
the expected change in prediction when the sensitive attribute changes from
unprivileged to privileged, while setting other attributes to whatever value
they would have attained for unprivileged individuals. Semantically, \nde
measures the portion of \te that is transmitted through the direct path $S
\rightarrow \hat{Y}$. Suppose, $Z \in \X$ is the set of attributes on the
indirect paths from $S$ to $\hat{Y}$.

\begin{align*}
    \mathit{NDE} = Pr(\hat{Y}_{S=\priv, Z_{S=\prot}} = 1) - Pr(\hat{Y}_{S=\prot} = 1)  \\
\end{align*}

$\mathit{NDE} \in [-1, 1]$ and $\mathit{NDE} = 0$ indicates that there is no
direct causal effect of the sensitive attribute on the prediction.

\begin{example} 
    \begin{sloppypar}
    From Figure~\ref{fig:dag}, the direct path corresponds to
    \texttt{gender}$\rightarrow$\texttt{admitted} and the indirect path is
    denoted by \texttt{gender}$\rightarrow$ \texttt{dept\_choice}
    $\rightarrow$\texttt{admitted}. Using Theorem 4 of Zhang et
    al.~\cite{zhang2017}, we can compute $\mathit{NDE}$ as $\underset{s,
    d}{\sum} \Big\{Pr(\hat{Y} = 1 \mid S = \priv, SAT=s, dept\_choice=d)
    Pr(dept\_choice = d\mid S = \prot) Pr(SAT=s)\Big\} - Pr(\hat{Y} = 1 \mid S =
    \prot)$, $\forall s \in \Dom(SAT), \forall d \in \Dom(dept\_choice)$. Thus,
    $\mathit{NDE}$ = $(1 \cdot \frac{1}{2}\cdot\frac{4}{6}\cdot\frac{7}{12} +
    1\cdot\frac{2}{6}\cdot\frac{7}{12} + 1*\cdot\frac{2}{6}\cdot\frac{5}{12}) -
    \frac{3}{6} = 0.01$. This indicates that the causal influence of the
    sensitive attribute is minimal through the direct path and there is no
    significant discrimination. 
    \end{sloppypar} 
    
\end{example}

\smallskip \noindent\textbf{Natural Indirect Effect
($\mathbf{NIE}$)~\cite{pearl}} is a group, causal, and intervention level
metric. It computes discrimination as the expected change in prediction when the
sensitive attribute is unprivileged, while other attributes change to values
they would have attained for privileged individuals. Semantically, \nie measures
the portion of \te that is transmitted through the indirect paths $S \cdots
\rightarrow Z \rightarrow \cdots \hat{Y}$.
\begin{align*}
    \mathit{NIE} = Pr(\hat{Y}_{S=\prot, Z_{S=\priv}} = 1) - Pr(\hat{Y}_{S=\prot} = 1) \\ 
\end{align*}

$\mathit{NIE} \in [-1, 1]$ and $\mathit{NIE} = 0$ indicates that no causal
effect is transmitted through indirect paths from the sensitive attribute to the
prediction.  

\begin{example} 
    Suppose, the Physics department has a low acceptance rate and females tend
    to apply there more. Using Theorem 5 of Zhang et al.~\cite{zhang2017}, we
    can compute $\mathit{NIE}$ as $\underset{s, d}{\sum} \Big\{Pr(\hat{Y} = 1
    \mid S = \prot, SAT=s, dept\_choice=d) Pr(dept\_choice = d\mid S = \priv)
    Pr(SAT=s)\Big\} - Pr(\hat{Y} = 1 \mid S = \prot)$, $\forall s \in \Dom(SAT),
    \forall d \in \Dom(dept\_choice)$. Thus, $\mathit{NIE}$ = $(1 \cdot
    \frac{1}{2}\cdot\frac{3}{6}\cdot\frac{7}{12} +
    1\cdot\frac{3}{6}\cdot\frac{7}{12} + 1\cdot\frac{3}{6}\cdot\frac{5}{12}) -
    \frac{3}{6} = 0.14$. This indicates that most of the causal influence is
    transmitted through the indirect path and females receive less favorable
    predictions due to their choice of department. 
\end{example}

\section{Description of Fair Approaches} 
In this section, we provide detailed discussion of the fair approaches that we
evaluate in this paper.

%

\def\A{\mathbf{A}} \def\I{\mathbf{I}}

\subsection{Pre-processing Approaches} 
\subsubsection{\large{\kam}}

Kamiran and Calders~\cite{kamiran} introduce a pre-pro\-cessing approach that
targets the notion of demographic parity. We refer to this approach as \kam.
Assuming that the predictions $\hat{Y}$ reasonably approximates the ground
truth $Y$, \kam argues that $\hat{Y}$ is likely to be independent of the
sensitive attribute $S$, when the classifier is deployed, if $Y$ and $S$ are
independent in the training data. To this end, \kam samples tuples from the
training dataset $\D$ to create a modified training dataset $\D'$ in a way that
ensures that $Y$ and $S$ are independent in $\D'$. This is based on the
intuition that the classifier is likely to learn the independence from $\D'$
and will ensure demographic parity when deployed.

If $S$ and $Y$ are independent in $\D$, then $\forall s \in S$ and $\forall
y\in Y$, their expected joint probability $Pr_{exp}(S = s \land Y = y)$ should
be sufficiently close to their observed joint probability $Pr_{obs}(S = s \land
Y = y)$. These probabilities (over $\D$) are computed using the following
formulas:
\allowdisplaybreaks
\begin{align*}
Pr_{exp}(S = s \land Y = y) &:= \frac{|\{t : S_t = s\}|}{|\D|} \cdot \frac{|\{t : Y_t = y\}|}{|\D|}\\
Pr_{obs}(S = s \land Y = y) &:= \frac{|\{t : S_t = s \wedge Y_t = y\}|}{|\D|}\\
\end{align*}
If $Pr_{obs}$ is different from $Pr_{exp}$, then $S$ and $Y$ are not
independent in $\D$. \kam's goal is to modify $\D$ to obtain $\D'$ such that
the differences between the expected and the observed probabilities are
mitigated. To achieve this, \kam employs a weighted sampling technique that
compensates for the differences in $Pr_{exp}$ and $Pr_{obs}$. The technique
involves computing a weight for each tuple in $\D$ and then sampling the tuples
from $\D$, with probability proportional to their weights, to construct $\D'$.
The weight $w(t)$ of a tuple $t \in \D$ is computed as:
\begin{align*}
w(t) = \frac{Pr_{exp}(S = S_t  \land  Y = Y_t)}{ Pr_{obs}(S = S_t  \land  Y
= Y_t)}\\
\end{align*}
This weighting scheme guarantees that $Pr_{exp}$ and $Pr_{obs}$ are
sufficiently close over $\D'$, which implies that $Y$ and $S$ are independent
in $\D'$. \kam also provides empirical evidence that classifiers trained on
$\D'$ indeed satisfy demographic parity.

\smallskip\noindent\textbf{Implementation.} We collected the source code for \kam 
from the open source AI Fairness 360 library.\footnote{
\url{https://github.com/Trusted-AI/AIF360/tree/master/aif360/algorithms/preprocessing}\label{aifairness}}

\subsubsection{\large{\feld}}

Feldman et al.~\cite{feldman} propose a pre-processing approach that also
enforces demographic parity. We refer to this approach as \feld. \feld argues
that demographic parity can be ensured if the marginal distribution of each $X
\in \X$ is similar across the privileged and the unprivileged groups in the
training data. The basis of their argument is that if a model learns from such
data, it is likely to predict based on attributes that are independent of $S$,
which in turn will satisfy demographic parity within the model's predictions.
Unlike \kam, which does not modify attribute values, \feld directly modifies
the values for each attribute $X$. 

Given data $\D = [\D_{\X},\D_S;\D_Y]$ with the schema $(\X, S; Y)$, \feld
produces a modified dataset $\D' = [\D_{\X}',\D_S;\D_Y]$ where the marginal
distribution of each attribute is similar across the privileged and the
unprivileged groups. \feld repairs values of each individual attribute
separately to equalize the marginal distribution of the sensitive groups for
each attribute. To this end, \feld determines the quantile of each value $x \in
\D_X$ and replaces $x$ with the median of the corresponding quantiles from the
original marginal distributions $Pr(\D_X \mid \D_S = 1)$ and $Pr(\D_X \mid \D_S
= 0)$. This repair produces the modified attribute $\D_X'$ such that $Pr(\D_X'
\mid \D_S = 1) = Pr(\D_X' \mid \D_S = 0)$, and, thus, ensures that the modified
attribute is independent of the sensitive attribute.

Repeating the repair process for all attributes produces the modified
$\D_{\X}'$ and the modified dataset $\D'$. The level of repair is controlled
through a hyper-parameter $\lambda \in [0,1]$, where $\lambda = 0$ yields the
unmodified dataset and $\lambda = 1$ implies that the values within each
attribute are completely moved to the median.

\smallskip\noindent\textbf{Implementation.} We collected the source code for
\feld from the AI Fairness 360 library.\footref{aifairness} As the preferred
value of $\lambda$ is highly application-specific, we only report our findings
for the highest level of repair ($\lambda = 1.0$). We also note that the
implementation of \feld modifies both training and test data.

\subsubsection{\large{\calm}} 

Calmon et al.~\cite{calmon} propose a pre-processing approach that also
enforces demographic parity. We refer to this approach as \calm. Given the
joint distribution associated with the training data $\D$, \calm computes a new
distribution to transform $\X$ and $Y$ such that the dependency between $Y$ and
$S$ is reduced, without significantly distorting the data distribution. The new
joint distribution yields repaired training data $\D' = [\D_{\X}',\D_S;\D_Y']$.

To compute the new distribution, \calm constructs the following constraints that
must be satisfied: (1)~the difference between $Pr(\D_Y' \mid D_S = 0)$ and
$Pr(\D_Y' \mid D_S = 1)$ is below an allowable threshold, (2)~the new joint
distribution is sufficiently close to the original one, and (3)~no attribute
value in $\D_{\X}$ is substantially distorted to compute $\D_{\X}'$. \calm then
formulates a convex optimization problem that searches for the optimal new
distribution subject to the constraints. The resulting new distribution maps
each tuple from $\D$ to the modified dataset $\D'$ and classifiers learned on
$\D'$ is expected to satisfy demographic parity.

\smallskip\noindent\textbf{Implementation.} We collected the source code for
\calm from the AI Fairness 360 library.\footref{aifairness} Further, \calm
requires a distortion function to ensure that all individual values are
distorted within acceptable limits.\footnote{\url{https://github.com/maliha93/Fairness-Analysis-Code/blob/master/Preprocessing/Calmon/aif360/algorithms/preprocessing/optim_preproc_helpers/distortion_functions.py}}

\subsubsection{\large{\zhangpcf}}

\begin{sloppypar}
Zhang, Wu, and Wu~\cite{zhang2017} propose two pre-processing approaches that
target \emph{path-specific fairness} and \emph{direct causal effect}. We refer
to them as \zhangpcft and \zhangdcet. Given training data $\D =
[\D_{\X},\D_S;\D_Y]$, both approaches utilize a graphical causal model to
estimate the direct and indirect causal influence of the sensitive attribute on
the label. Then they repair $\D_Y$ minimally to produce $\D_Y'$ such that their
target fairness is achieved. Then the classifiers trained on the modified
training data $\D' = [\D_{\X},\D_S;\D_Y']$ are expected to be fair, under the
assumption that the distribution of the predictions made by a classifier follows
the distribution of the ground truth in the training data.
\end{sloppypar}

\zhangpcft enforces path specific fairness: a causal notion that ensures that
the causal influence of $S$ is not carried to $Y$ through any direct or
indirect paths. To repair $\D_Y$, \zhangpcft first verifies if $\D_Y$ violates
path-specific fairness. Specifically, $\D_S$ is a direct or indirect cause of
$\D_Y$ if intervening on $\D_S$ changes the expectations of $\D_Y$. \zhangpcft
utilizes the graphical causal model and estimates the effect of intervening on
$\D_S$ as the expected difference in $\D_Y$ when $\D_S$ changes from privileged
to unprivileged. Instead of measuring causal association through all paths
between $\D_S$ and $\D_Y$ in the causal graph, \zhangpcft can measure this
association through specific paths if desired. Path-specific fairness is
violated if the expected difference in $\D_Y$ is above some threshold
$\epsilon$. Next, \zhangpcft designs an optimization problem to repair $\D_Y$
such that the direct and indirect causal effects of $\D_S$ are removed, and the
causal model is minimally altered. The modified training dataset $\D'$ is then
used to train classifiers that enforce path-specific fairness.

In contrast, \zhangdcet aims to ensure that the direct causal effect, the
influence of $S$ that is transmitted through the direct path, is within an
allowable threshold $\tau$. It exploits the causal graph to determine a set of
parents ($Q$) of $Y$, such that $Q$ blocks all indirect paths from $S$ to $Y$.
To certify the presence of direct causal effect, \zhangdcet then computes
$\Delta_q = Pr(\D_Y = 1 \mid \D_S = 1, Q = q) - Pr(\D_Y = 1 \mid \D_S = 0, Q =
q)$, $\forall q \in \Dom\{Q\}$. Since $Q$ blocks all indirect paths, the
graphical criterion of d-separation~\cite{pearl} supports that $\Delta_q > 0$
occurs only if there exists some direct causal effect. Finally, \zhangdcet
modifies $D_Y$ to ensure that $\Delta_q \le \tau$ in each subpopulation $q$.

\smallskip\noindent\textbf{Implementation.} We retrieved the source code for the
approaches from the authors'
website.\footnote{\url{https://www.yongkaiwu.com/publication/}} In accordance
with the original papers, we set both $\epsilon$ and $\tau$ to be $0.05$. The
causal graphs for the datasets are constructed from prior
literature (detailed in Appendix~\ref{app:dags}).

\subsubsection{\large{\sal}} 

Salimi et al.~\cite{Salimi} propose a pre-processing approach that enforces
\emph{justifiable fairness}: a causal fairness notion that prohibits causal
dependency between the sensitive attribute $S$ and the prediction $\hat Y$,
except through admissible attributes. We refer to this approach as \sal. Unlike
other causal mechanisms, \sal does not require access to the causal model. \sal
assumes that $\hat Y$ is likely to be fair if a classifier is trained on data
$\D$ where ground truth $Y$ satisfies the target fairness notion. To that end,
it expresses justifiable fairness as an integrity constraint and repairs $\D$
to ensure that the constraint holds on the repaired training data $\D'$. Unlike
\kam, \sal does not modify the attributes and only repairs $\D$ by inserting or
deleting tuples.

As \sal does not depend on the causal model, it translates the condition for
justifiable fairness into an integrity constraint that must hold over the
training data. \sal partitions all attributes, except the ground truth, into
two disjoint sets: \emph{admissible} ($\A$) and \emph{inadmissible} ($\I$).
$\A$ contains the attributes that are allowed to influence or have causal
associations with prediction $\hat Y$, while $\I$ contains the rest of the
attributes. Given $\A$ and $\I$, justifiable fairness holds in $\D$ if $Y$ is
independent of $\I$ conditioned on $\A$. If the probability distribution
associated with $\D$ is uniform,\footnote{Datasets do not always have uniform
probability distribution in practice and additional pre-processing is required
to ensure that.} this integrity constraint can be checked through the following
multi-valued dependency: $\D = \Pi_{\A Y}(\D)\bowtie \Pi_{Y\I}(\D)$.

The goal of \sal is then to minimally repair $\D$ to form a new training
dataset $\D'$, such that the multi-valued dependency is satisfied. \sal
leverages techniques from maximum satisfiability~\cite{maxsat} and matrix
factorization~\cite{lee2001} to compute the minimal repair of $\D$ that
produces the optimal $\D'$ for training classifiers. However, these techniques
are NP-hard and application-specific knowledge is generally needed to determine
the sets of admissible and inadmissible attributes.

\smallskip\noindent\textbf{Implementation.} We collected the source code for
\sal from the authors via email, as no public repository is available.
Following the original paper, we choose race, gender, marital/relationship
status as inadmissible attributes whenever applicable, and the rest of the
attributes as admissible. Moreover, Salimi et al.\ discuss a second variation
of \salmst that partially repairs the data, but we do not include it as there
are no instructions on how to tune the level of repair for that. Lastly,
although there are experiments in the original paper that discuss techniques to
partition the training data and repair them in parallel, our evaluation is
limited to a single-threaded implementation.

%

\subsection{In-processing Approaches}

\subsubsection*{\large{\zaf}} Zafar et al.~\cite{zafar, zafarDM} propose two
in-processing approaches to enforce demographic parity and equalized odds. We
refer to them as \zdi and \zdmt, respectively. Both of these approaches
translate their corresponding fairness notion to a convex function of the
classifier parameters, and compute the optimal parameters that minimize
prediction errors while satisfying the notion.

To compute the optimal fair classifier, \zaf first formulates the learning
process as a constrained optimization problem. Given the training data $\D$,
the task of a classifier is to learn a decision boundary that separates the
tuples according to the ground truth. The optimal decision boundary, defined by
a set of parameters $\theta$, is the one that minimizes a convex loss function
$L(\theta)$ that measures the cost of prediction errors. For any tuple $t$, the
signed distance from the decision boundary determines the prediction.
Specifically, $\hat Y_t = 1$ if $d_\theta(\X_t) \ge 0$, where $d_\theta(\X_t)$
denotes the signed distance. \zaf does not explicitly use $S$ to determine the
prediction, rather they utilize $S$ to define the fairness constraint only.

\zdi introduces a proxy constraint for demographic parity, as directly
including the notion as a constraint leads to non-convexity in the loss
function.\footnote{Non-convex functions are computationally harder to optimize
than convex functions.} \zdi utilizes $d_\theta$ as a proxy for $\hat Y$ and
argues that the empirical covariance between the sensitive attribute and the
signed distance from the decision boundary is approximately zero, if the
prediction of a classifier is independent of the sensitive attribute. As
covariance is a convex function of $\theta$, it can be used define the proxy
constraint for demographic parity. Formally, covariance is computed as: $cov =
\frac{1}{|\D|} \sum_{t \in \D}(S_t - \bar{S}) d_\theta(\X_t)$, where $\bar{S}$
denotes the mean of $S$. Given the proxy constraint, \zdi proposes the
following two variations that work under different constraint settings:

\smallskip

\begin{itemize} 
	\item \textbf{Maximizing accuracy under fairness constraint.} This variation
(\zdifairt) computes the optimal classifier by minimizing $L(\theta)$ under the
condition that $cov \approx 0$. 

	\item \textbf{Maximizing fairness under accuracy constraint.} This
variation (\zdiacct) minimizes $cov$ as much as possible while ensuring
$L(\theta)$ is below a specified threshold. This is to avoid cases where
enforcing $cov \approx 0$ leads to high loss in the first variation.

\end{itemize} 

\smallskip

Both of the above variations produce a fair classifier that approximately
satisfies demographic parity. Similar to \zdi, \zdmt introduces a proxy
constraint for equalized odds. In particular, \zdmt proposes to use the
covariance between $S$ and $d_\theta$ of the misclassified tuples, since
covariance is approximately zero when a classifier satisfies equalized odds.
This covariance is computed as: $cov = \frac{1}{|\D|} \sum_{t \in \D}(S_t -
\bar{S}) g_\theta(\X_t)$, where $ g_\theta(\X_t) = -d_\theta(\X_t)$ if tuple
$t$ is misclassified, and $0$ otherwise. While this proxy is still not a convex
function of $\theta$, \zdmt efficiently computes classifier parameters that
maximize prediction accuracy under this proxy constraint through a disciplined
convex-concave program~\cite{dccp}.

\smallskip\noindent\textbf{Implementation.} We collected the source code for
\zaf from the authors' public
repository.\footnote{\url{https://github.com/mbilalzafar/fair-classification}}
We set all the hyper-parameters following the instructions specified within 
the source code (more details are in the authors' repository).

\subsubsection*{\large{\zhang}} Zhang, Lemoine, and others~\cite{zhang} propose
an in-processing approach that can enforce demographic parity, equalized odds,
or equal opportunity, by leveraging \emph{adversarial learning}, a technique
where a classifier and an adversary with mutually competing goals are trained
together. We refer to this approach as \zhang. Given the training data $\D =
(\X, S; Y)$, the goal of a classifier $f$ is to maximize the accuracy of
prediction $\hat Y$, while an adversary $a$ attempts to correctly predict the
sensitive attribute using $\hat Y$ (and $Y$). \zhang enforces the target notion
of fairness by designing the classifier to converge to optimal parameters such
that $\hat Y$ does not contain any information about $S$ that the adversary can
exploit.

In order to determine the optimal parameters, classifier $f$ minimizes a loss
function $L_f(\hat Y, Y)$. Adversary $a$ receives both $\hat Y$ and $Y$ if
equalized odds or equal opportunity is the target notion, otherwise $a$ only has
access to $\hat Y$ if demographic parity is enforced. The loss of adversary is
denoted as $L_a(\hat S, S)$. Both the classifier and adversary apply gradient
based optimizations~\cite{sgd} to iteratively update their parameters. Adversary
$a$ updates its parameters in a direction that minimizes $L_f$, while the
classifier $f$ only updates its parameters in a direction that both decreases
$L_f$ and increases $L_a$. This process of update guarantees that $f$ converges
to a solution where $L_f(\hat Y, Y)$ is minimized while $L_a(\hat S, S)$ is
approximately equal to the entropy of $S$, i.e., adversary gains no information
about $S$ from $\hat Y$ (and $Y$). Hence, the optimal classifier satisfies the
target fairness notion.

\smallskip\noindent\textbf{Implementation.} We collected the source code for \zhang 
from the open source AI Fairness 360 library.\footnote{
\url{https://github.com/Trusted-AI/AIF360/tree/master/aif360/algorithms/inprocessing}\label{inpro}}

\subsubsection*{\large{\kearns}} Kearns et al.~\cite{kearns} propose an
in-processing approach that enforces demographic parity and predictive
equality, a notion that requires equal $\mathit{FPR}$ for the privileged and
the unprivileged groups. We refer to this approach as \kearns. \kearns
approximately enforces the target fairness notion within a large set of
subgroups\footnote{The number of subgroups must be bounded by the classifier's
VC dimension~\cite{vcdom2015}.} defined using one or more sensitive attributes
(or user-specified attributes). To that end, \kearns solves a constrained
optimization problem to obtain optimal classifier parameters such that the
proportion of positive outcomes (demographic parity) or FPR (predictive
equality) is approximately equal to that of the population.

\looseness-1 \kearns begins by formulating the learning process and constraint
for the target fairness notion. Let $f: f(\X, S) \rightarrow \hat Y$ be a
classifier learned over training data $\D = (\X, S, Y)$. Moreover, let $G$ be
the set of the subgroups for which fairness must be ensured. Each $g \in G$
indicates a subgroup such that $g(\X_t, S_t) = 1$ means tuple $t$ belongs to
subgroup $g$. If predictive equality is the target notion, a group function
$\beta(g) = Pr(\hat Y = 1 \mid Y = 0) - Pr(\hat Y = 1 \mid Y = 0, g(\X, S) =
1)$ denotes the difference between overall $\mathit{FPR}$ and $\mathit{FPR}$
for group $g$. The fairness constraint is formally expressed as: $\alpha(g)
\beta(g) \leq \gamma, \forall g \in G$, where $\alpha(g)$ denotes the
proportion of tuples in group $g$ in order to exclude very small groups from
calculation and $\gamma$ is a tolerance parameter. Similar $\alpha(g)$ and
$\beta(g)$ can be derived for demographic parity.

Next, \kearns constructs the following optimization problem to compute optimal
$f$ that minimizes a loss function $l(\hat Y, Y)$:
\begin{align*}
    & \underset{f}{\min}\ \mathbb E[l(\hat Y, Y)]\\
    s.t.\ &\alpha(g) \beta(g) \leq \gamma,\ \forall g \in G\\
\end{align*}

While this optimization problem can be computationally hard in the worst case,
\kearns computes an approximate solution by solving an equivalent zero-sum
game~\cite{zerosum} in polynomial time and the optimal classifier approximately
satisfies the target fairness notion.

\smallskip\noindent\textbf{Implementation.} We collected the source code for
\kearns from the open source AI Fairness 360 library.\footref{inpro} The current
version does not include any implementation for demographic parity, and, thus,
our evaluation is limited to predictive equality. We use $\gamma = 0.005$, as
suggested in the source code.

\subsubsection*{\large{\celis}} Celis et al.~\cite{celis} propose an
in-processing approach that supports multiple fairness notion within a single
framework. We refer to this approach as \celis. \celis can accommodate a wide
range of notions: predictive parity, demographic parity, equalized odds, and
conditional accuracy equality. \celis reduces each fairness notion to a linear
function and presents an approach to solve the resulting linear constrained
optimization problem for obtaining a fair classifier that minimizes prediction
error.

In order to derive the fairness constraint, \celis first partitions the
training data $\D = (\X, S, Y)$ into groups according to the sensitive
attribute. Let $G$ be the set of groups and each $g_i \in G$ denotes a group
such that $g_i = (\X, S = i, Y) \subseteq \D$. For each group in $G$, \celis
then defines $q_i(f)$ that is a linear function or quotient of linear functions
of $Pr(\hat Y = 1 \mid g_i, \varepsilon_i)$, where $\varepsilon_i$ can be any
event relevant to the target fairness notion. Intuitively, $q_i(f)$ represents
the performance of classifier $f$ for group $g_i$. For example, $q_i(f)$
represents the probability of positive outcome when the target notion is
demographic parity. Given the function, a fairness notion can be expressed as
the following constraint: $\frac{\min_{i \in S}\ q_i(f)}{\max_{i \in S} q_i(f)}
\geq \tau$, where $\tau \in [0, 1]$ denotes a tolerance parameter. $\tau = 1$
implies that a classifier's performance must be equal across all groups.
Multiple constraints can be derived similarly if multiple notions need to be
enforced simultaneously.

Given the fairness constraint, \celis then formulates the process of finding the
optimal $f$ as the following constrained optimization problem:
\begin{align*}
    		& \underset{f}{\min}\ Pr(f(\X) \neq Y)\\
    s.t.\ 	& \frac{\min_{i \in S}\ q_i(f)}{\max_{i \in S} q_i(f)} \geq \tau\\
\end{align*}
To solve the above problem efficiently, \celis solves its dual instead using
Lagrange duality~\cite{lagrange1976}, which produces an approximately fair
classifier. This fair classifier can only guarantee $\min_{i \in S} q_i(f)\geq
\tau\cdot \max_{i \in S}\ q_i(f) - \epsilon - k$, where $\epsilon > 0$
represents some error that results from the approximation and $k$ denotes
additional error from estimating the probability distribution of data from
samples in $\D$.

\smallskip\noindent\textbf{Implementation.} We collected the source code for
\celis from the open source AI Fairness 360 library.\footref{inpro} We use $\tau
= 0.8$ as suggested in the source code. Further, we noted that the difference in
accuracy was minimal ($\le 1\%$) for any $\tau \in [0.8, 1.0]$, and, thus,
further hyper-parameter tuning was not necessary.  

\subsubsection*{\large{\thomas}} Thomas et al.~\cite{thomas2019} propose an
in-processing approach that can enforce demographic parity, equalized odds,
equal opportunity, and predictive equality. We refer to this approach as
\thomas. Given a training data $\D$ and a target fairness notion, \thomas
ensures that a classifier $f$ trained on $\D$ only picks solutions that satisfy
the fairness notion with high probability. \thomas computes an upper bound
(with high confidence) of the maximum possible fairness violation that a
classifier can incur at test time, and returns optimal classifier parameters
for which this worst possible violation is within an allowable threshold.

Given a function $g$ that quantifies discrimination according to the target
fairness notion and an objective function $L$ denoting a classifier's
correctness, \thomas's goal is formalized below:
\begin{align*}
    & \underset{f}{\argmax}\ L(f)\\
    s.t.\ &Pr(g(f(\D)) \le 0) \ge 1 - \delta\\
  \end{align*}
Here, $1 - \delta$ denotes the confidence upper bound. While \thomas allows
multiple $g$ to specify multiple fairness constraints simultaneously, it fails
to compute a feasible solution if the specified fairness notions cannot be
enforced at the same time. In order to compute the optimal fair solution,
\thomas splits the training data into two partitions: $\D_1$ and $\D_2$.
\thomas then uses gradient descent to compute a candidate solution that
maximizes the objective function on $\D_1$. Using $\D_2$, \thomas derives an
upper bound on the amount of discrimination that the candidate solution can
incur. This upper bound is computed using concentration inequalities, such as
Hoeffding's inequality~\cite{hoeffding2004} or Student's
t-test~\cite{brown1967}; and denotes the maximum amount of discrimination that
can occur, with a confidence of $1 - \delta$. Finally, \thomas selects the
candidate solution as the optimal solution if the upper bound is acceptable in
the context of the problem, and returns no solution otherwise.

\smallskip\noindent\textbf{Implementation.} We collected the source code for
\thomas from the authors via email, as no public repository is available.
Although \thomas supports multiple notions of fairness
(Figure~\ref{table:methods}), we exclude two notions---equal opportunity and
predictive equality---from our evaluation, as equalized odds encompasses both
these notions. We use $\delta = 0.05$, in accordance with the paper.

%
\subsection{Post-processing Approaches}

\subsubsection{\large{\kamdp}}

Kamiran, Karim, and others~\cite{kamiran2012} propose a post-processing
approach that enforces demographic parity. We refer to this approach as \kamdp.
\kamdp is based on the intuition that discriminatory decisions are most often
made for tuples close to the decision boundary, because the prediction
confidence (i.e., the probability of belonging to the predicted class) is low
for those tuples. Given a classifier, \kamdp derives a critical region around
the decision boundary and modifies the predictions for tuples in that region
such that demographic parity is satisfied.

Let $f: f(\X, S) \rightarrow \hat Y$ be a classifier and $Pr(\hat Y \mid \X,
S)$ be the prediction confidence. \kamdp defines a critical region around the
decision boundary where the prediction confidence is below a threshold
$\theta$, i.e., $\max(Pr(\hat Y = 1 \mid \X, S), Pr(\hat Y = 0 \mid \X, S)) <
\theta$. Here, $\theta$ is a hyper-parameter that can be tuned to find the
optimal critical region for the desired level of demographic parity. \kamdp
rejects the predictions for tuples that belong to the critical region as those
predictions are most likely to be discriminatory.

In order to enforce demographic parity, \kamdp modifies the predictions for the
tuples in the critical region using the following method: $\hat Y = 1$ is
assigned to all tuples belonging to the unprivileged group, while $\hat Y = 0$
is assigned to all tuples belonging to the privileged group. 

\smallskip\noindent\textbf{Implementation.} We collected the source code for
\kamdp from the open source AI Fairness 360
library.\footnote{\url{https://github.com/Trusted-AI/AIF360/tree/master/aif360/algorithms/postprocessing}\label{postpro}} 
We set all the hyper-parameters following the instructions specified within the source
code (more details are in the authors' repository).

\subsubsection{\large{\hardt}} Hardt et al.~\cite{hardt} propose a
post-processing approach that enforces equalized odds. We refer to this
approach as \hardt. Given the ground truth $Y$ and the sensitive attribute $S$
in the training data, \hardt learns the parameters of a new mapping $g:g(\hat
Y, S) \rightarrow \tilde Y$ to replace $\hat Y$ such that $\mathit{TPR}$ and
$\mathit{TNR}$ are equalized across the privileged and the unprivileged groups.

In order to enforce equalized odds, the new mapping $g$ must satisfy the
following condition: $Pr(\tilde{Y} = 1 \mid S = 1, Y = y) = Pr(\tilde{Y} = 1
\mid S = 0, Y = y), \forall y \in Y$. Given any standard loss function $l :
l(Y, \tilde Y) \rightarrow \mathbb{R}$ that quantifies the cost of incorrect
predictions, \hardt solves the following linear program to obtain the optimal
mapping:
\begin{equation*}
\begin{aligned}
  & \underset{g}{\min} \ \mathbb{E}[l(Y, \tilde{Y})]\\
  \text{s.t.\ } & Pr(\tilde{Y} = 1\mid S = 1, Y = y) = Pr(\tilde{Y} = 1\mid S = 0, Y = y), \forall y \in Y,\\
  \text{and } & Pr(\tilde{Y} = 1 \mid S = s, Y = y) \in [0, 1], \forall y \in Y,\ s \in S,\\
\end{aligned}
\end{equation*} 
where $\mathbb{E}[l(Y, \tilde{Y})]$ is the expected loss. The solution to this
linear program always provides a mapping for modifying the predictions such
that equalized odds is satisfied.

\smallskip\noindent\textbf{Implementation.} We collected the source code for
\hardt from a public repository.\footnote{\url{https://github.com/gpleiss/equalized_odds_and_calibration}\label{hardtpleiss}}


\begin{figure}[t]
  \centering
  \begin{subfigure}[t]{0.49\textwidth}
  \includegraphics[width=1\textwidth]{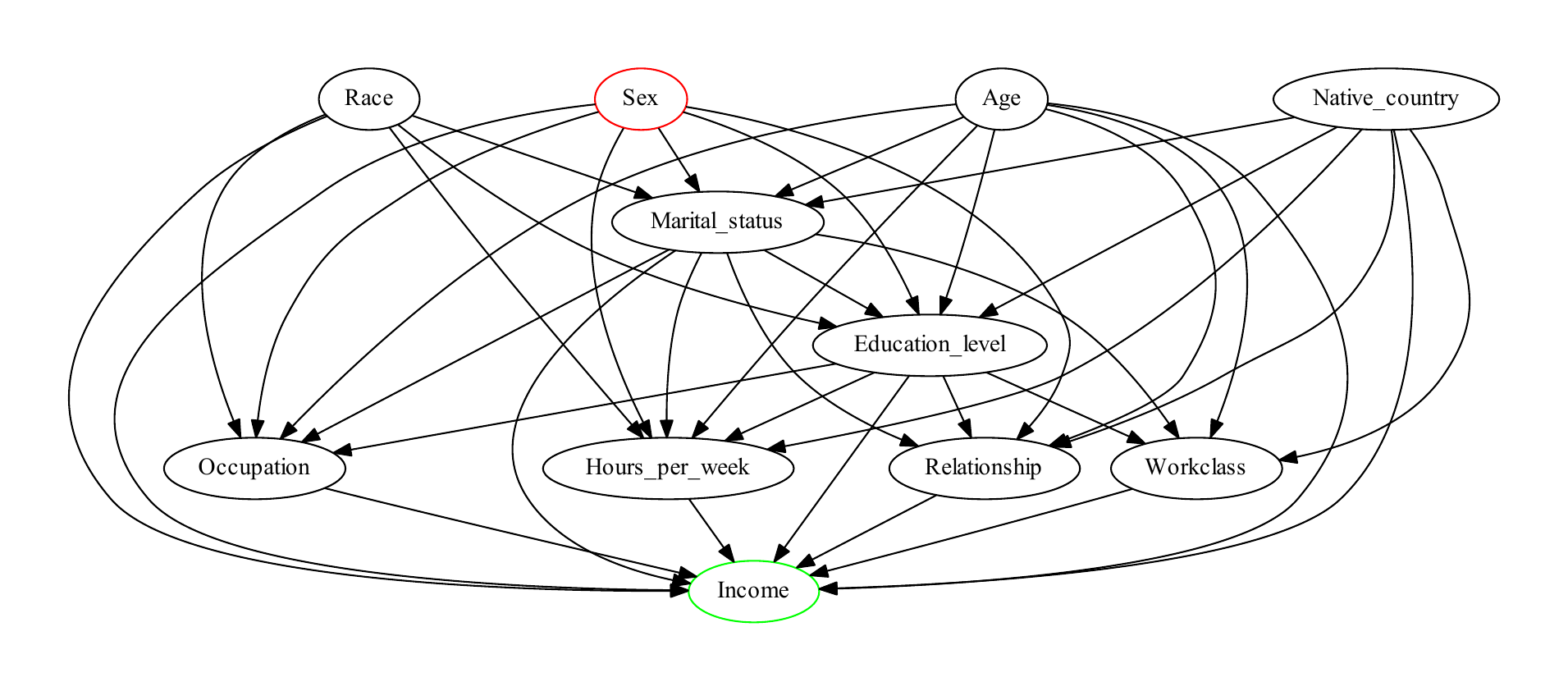}
  \vspace{-7mm}
  \caption{Adult}
  \end{subfigure}

  \begin{subfigure}[t]{0.49\textwidth}
  \includegraphics[width=1\textwidth]{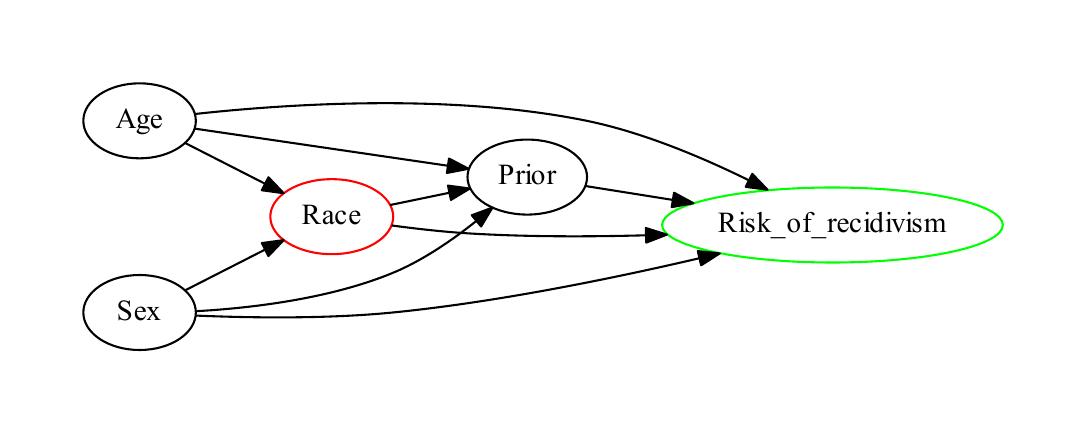}
  \vspace{-7mm}
  \caption{COMPAS}
  \end{subfigure}
  
  \begin{subfigure}[t]{0.49\textwidth}
  \includegraphics[width=1\textwidth]{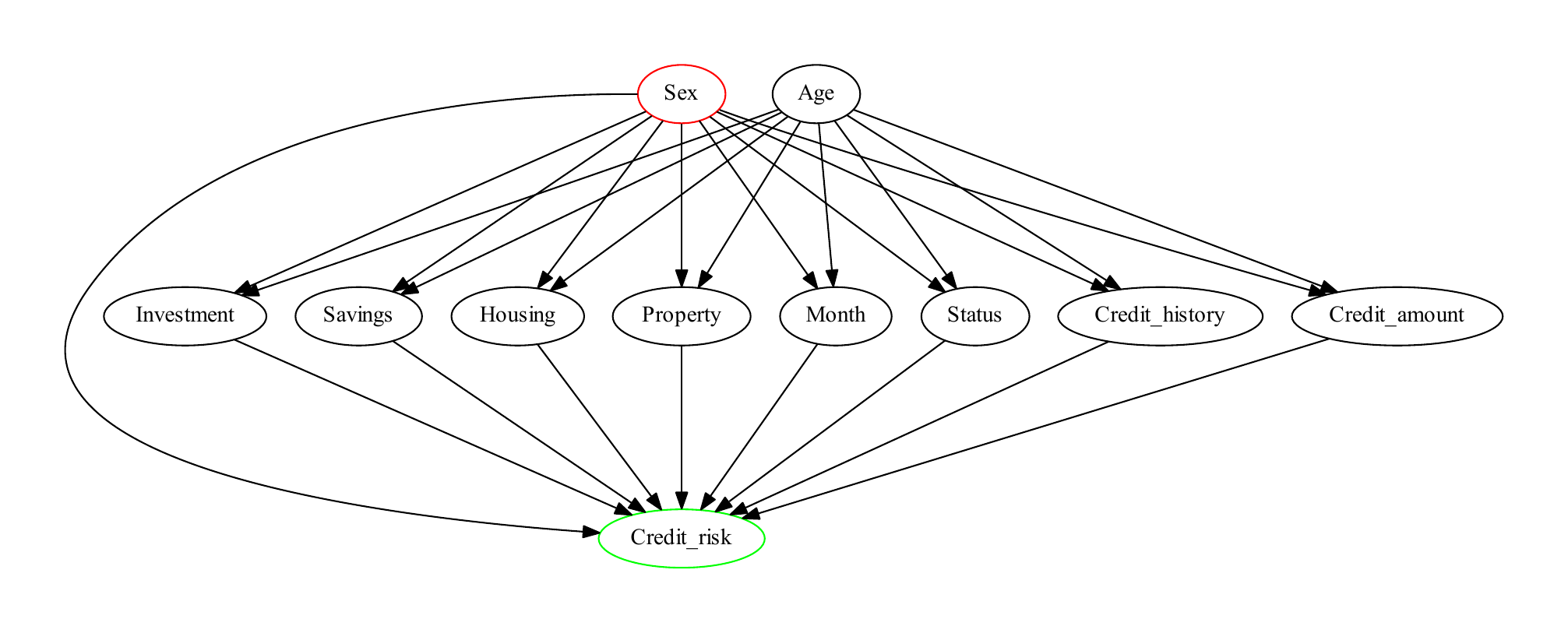}
  \vspace{-7mm}
  \caption{German}
  \end{subfigure}
  
  
  \caption{The causal graph underlying each dataset in our evaluation. The nodes
  highlighted in red and green denote the sensitive attributes and the labels
  respectively.  }
  \vspace{-2mm}
  \label{fig:dags}
  
\end{figure}

\subsubsection{\large{\pleiss}} 

Pleiss et al.~\cite{pleiss} propose a post-processing approach to ensure that a
calibrated classifier satisfies \emph{equal opportunity}---equal $\mathit{TPR}$
across the sensitive groups---or \emph{predictive equality}---equal
$\mathit{FPR}$ across the sensitive groups---or a weighted combination thereof.
We refer to this approach as \pleiss. \pleiss derives a new predictor for the
group with higher $\mathit{TPR}$ (or lower $\mathit{FPR}$) and replaces $\hat
Y$ in order to enforce the fairness notion.

\pleiss begins by assuming that the optimal classifier $f$, learned on the
training data $\D$, is reliable and calibrated, i.e., $Pr(Y = 1 \mid \hat{Y} =
y) = y, \forall y \in Y$. Given $f$, \pleiss derives two cost functions, $C_0
(f)$ and $C_1 (f)$, for the unprivileged and the privileged groups,
respectively. Depending on the target fairness notion, this cost function
denotes the $\mathit{TPR}$, or the $\mathit{FPR}$, or a weighted combination
thereof, for the corresponding group. $f$ violates fairness if it favors one
group, i.e., $C_0(f) \neq C_1(f)$.

To enforce the target fairness notion, \pleiss derives a new predictor for the
favored group, such that it replaces a random subset of $\hat Y$ to decrease the
$\mathit{TPR}$ (or increase $\mathit{FPR}$) to make it approximately equal to
the other (unfavored) group. For any tuple $t$ in the favored group, the actual
prediction $\hat Y_t$ is withheld with probability $\alpha \in [0,1]$, where
$\alpha$ depends on the difference between $C_0$ and $C_1$. Then $\hat Y_t$ is
replaced with $\tilde Y_t$, such that $\tilde Y_t = 1$ with probability
proportional to the fraction of positive tuples in the favored group. This
modification technique decreases the classifier's performance for the favored
group while maintaining classifier calibration, and approximately satisfies the
target fairness notion. Pleiss et al. acknowledge that their approach satisfies
group-level fairness while intentionally violating individual-level fairness due
to randomness in predictions.

\vspace{3mm}\noindent\textbf{Implementation.} We collected the source code for
\pleiss from the authors' public repository.\footref{hardtpleiss} We use equal
opportunity as the fairness notion, since minimizing the difference in terms
of favorable outcomes---i.e., equal $\mathit{TPR}$ across the sensitive
groups---is more appropriate as the fairness goal in the context of our
datasets. Further, a weighted combination of equal opportunity and predictive
equality led to very poor performance in terms of correctness in most cases.

\vspace{-3mm}
\subsection{Additional Approaches Under Evaluation}
We evaluated 3 additional variations of 2 approaches by Madras et
al.~\cite{madras2018} and Agarwal et al.~\cite{agarwal2018}. Madras et al.
propose a pre-processing approach to learn a fair representation of data that
assures that naively trained classifiers on it will be reasonably fair and
accurate. Specifically, this approach utilizes adversarial learning by providing
appropriate adversarial objective functions that upper bounds the unfairness of
arbitrary downstream classifiers in the limit of adversarial training. We refer
to this approach as \madrast as it targets demographic parity. On the other
hand, Agarwal et al. propose an in-processing approach that can enforce multiple
definitions of fairness. The key mechanism is to break down fair classification
to a sequence of cost-sensitive classification problems, whose solutions yield a
randomized classifier with the lowest empirical error subject to the target
fairness constraints. We evaluate two variations of this approach---\agarwaldpt
and \agarwaleot---that target demographic parity and equalized odds. 

The aforementioned approaches are not included in the main report as Zhang et
al. (\zhangt) utilizes adversarial learning techniques similar to Madras et al.,
and Celis et al. (\celist) applies a mechanism similar to Agarwal et al. to
cover a wider range of fairness notions. Figure~\ref{fig:others} shows their
correctness and fairness over Adult, COMPAS and German.
\Cref{fig:runtime_oth_experiment,fig:other_robustness,fig:experiment_sensitivity,fig:experiment_var,fig:experiment_data_eff}
show the results of their efficiency and scalability, robustness to data errors,
sensitivity to the underlying ML model, stability, and data efficiency,
respectively.

\vspace{-2.5mm}
\section{Causal graphs of all datasets}\label{app:dags}
The computation of causal metrics in our evaluation require the causal structure
or graphical model underlying the datasets. Figure~\ref{fig:dags} shows the
causal graphs corresponding to each dataset. These causal graphs are well
accepted and widely used in prior literature~\cite{nabi, zhang2017, zhang2018}.

%

\begin{figure*}[t]
  \centering
  \hspace{5mm}\includegraphics[width=0.78\textwidth]{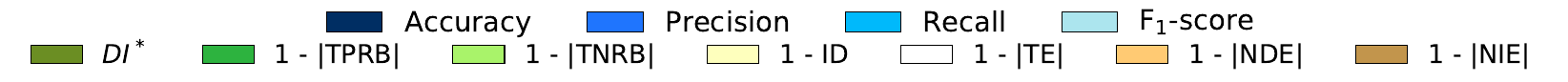}
  \vspace{2mm}
  
  \begin{subfigure}[t]{0.32\textwidth}
  \includegraphics[width=1\textwidth]{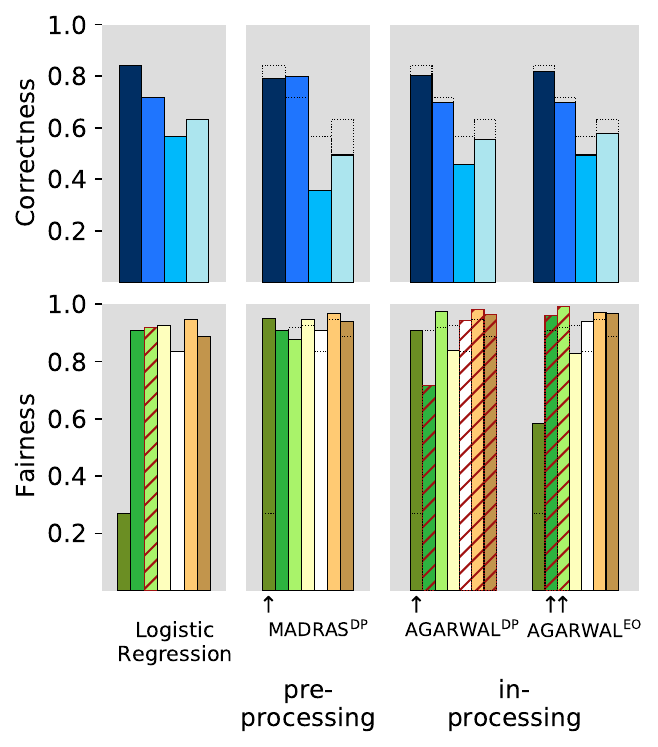}
  \vspace{-6mm}
  \caption{Adult}
  \end{subfigure}
  \begin{subfigure}[t]{0.32\textwidth}
  \includegraphics[width=1\textwidth]{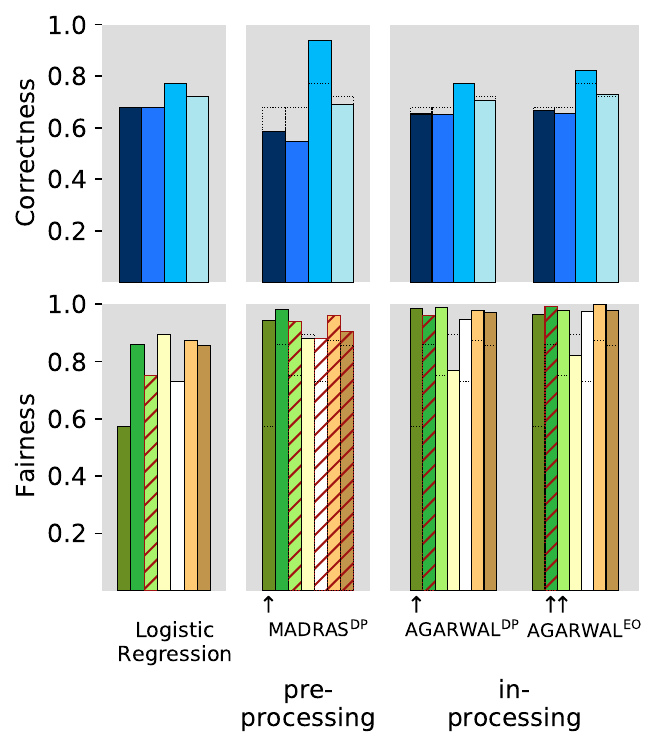}
  \vspace{-6mm}
  \caption{COMPAS}
  \end{subfigure}
  \begin{subfigure}[t]{0.32\textwidth}
  \includegraphics[width=1\textwidth]{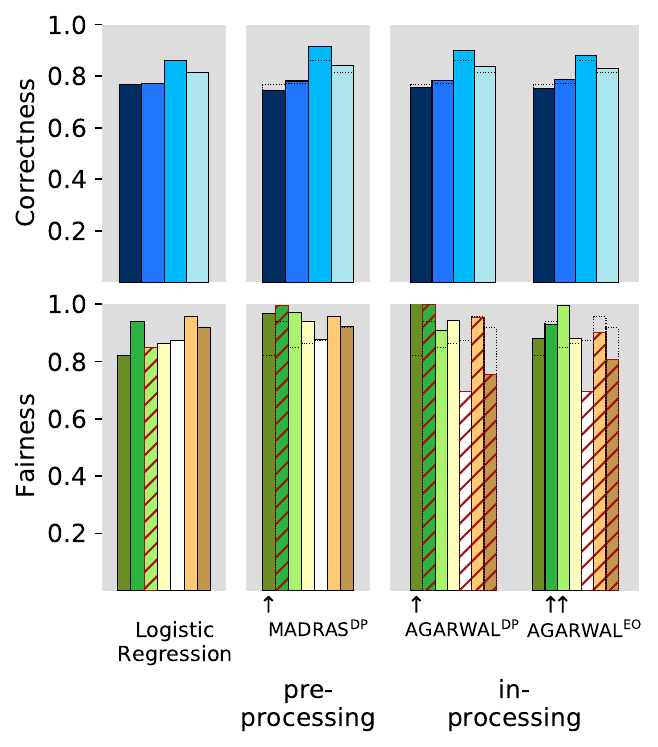}
  \vspace{-6mm}
  \caption{German}
  \end{subfigure}
  
  \vspace{-2mm} 
  \caption{Correctness and fairness scores of the 3 additional fair classification
  approaches over (a)~Adult, (b)~COMPAS, and (c)~German datasets. Higher scores
  for correctness (fairness) metrics correspond to more correct (fair) outcomes.
  The bars highlighted in red denote the reverse direction of the remaining
  discrimination---favoring the unprivileged group more than the privileged
  group. The arrows ($\uparrow$) denote the fairness metric(s) each approach is
  optimized for. The bar plots for \orig are overlaid for aiding visual
  comparison.  } 
  \vspace{4mm}
  \label{fig:others}
\end{figure*}
   
\balance

\section{Results of 5-fold cross validation}
We cross validated each of our approach through 5-fold cross validation with
50\%-20\%-30\% split for the train-validation-test sets.
\Cref{fig:cross_val_adult,fig:cross_val_compas,fig:cross_val_german} presents
the average of each metric over all the datasets.

\section{Complete results of robustness to data errors}

Figure~\ref{fig:experiment_robustness} shows the results of our robustness
experiments in terms of all correctness and fairness metrics over 3 different
erroneous training set derived from COMPAS. 

\section{Complete results of sensitivity to underlying ML model}

We study the sensitivity of pre- and post-processing approaches to the choice of
ML model by pairing them with each of the following models: Logistic Regression
(LR), Sup-port Vector Machine (SVM), Random Forest (RF), k-Nearest Neigh-bors
(k-NN), and Multi-layer Perceptron (MLP). We implemented each classifier using
Scikit-learn (version 0.22.1). We chose hyper-parameters that maximize
correctness in the fairness-unaware setting; they are as follows:

\begin{itemize}
  \item \textbf{LR:} \texttt{l2} regularization.
  \item \textbf{SVM:} \texttt{rbf} kernel with \texttt{scaled} gamma co-efficient.
  \item \textbf{RF:} A forest of \texttt{40} trees, each with a maximum depth of \texttt{100}.
  \item \textbf{k-NN:} \texttt{33} nearest neighbors.
  \item \textbf{MLP:} \texttt{1} hidden layer with \texttt{20} neurons,
  \texttt{l2} regularization with \texttt{alpha=0.01}, and \texttt{sigmoid}
  activation for output.
\end{itemize}

\smallskip

\noindent We do not mention the additional hyper-parameters associated with each
classifier as they were kept at the default level.
Figure~\ref{fig:experiment_sensitivity} presents the sensitivity of all pre- and
post-processing approaches to the underlying ML model over the Adult dataset. 

\section{Stability of Fair Approaches}\label{sec:stability} We evaluate the
stability of all the approaches through a variance test on their correctness and
fairness over the Adult dataset. We executed each fair approach $10$ times with
random folds, using $66.67\%$ of the data for training and the rest for testing.
We report our findings on the stability of all correctness and fairness metrics
in Figure~\ref{fig:experiment_var}; the results are similar over the other
datasets.

\smallskip\noindent\emph{Approaches are generally stable}. Most approaches show
low variance and have a very small number of outliers. \hardtt shows high
variance in precision and F$_1$-score, but are stable on the other metrics. In
general, approaches can exhibit slightly higher variance in the metrics they do
not target.

\medskip
\noindent
\fbox{
\parbox{0.96\columnwidth}{
\emph{Key takeaway:}
All approaches generally exhibit low variance in terms of correctness and
fairness over different train-test splits. High-variance behavior is rare, and
there is no significant trend across the dimension of pre-, in-, and
post-processing.
}
}
\bigskip

\section{Data Efficiency of Fair Approaches}\label{sec:data-efficiency}

In this section, we examine how the size of the training set impacts the
accuracy and fairness of all approaches. We used the Adult dataset, as it is
the largest, and executed a new instance of each approach ranging the size of
the training set from 0.1K to 36K data points, sampled from the dataset. We do
not present separate variants of the same approach unless they differ
significantly in behavior. We report our findings in
Figure~\ref{fig:experiment_data_eff}.

\smallskip
\noindent\emph{Approaches are generally data-efficient}. Most approaches we
evaluate produce stable results when trained on 1K data points or higher.
However, the data-efficiency can be unpredictable in metrics not optimized by an
approach. \kamt, \hardt, and \pleisst appear to be the most data-efficient,
achieving their best correctness-fairness balance with as few as 100 data
points. Further, we don't observe any significant pattern across the dimension
of pre-, in-, and post-processing. 

\bigskip
\noindent
\fbox{
\parbox{0.96\columnwidth}{
\emph{Key takeaway:}
Most approaches are data-efficient, with the size of training data not having
significant impact on their correctness-fairness balance. No stage (pre-, in-,
and post-processing) appears to have an edge in this aspect.
}
}

\begingroup
\setlength{\tabcolsep}{2.5pt}
\renewcommand{\arraystretch}{1.0}

\begin{figure*}[t]
  \centering 
  \begin{tabular}{lcccccccccccc}
      \toprule
      &\bf{Accuracy}	 &\bf{Precision}  &\bf{Recall}  &\bf{F$_1$-score}  &\bf{DI$^*$}   &\bf{1 - |TPRB|}   &\bf{1 - |TNRB|}  &\bf{1 - CD}  &\bf{1 - |TE|} &\bf{1 - |NDE|}  &\bf{1 - |NIE|} \\
      \midrule
      \orig         &0.84     & 0.72      & 0.56   & 0.63        & 0.27  & 0.91 & 0.92 & 0.93 & 0.84 & 0.95 & 0.89 \\
      \kamt         &0.77     & 0.64      & 0.21   & 0.32        & 0.79  & 0.96 & 0.99 & 1.00 & 0.86 & 1.00 & 0.86 \\
      \feldt        &0.79     & 0.63      & 0.21   & 0.31        & 0.80  & 0.92 & 0.97 & 1.00 & 0.87 & 0.99 & 0.87 \\
      \calmt        &0.75     & 0.50      & 0.49   & 0.50        & 0.89  & 0.96 & 0.96 & 1.00 & 0.96 & 0.99 & 0.97 \\
      \zhangpcft    &0.82     & 0.66      & 0.54   & 0.59        & 0.57  & 0.94 & 0.88 & 0.86 & 0.91 & 0.97 & 0.94 \\
      \zhangdcet    &0.81     & 0.72      & 0.42   & 0.53        & 0.21  & 0.80 & 0.93 & 0.99 & 0.87 & 0.97 & 0.91 \\
      \salmst       &0.81     & 0.72      & 0.54   & 0.62        & 0.61  & 0.90 & 0.93 & 0.96 & 0.90 & 0.97 & 0.94 \\
      \salmft       &0.78     & 0.68      & 0.61   & 0.65        & 0.61  & 0.87 & 0.93 & 0.95 & 0.90 & 0.98 & 0.92 \\
      \madrast      &0.79	    &0.79     	&0.35    &0.49	       &0.95	 &0.91  &0.88  &0.95	&0.91	 & 0.97	& 0.94\\

      \zdifairt     &0.76     & 0.53      & 0.56   & 0.54        & 0.95  & 0.81 & 0.96 & 1.00 & 0.97 & 0.99 & 0.98 \\
      \zdiacct      &0.82     & 0.70      & 0.49   & 0.58        & 0.74  & 0.82 & 0.99 & 1.00 & 0.98 & 0.99 & 0.99 \\
      \zdmt         &0.82     & 0.73      & 0.47   & 0.57        & 0.52  & 0.93 & 0.98 & 1.00 & 0.83 & 0.99 & 0.84 \\
      \zhangt       &0.80     & 0.64      & 0.44   & 0.52        & 0.57  & 0.91 & 1.00 & 0.94 & 0.95 & 0.99 & 0.96 \\
      \kearnst      &0.82     & 0.79      & 0.36   & 0.50        & 0.26  & 0.89 & 0.96 & 0.96 & 0.80 & 0.96 & 0.84 \\
      \celist       &0.72     & 0.53      & 0.88   & 0.66        & 0.99  & 0.98 & 0.92 & 0.98 & 0.93 & 0.95 & 0.98 \\
      \thomasdpt    &0.72     & 0.60      & 0.30   & 0.40        & 0.98  & 0.88 & 0.94 & 0.93 & 0.99 & 1.00 & 0.99 \\
      \thomaseot    &0.79     & 0.55      & 0.46   & 0.50        & 0.60  & 0.99 & 1.00 & 0.90 & 0.93 & 0.99 & 0.94 \\
      \agarwaldpt   &0.80	    &0.69	      &0.45	   & 0.55	       &0.90	 & 0.71	& 0.97 & 0.84	& 0.94 &0.98  &0.96  \\
      \agarwaleot   &0.81	    &0.69     	&0.49	   & 0.57        &0.58	 &0.96	&0.99	 &0.82	& 0.93 &0.97	&0.96  \\

      \kamdpt       &0.76     & 0.52      & 0.78   & 0.62        & 0.95  & 0.85 & 0.92 & 0.93 & 0.96 & 1.00 & 0.96 \\
      \hardtt       &0.78     & 0.58      & 0.45   & 0.51        & 0.97  & 0.90 & 0.93 & 0.65 & 0.88 & 0.92 & 0.96 \\
      \pleisst      &0.74     & 0.71      & 0.34   & 0.46        & 0.54  & 0.97 & 0.87 & 0.79 & 0.95 & 0.98 & 0.97\\
      \bottomrule
    \end{tabular}

\caption{The average of all correctness and fairness metrics after 5-fold cross validation on Adult.}
\vspace{2mm}
\label{fig:cross_val_adult}
\end{figure*}

\begin{figure*}[t]
  \centering 
  \begin{tabular}{lccccccccccc}
      \toprule
      &\bf{Accuracy}	 &\bf{Precision}  &\bf{Recall}  &\bf{F$_1$-score}  &\bf{DI$^*$}   &\bf{1 - |TPRB|}   &\bf{1 - |TNRB|}  &\bf{1 - CD}  &\bf{1 - |TE|} &\bf{1 - |NDE|}  &\bf{1 - |NIE|} \\
      \midrule
      \orig           &0.68     & 0.68      & 0.77   & 0.72        & 0.57  & 0.86 & 0.75 & 0.90 & 0.73 & 0.87 & 0.86 \\
      \kamt           &0.61     & 0.66      & 0.78   & 0.71        & 0.94  & 0.93 & 0.91 & 0.74 & 0.78 & 0.99 & 0.79 \\
      \feldt          &0.63     & 0.62      & 0.86   & 0.72        & 0.82  & 0.90 & 0.85 & 1.00 & 0.80 & 0.91 & 0.90 \\
      \calmt          &0.64     & 0.64      & 0.77   & 0.70        & 0.78  & 0.9  & 0.83 & 0.98 & 0.83 & 0.92 & 0.91 \\
      \zhangpcft      &0.59     & 0.57      & 0.87   & 0.69        & 0.91  & 0.95 & 0.92 & 0.89 & 0.96 & 0.99 & 0.98 \\
      \zhangdcet      &0.64     & 0.62      & 0.84   & 0.72        & 0.81  & 0.89 & 0.86 & 0.95 & 0.85 & 1.00 & 0.85 \\
      \salmst         &0.62     & 0.64      & 0.51   & 0.57        & 0.74  & 0.81 & 0.83 & 0.91 & 0.91 & 0.97 & 0.94 \\
      \salmft         &0.62     & 0.72      & 0.64   & 0.67        & 0.61  & 0.88 & 0.81 & 0.98 & 0.88 & 0.96 & 0.92 \\
      \madrast        &0.58	    & 0.54     	& 0.93	 & 0.69  	     & 0.94	 & 0.98	& 0.94 & 0.88	& 0.87 & 0.96	& 0.90\\

      \zdifairt       &0.55     & 0.55      & 0.94   & 0.69        & 0.99  & 1.00  & 0.99 & 1.00  & 0.85 & 0.98 & 0.87 \\
      \zdiacct        &0.65     & 0.57      & 0.96   & 0.72        & 0.90  & 0.96 & 0.88 & 1.00  & 0.83 & 0.92 & 0.91 \\
      \zdmt           &0.61     & 0.55      & 1.00   & 0.71        & 0.92  & 1.00  & 0.97 & 1.00  & 0.99 & 1.00  & 0.99 \\
      \zhangt         &0.68     & 0.68      & 0.77   & 0.72        & 0.87  & 0.97 & 0.94 & 0.80  & 0.87 & 0.99 & 0.88 \\
      \kearnst        &0.54     & 0.54      & 1.00   & 0.70        & 1.00  & 0.99 & 1.00  & 1.00  & 0.89 & 0.98 & 0.91 \\
      \celist         &0.67     & 0.66      & 0.80   & 0.72        & 0.61  & 0.75 & 0.68 & 0.90  & 0.74 & 0.95 & 0.79 \\
      \thomasdpt      &0.62     & 0.54      & 0.94   & 0.68        & 1.00  & 0.96 & 1.00  & 1.00  & 0.91 & 1.00  & 0.91 \\
      \thomaseot      &0.60     & 0.54      & 0.99   & 0.70        & 0.92  & 1.00  & 0.99 & 0.97 & 0.90  & 0.99 & 0.91 \\
      \agarwaldpt     &0.65	    &0.65	      &0.77	   &0.70	       &0.98	&0.96	&0.98	&0.76	&0.94	&0.97	&0.97\\
      \agarwaleot     &0.66	    &0.65	      &0.82	   &0.72	       &0.96	&0.99	&0.97	&0.82	&0.97	&0.99	&0.97\\

      \kamdpt         &0.60      & 0.72      & 0.60    & 0.65      & 0.93  & 0.98 & 1.00  & 0.88 & 0.92 & 0.98 & 0.95 \\
      \hardtt         &0.62     & 0.62      & 0.74   & 0.68        & 0.92 & 0.91 & 0.93 & 0.53 & 0.80  & 0.88 & 0.92 \\
      \pleisst        &0.59     & 0.66      & 0.79   & 0.72        & 0.53  & 0.91 & 0.87 & 0.46 & 0.65 & 0.88 & 0.78 \\
      \bottomrule
    \end{tabular}

\caption{The average of all correctness and fairness metrics after 5-fold cross validation on COMPAS.}
\vspace{4mm}
\label{fig:cross_val_compas}
\end{figure*}

\begin{figure*}[t]
  \centering 
  \begin{tabular}{ccccccccccccc}
      \toprule
      &\bf{Accuracy}	 &\bf{Precision}  &\bf{Recall}  &\bf{F$_1$-score}  &\bf{DI$^*$}   &\bf{1 - |TPRB|}   &\bf{1 - |TNRB|}  &\bf{1 - CD}  &\bf{1 - |TE|} &\bf{1 - |NDE|}  &\bf{1 - |NIE|} \\
      \midrule
      \orig           &0.77     & 0.77      & 0.86   & 0.81        & 0.82  & 0.94 & 0.85 & 0.86 & 0.87 & 0.96 & 0.92 \\
      \kamt           &0.72     & 0.72      & 0.98   & 0.83        & 0.96 & 0.97 & 0.90  & 0.98 & 0.87 & 0.91 & 0.96 \\
      \feldt          &0.73     & 0.73      & 0.96   & 0.83        & 0.93 & 0.97 & 0.82 & 1.00  & 0.82 & 0.95 & 0.86 \\
      \calmt          &0.76     & 0.78      & 0.91   & 0.84        & 0.99  & 1.00  & 0.97 & 1.00  & 0.87 & 0.95 & 0.92 \\
      \zhangpcft      &0.71     & 0.77      & 0.90    & 0.83        & 0.96  & 1.00  & 0.96 & 0.96 & 0.95 & 1.00  & 0.96 \\
      \zhangdcet      &0.74     & 0.77      & 0.89   & 0.83        & 0.70   & 0.80  & 0.65 & 0.79 & 0.90  & 0.98 & 0.92 \\
      \salmst         &0.68     & 0.64      & 0.63   & 0.63        & 0.84  & 0.94 & 0.91 & 0.84 & 0.91 & 0.99 & 0.92 \\
      \salmft         &0.70      & 0.63      & 0.67   & 0.65        & 0.85  & 0.97 & 0.89 & 0.89 & 0.90  & 0.95 & 0.95 \\
      \madrast        &0.74	    &0.78	      &0.91	    &0.84	       &0.96	&0.99	&0.97	&0.94	&0.87	&0.95	&0.92\\

      \zdifairt       &0.75     & 0.77      & 0.91   & 0.83        & 0.99  & 0.96 & 0.90  & 1.00  & 0.72 & 0.99 & 0.74 \\
      \zdiacct        &0.74     & 0.79      & 0.86   & 0.82        & 0.93  & 0.92 & 0.93 & 1.00  & 0.82 & 0.97 & 0.86 \\
      \zdmt           &0.75     & 0.78      & 0.89   & 0.83        & 0.94  & 0.92 & 0.91 & 1.00  & 0.76 & 0.89 & 0.86 \\
      \zhangt         &0.75     & 0.79      & 0.87   & 0.83        & 0.86  & 0.88 & 0.98 & 0.90  & 0.82 & 0.98 & 0.85 \\
      \kearnst        &0.61     & 0.66      & 0.88   & 0.76        & 0.92 & 0.90  & 1.00  & 0.92 & 0.95 & 1.00  & 0.95 \\
      \celist         &0.74     & 0.79      & 0.85   & 0.82        & 0.96  & 0.95 & 0.93 & 0.98 & 0.89 & 0.98 & 0.91 \\
      \thomasdpt      &0.73     & 0.71      & 0.88   & 0.79        & 0.99  & 0.98 & 0.94 & 1.00  & 0.87 & 0.94 & 0.93 \\
      \thomaseot      &0.73     & 0.73      & 0.89   & 0.80        & 0.94  & 0.99 & 0.98 & 0.98 & 0.93 & 0.96 & 0.97 \\
      \agarwaldpt     &0.75	    &0.78	      &0.89	   &0.83	       &0.99	&0.99	&0.90	&0.94	&0.69	&0.95	&0.75\\
      \agarwaleot     &0.75	    &0.78	      &0.87	   &0.83	       &0.87	&0.92	&0.99	&0.88	&0.69	&0.90	&0.80\\

      \kamdpt         &0.67     & 0.87      & 0.61   & 0.72        & 0.95 & 0.94 & 0.95 & 0.89 & 0.75 & 0.92 & 0.84 \\
      \hardtt         &0.71     & 0.77      & 0.84   & 0.80         & 0.93 & 0.93 & 0.91 & 0.64 & 0.70  & 0.90  & 0.80  \\
      \pleisst        &0.72     & 0.76      & 0.92   & 0.83        & 0.93 & 0.96 & 0.84 & 0.73 & 0.97 & 1.00  & 0.97\\
      \bottomrule
    \end{tabular}

\caption{The average of all correctness and fairness metrics after 5-fold cross validation on German.}
\vspace{4mm}
\label{fig:cross_val_german}

\end{figure*}
\endgroup

\begin{figure*}[t]
  \centering
  \begin{subfigure}[t]{0.99\textwidth}
  \includegraphics[width=1\textwidth]{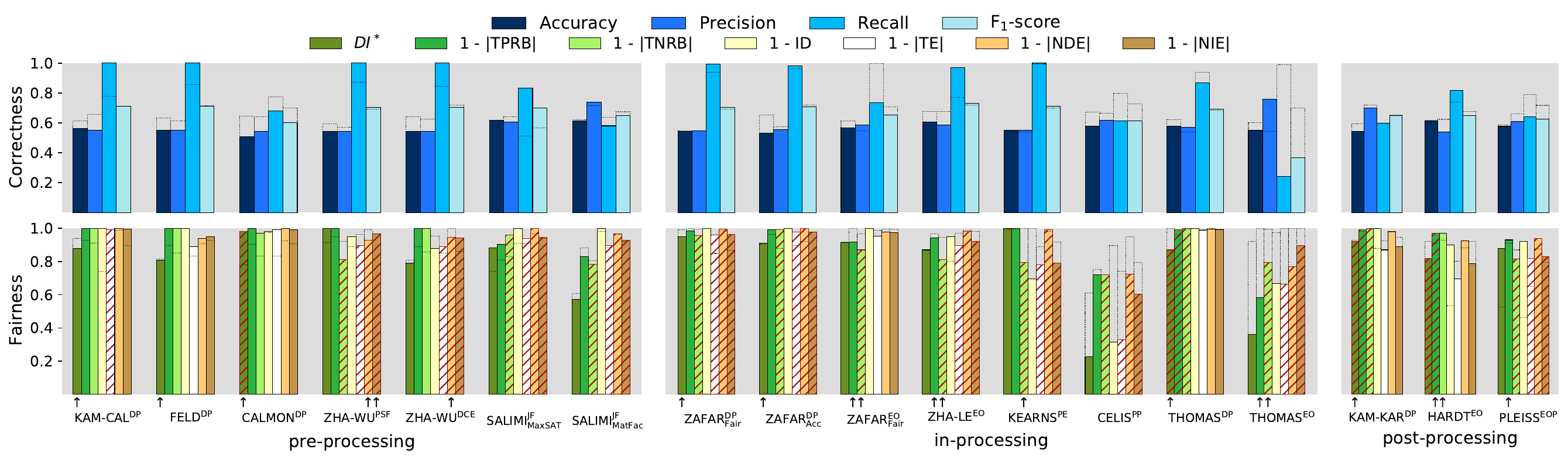}
  \vspace{-6mm}
  \caption{$T_1$}
  \vspace{5mm}
  \label{fig:experiment_t1}
  \end{subfigure}

  \begin{subfigure}[t]{0.99\textwidth}
  \includegraphics[width=1\textwidth]{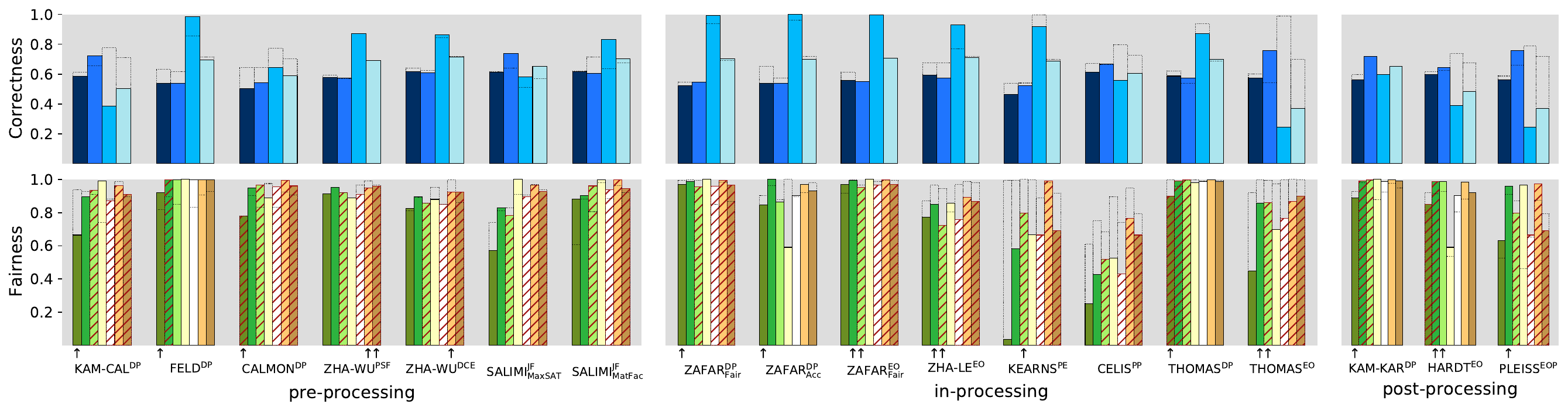}
  \vspace{-6mm}
  \caption{$T_2$}
    \vspace{5mm}
  \label{fig:experiment_t2}
  \end{subfigure}
  
  \begin{subfigure}[t]{0.99\textwidth}
  \includegraphics[width=1\textwidth]{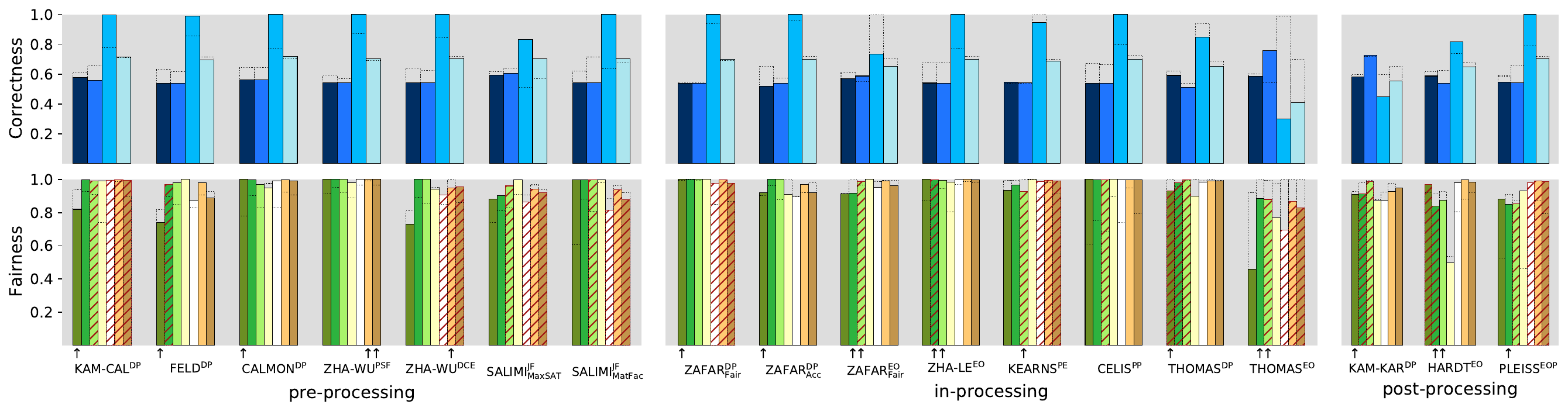}
  \vspace{-6mm}
  \caption{$T_3$}
    \vspace{5mm}
  \label{fig:experiment_t3}
  \end{subfigure}
  
  \vspace{-2mm} 
  
  \caption{The results of our robustness experiment on 3 erroneous dataset.
  Higher scores for correctness (fairness) metrics correspond to more correct
  (fair) outcomes. The bars highlighted in red denote the reverse direction of
  the remaining discrimination---favoring the unprivileged group more than the
  privileged group. The arrows ($\uparrow$) denote the fairness metric(s) each
  approach is optimized for. The bar plots for each approach on the error-free
  dataset are overlaid for aiding visual comparison. } \vspace{-2mm}
  \label{fig:experiment_robustness}
  
\end{figure*}

\begin{figure*}[t]
  \centering
  \hspace{0mm}
  \includegraphics[width=0.65\textwidth]{images/main_legend.pdf}
  \vspace{1mm}

  \hspace{10mm}
  \begin{subfigure}[t]{0.288\textwidth}
  \includegraphics[height=1\textwidth]{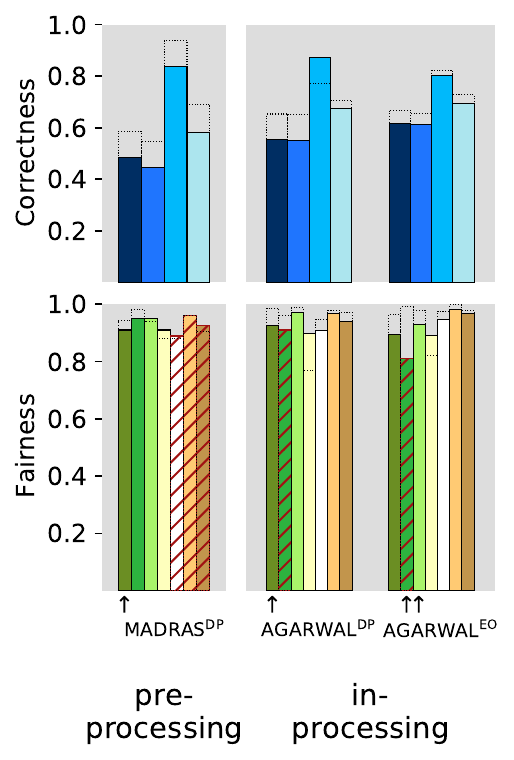}
  \caption{$T_1$}
  \end{subfigure}
  \hspace{-13mm}
  \begin{subfigure}[t]{0.288\textwidth}
  \includegraphics[height=1\textwidth]{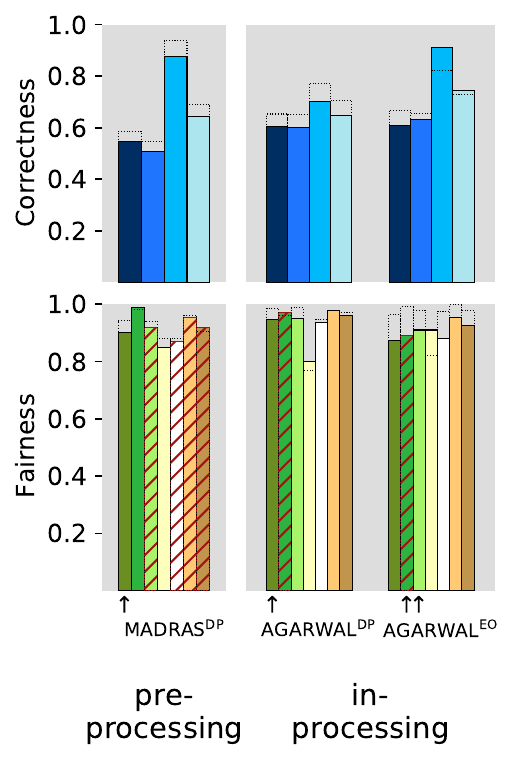}
  \caption{$T_2$}
  \end{subfigure}
  \hspace{-13mm}
  \begin{subfigure}[t]{0.288\textwidth}
  \includegraphics[height=1\textwidth]{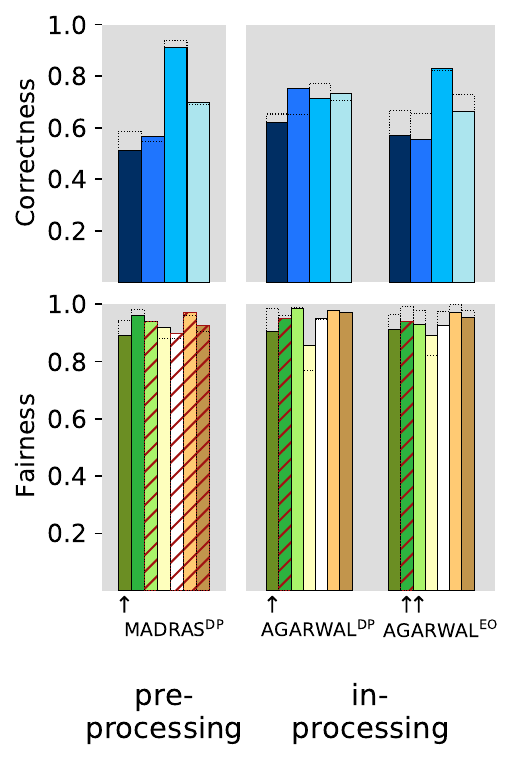}
  \caption{$T_3$}
  \end{subfigure}
  
  
  \caption{The results of robustness experiment on 3 erroneous dataset over
  3 additional fair classifiers. Higher scores for correctness (fairness)
  metrics correspond to more correct (fair) outcomes. The bars highlighted in
  red denote the reverse direction of the remaining discrimination---favoring
  the unprivileged group more than the privileged group. The arrows ($\uparrow$)
  denote the fairness metric(s) each approach is optimized for. The bar plots
  for each approach on the error-free dataset are overlaid for aiding visual
  comparison. } 
  \label{fig:other_robustness}
  
\end{figure*}

  \begin{figure*}[t]
    \centering
    \begin{subfigure}[t]{0.99\textwidth}
    \includegraphics[width=1\textwidth]{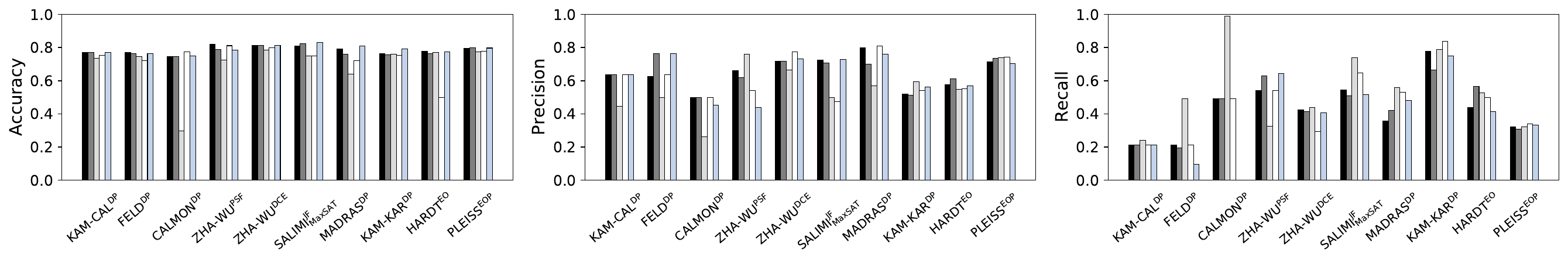}
    \vspace{-6mm}
    \end{subfigure}
  
    \begin{subfigure}[t]{0.99\textwidth}
    \includegraphics[width=1\textwidth]{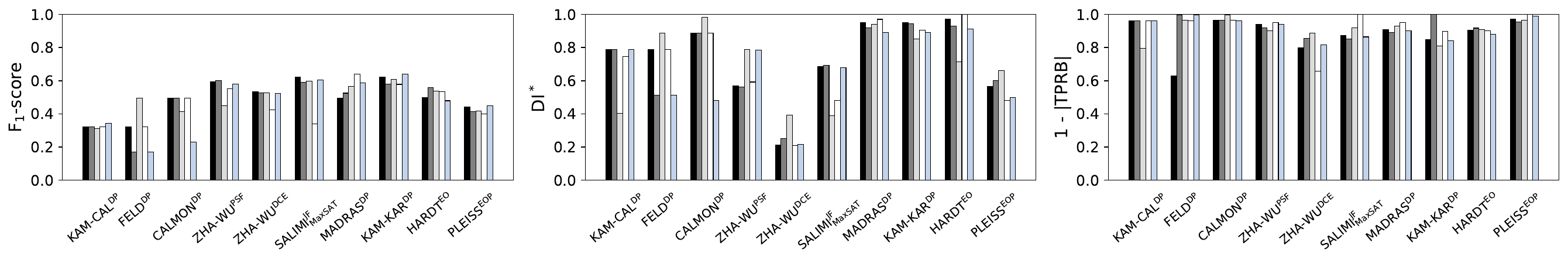}
    \vspace{-6mm}
    \end{subfigure}
    
    \begin{subfigure}[t]{0.99\textwidth}
    \includegraphics[width=1\textwidth]{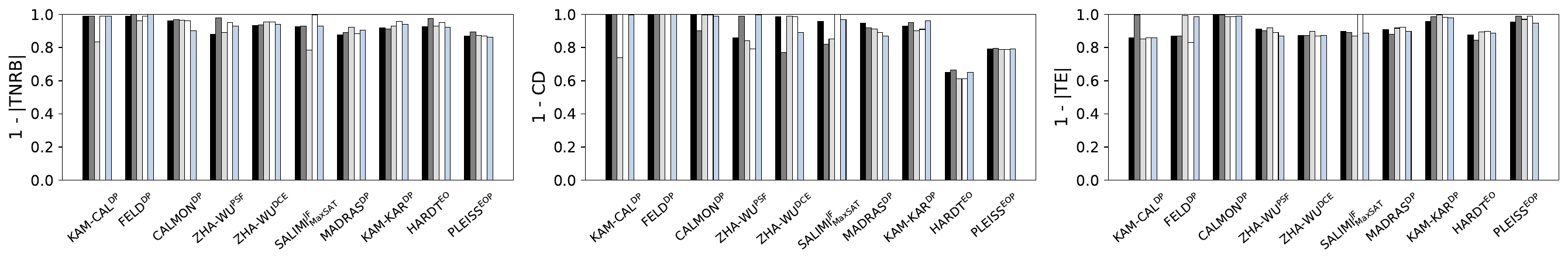}
    \vspace{-6mm}
    \end{subfigure}

    \begin{subfigure}[t]{0.66\textwidth}
        \includegraphics[width=1\textwidth]{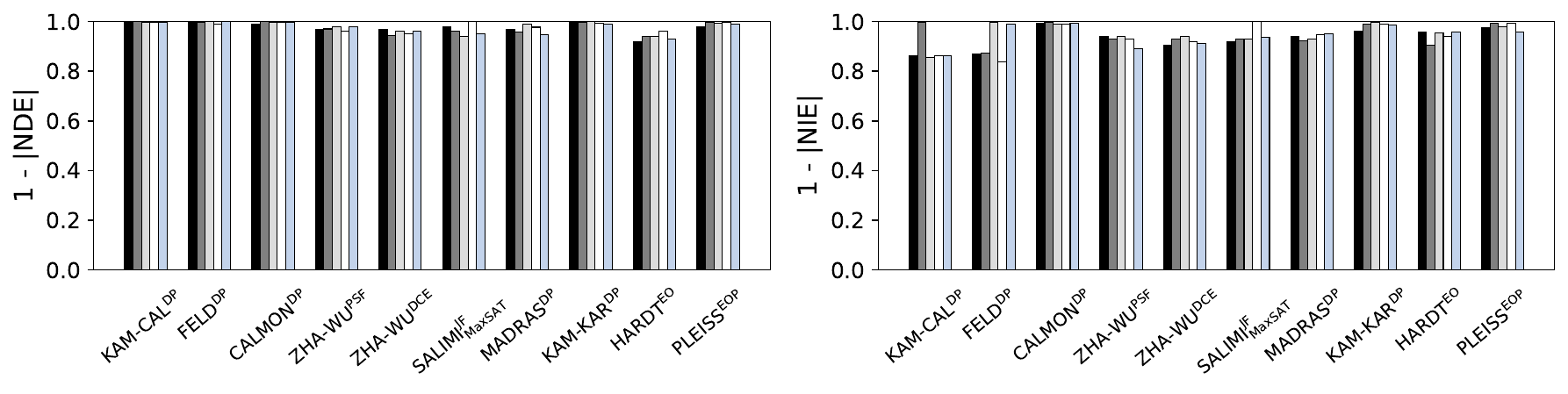}
        \vspace{-6mm}
    \end{subfigure}
    
    \vspace{-2mm} 
    
    \caption{The sensitivity of pre- and post-processing approaches (including
    the additional ones) to the choice of ML model in terms of all correctness
    and fairness metrics on the Adult dataset. }
    \vspace{-2mm}
    \label{fig:experiment_sensitivity}   
  \end{figure*}

  \begin{figure*}[t]
    \centering
    \begin{subfigure}[t]{0.99\textwidth}
    \includegraphics[width=1\textwidth]{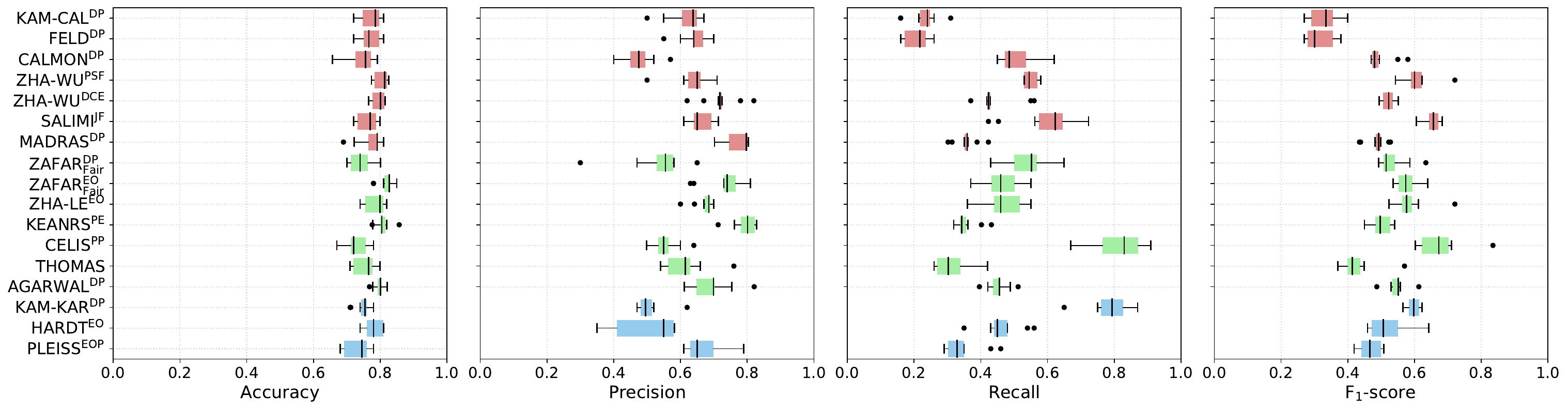}
	    	\vspace{5mm}
    \end{subfigure}

    \begin{subfigure}[t]{0.97\textwidth}
    \includegraphics[width=1\textwidth]{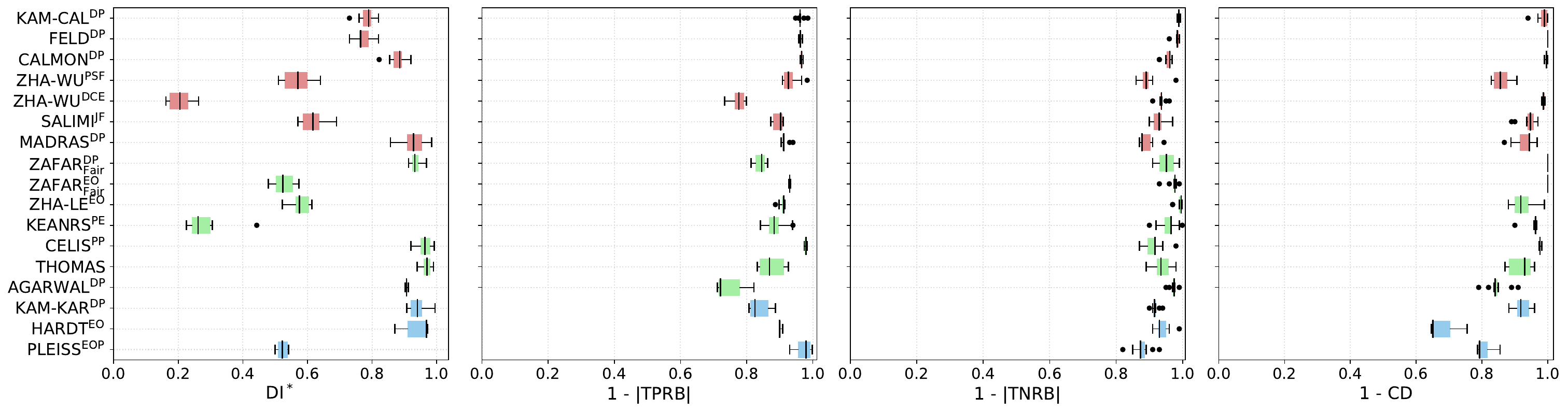}
    \vspace{5mm}
    \end{subfigure}

    \begin{subfigure}[t]{0.74\textwidth}
        \includegraphics[width=1\textwidth]{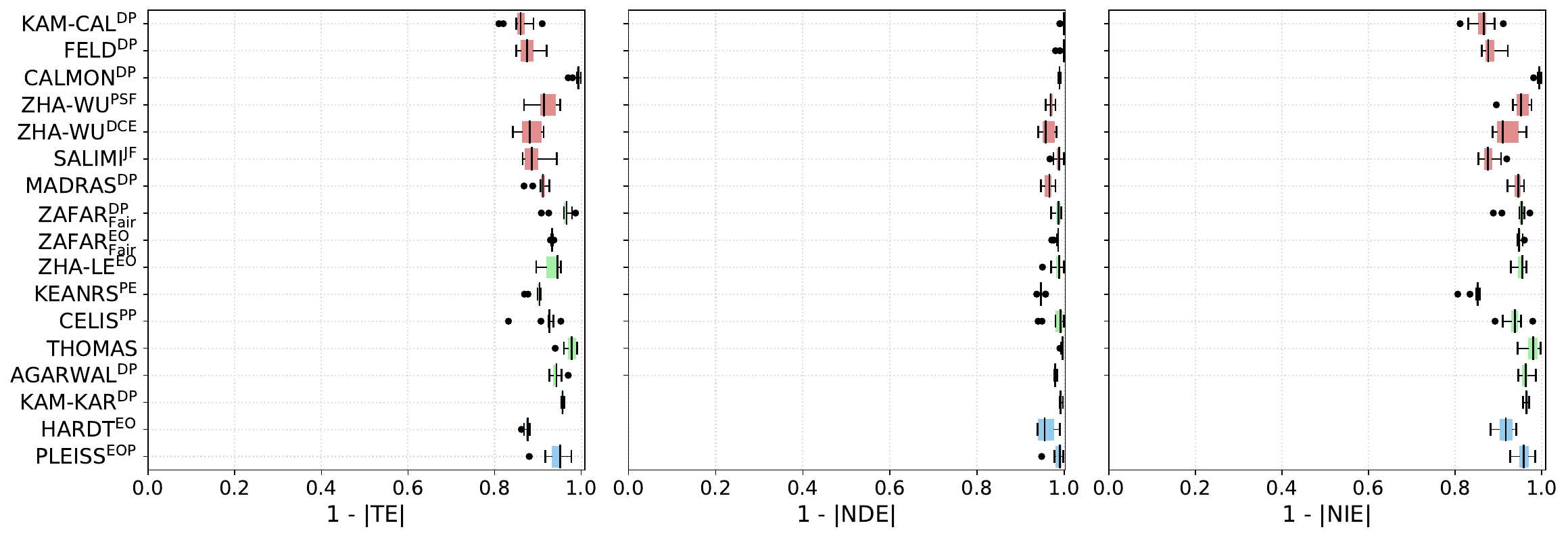}
        \vspace{-6mm}
    \end{subfigure}
    
    
    \caption{Variance of the fair approaches (including the additional ones) in
    terms of correctness and fairness metrics on arbitrary folds over Adult
    dataset.}
    \label{fig:experiment_var}
    
  \end{figure*}
  
\begin{figure*}[t]
    \centering
    \begin{subfigure}{\linewidth}
        \hspace{4.5mm}
        \includegraphics[width=0.95\linewidth]{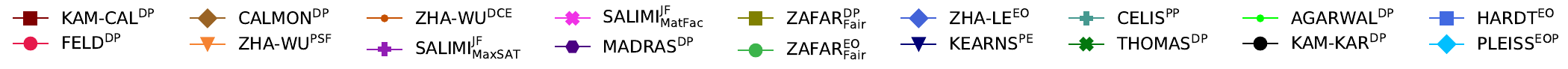}
		\vspace{1.7mm}
    \end{subfigure}

    \begin{subfigure}[t]{0.48\textwidth}
    \includegraphics[width=1\textwidth]{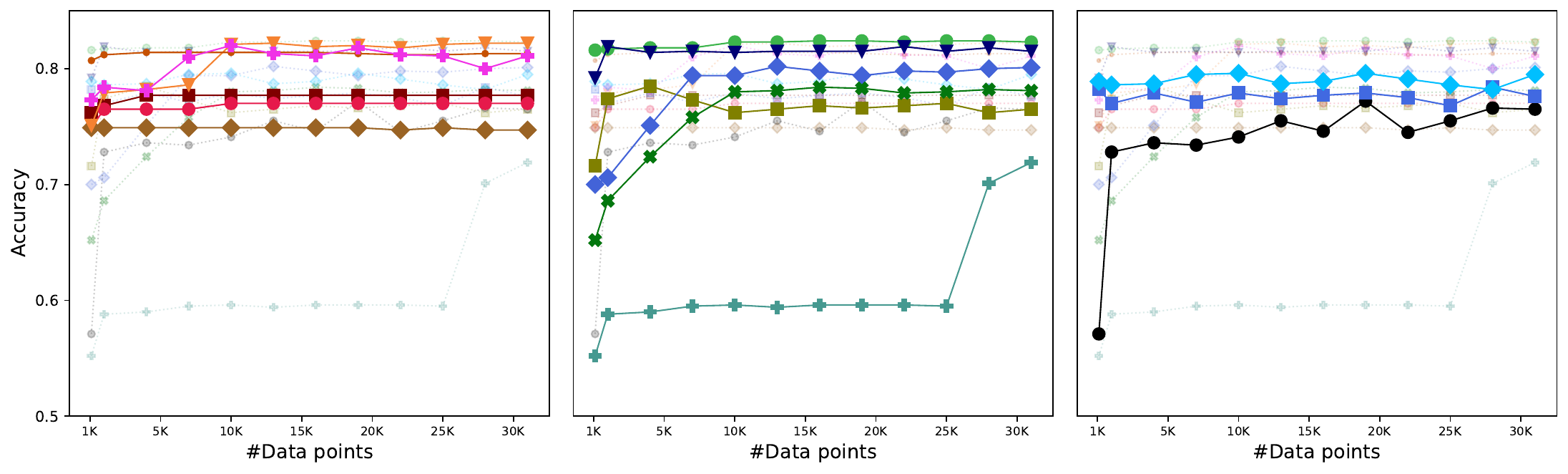}
    \vspace{-6mm}
    \caption{Accuracy}
    \end{subfigure}
    \begin{subfigure}[t]{0.48\textwidth}
    \includegraphics[width=1\textwidth]{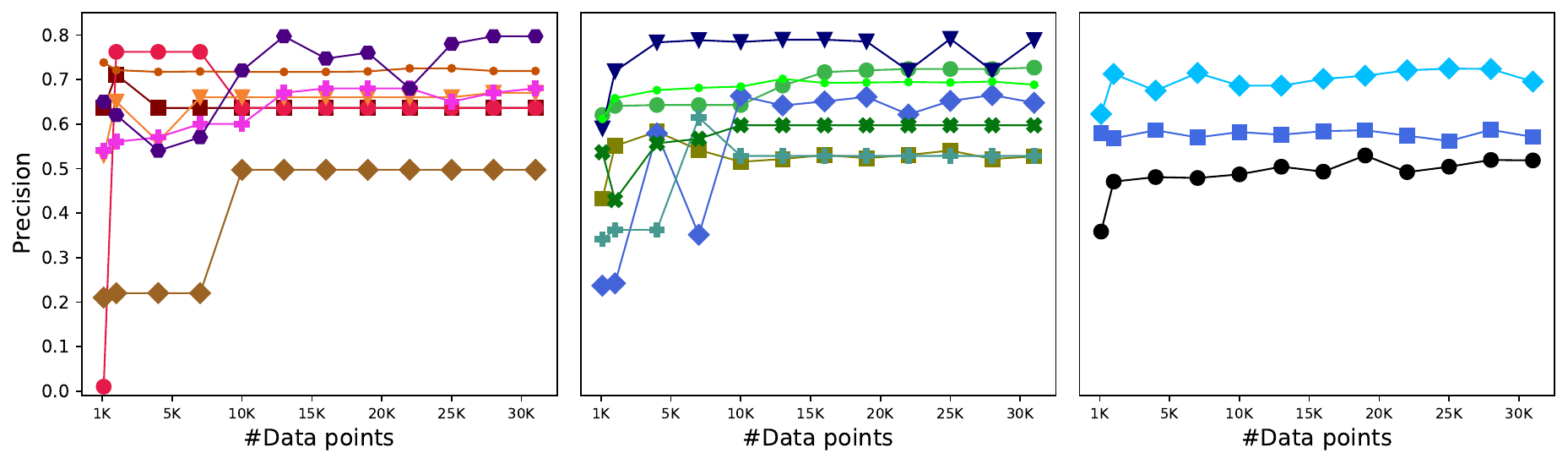}
    \vspace{-6mm}
    \caption{Precision}
    \end{subfigure}
    \vspace{5mm}
	
    \begin{subfigure}[t]{0.48\textwidth}
    \includegraphics[width=1\textwidth]{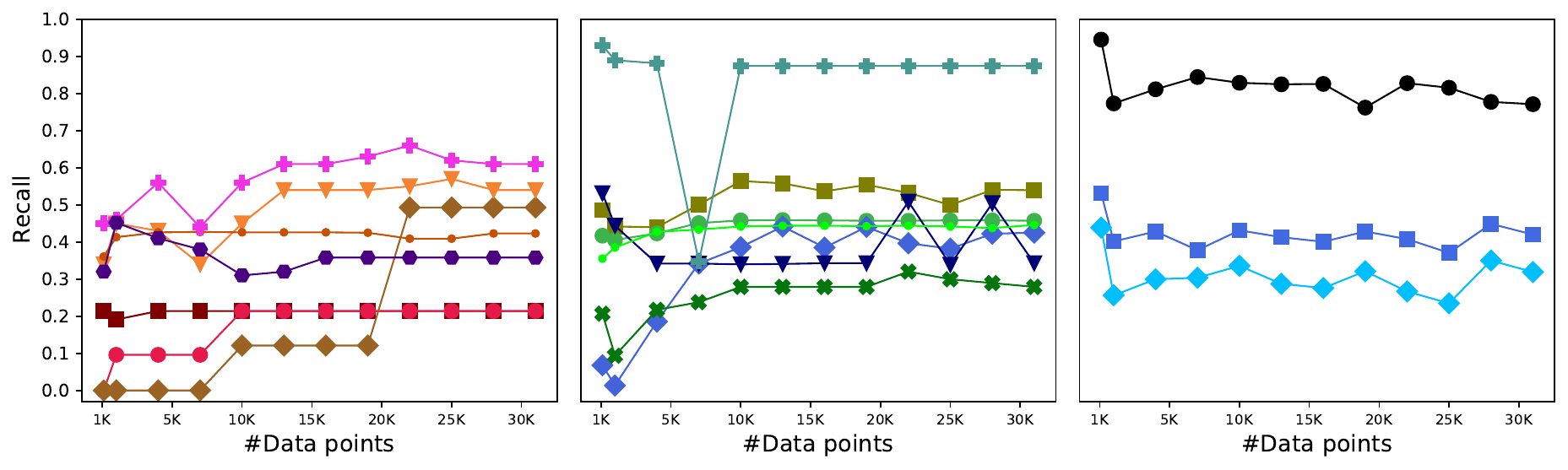}
    \vspace{-6mm}
    \caption{Recall}
    \end{subfigure}
    \begin{subfigure}[t]{0.48\textwidth}
      \includegraphics[width=1\textwidth]{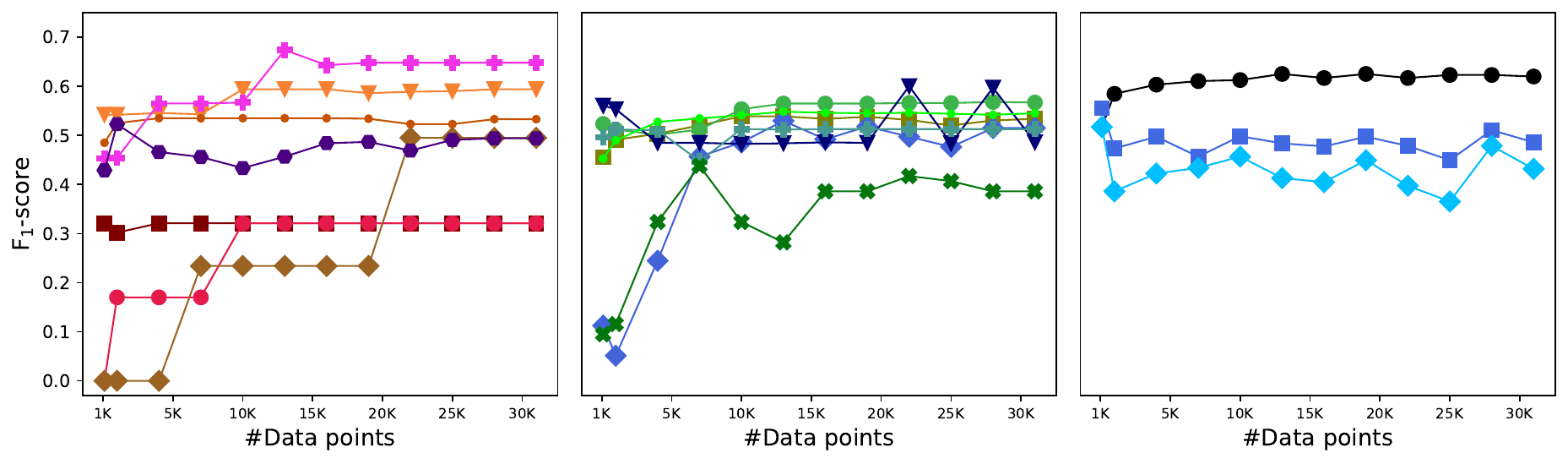}
      \vspace{-6mm}
      \caption{F$_1$-score}
    \end{subfigure}
  \vspace{5mm}
  
    \begin{subfigure}[t]{0.48\textwidth}
      \includegraphics[width=1\textwidth]{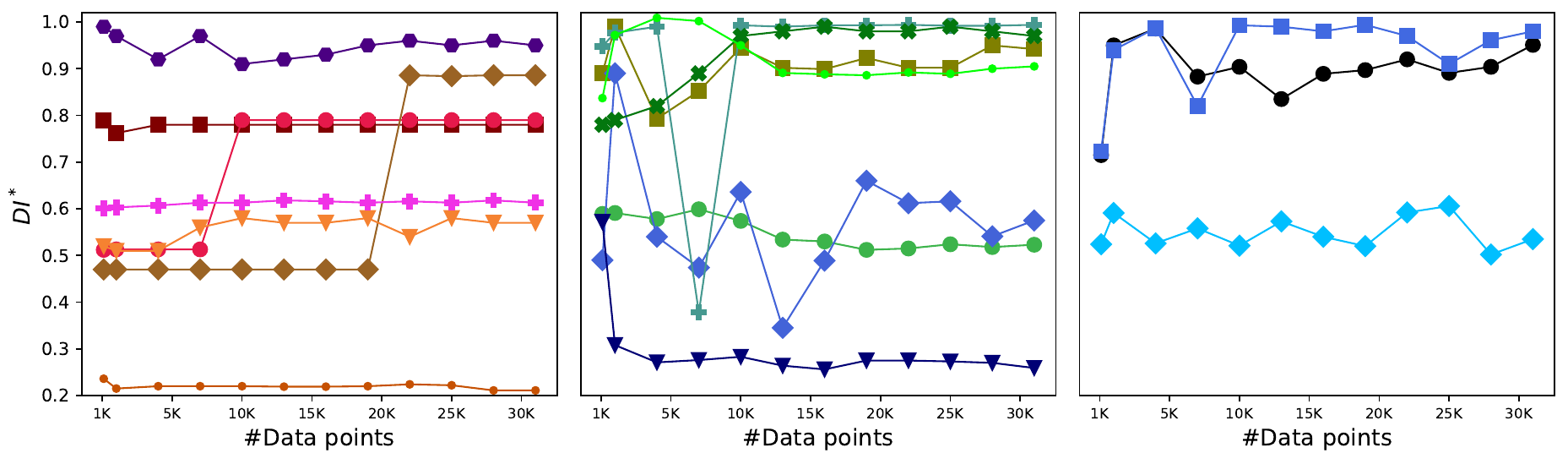}
      \vspace{-6mm}
      \caption{DI$^*$}
      \end{subfigure}
      \begin{subfigure}[t]{0.48\textwidth}
        \includegraphics[width=1\textwidth]{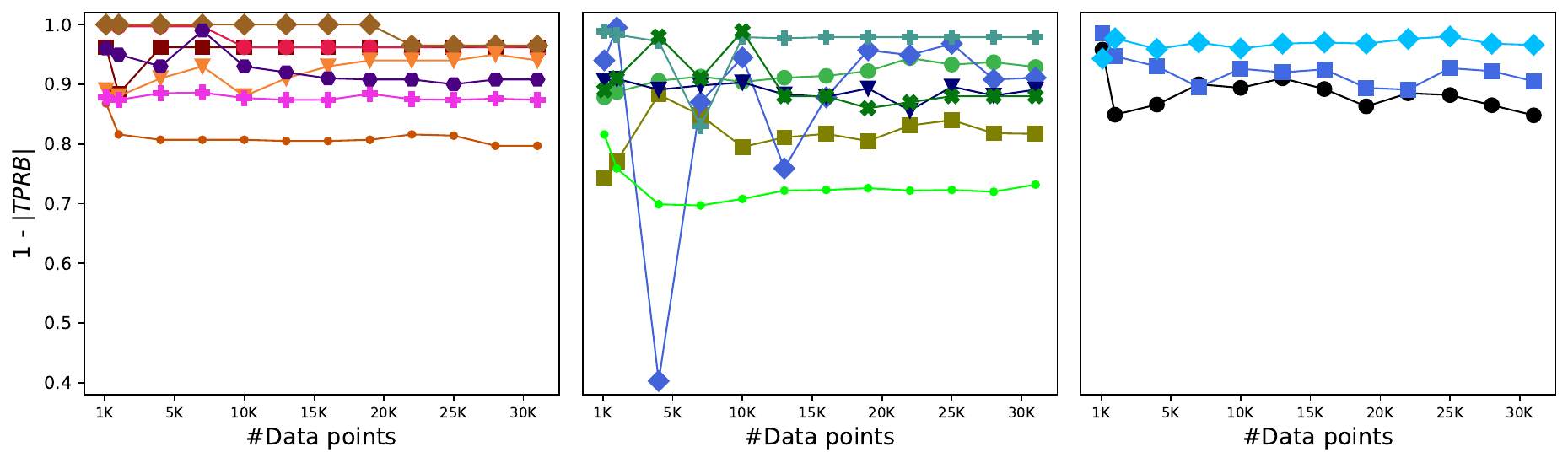}
        \vspace{-6mm}
        \caption{1 - |TPRB|}
    \end{subfigure}
      \vspace{5mm}
	  
    \begin{subfigure}[t]{0.48\textwidth}
      \includegraphics[width=1\textwidth]{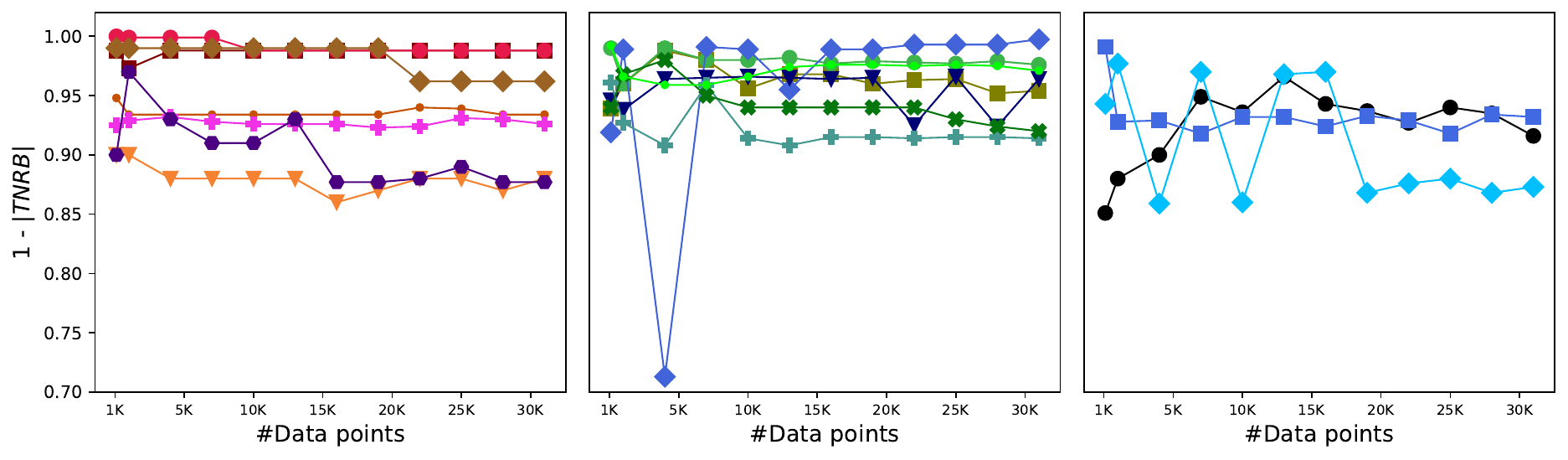}
      \vspace{-6mm}
      \caption{1 - |TNRB|}
      \end{subfigure}
      \begin{subfigure}[t]{0.48\textwidth}
        \includegraphics[width=1\textwidth]{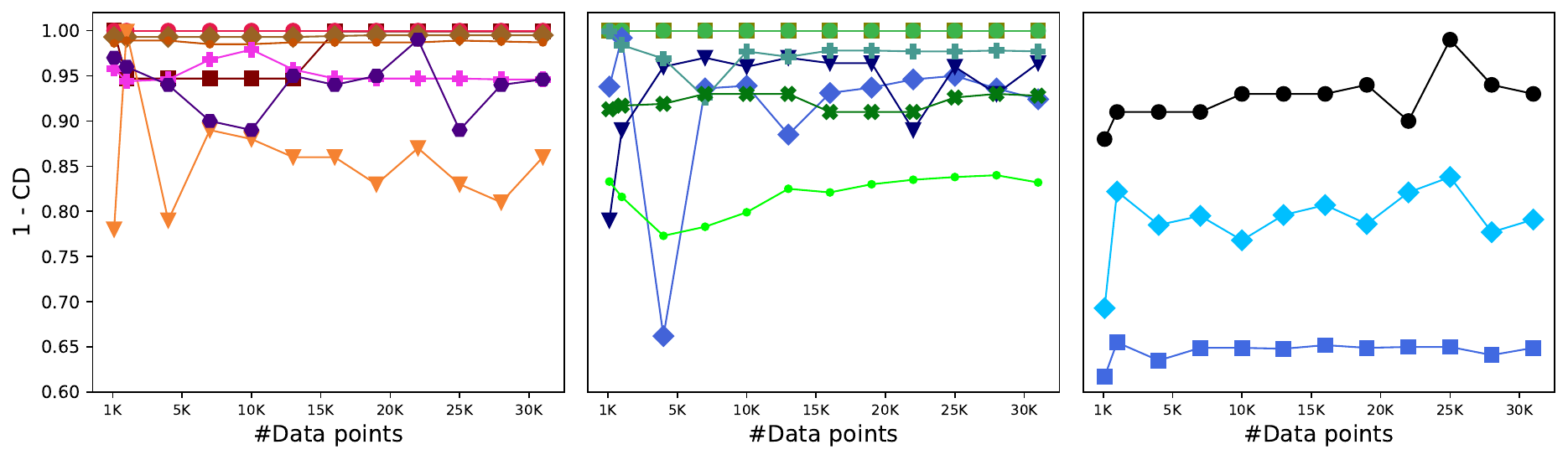}
        \vspace{-6mm}
        \caption{1 - CD}
    \end{subfigure}
	\vspace{5mm}
    
    \begin{subfigure}[t]{0.48\textwidth}
      \includegraphics[width=1\textwidth]{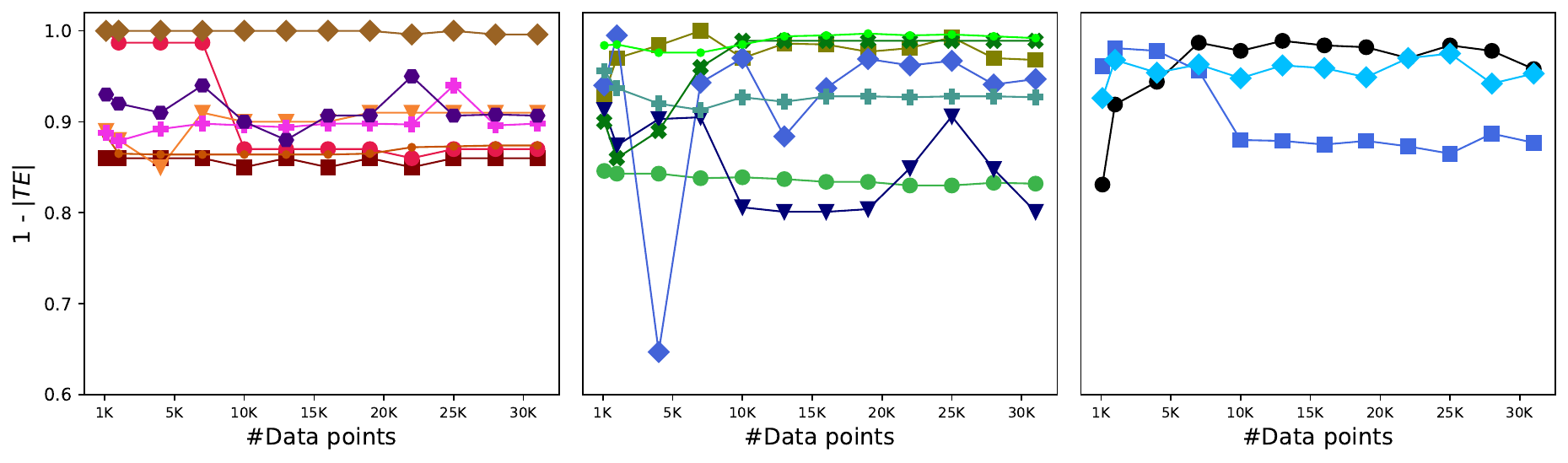}
      \vspace{-6mm}
      \caption{1 - |TE|}
      \end{subfigure}
      \begin{subfigure}[t]{0.48\textwidth}
        \includegraphics[width=1\textwidth]{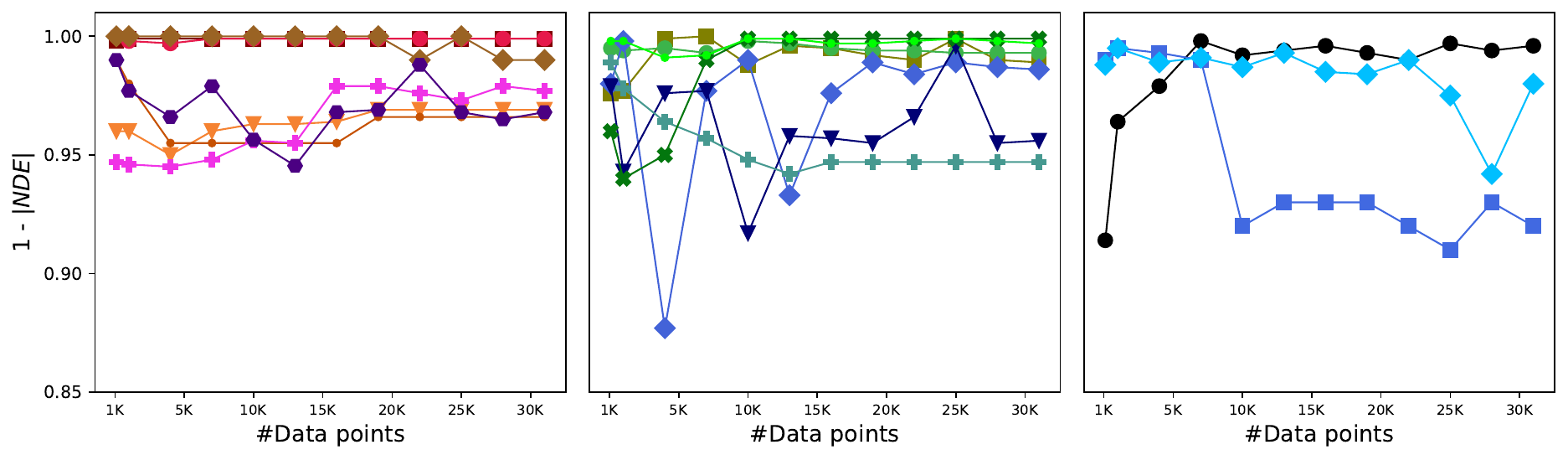}
        \vspace{-6mm}
        \caption{1 - |NDE|}
    \end{subfigure}
    \vspace{5mm}
	
    \begin{subfigure}[t]{0.48\textwidth}
        \includegraphics[width=1\textwidth]{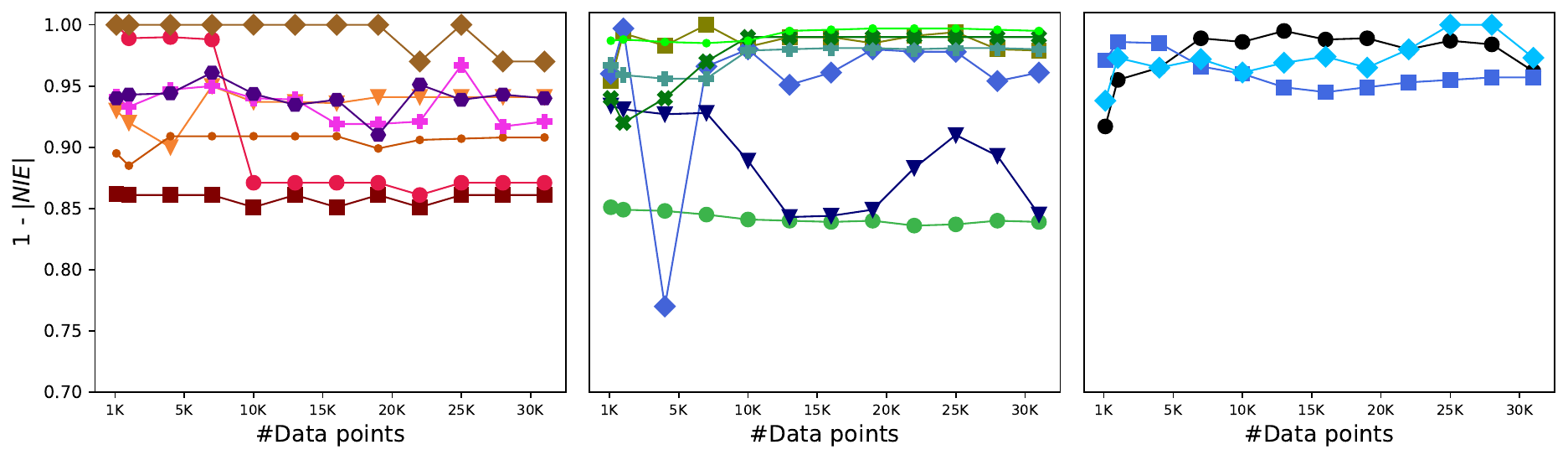}
        \vspace{-6mm}
        \caption{1 - |NIE|}
    \end{subfigure}

    \caption{The data efficiency of all fair approaches (including the additional
    ones) on Adult in terms of all correctness and fairness metrics. Note that
    the y-axis is scaled differently for each metric.}
    \vspace{-2mm}
    \label{fig:experiment_data_eff}
    
  \end{figure*}

  \begin{figure*}[t]
    \centering
  
    \vspace{6mm}
    \hspace{5mm}\includegraphics[width=0.8\textwidth, height=0.04\textwidth]{images/runtime_legend_1.pdf}\vspace{2mm}
    \begin{subfigure}[b]{0.188\textwidth}
      \includegraphics[width=1\textwidth]{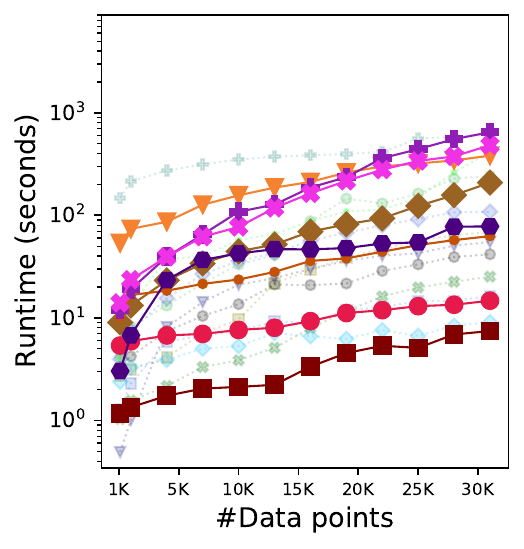}
    \vspace{-6mm}
      \caption{pre-processing}
    \end{subfigure} \hspace{-1em}
    \hspace{0.5mm}
    \begin{subfigure}[b]{0.16\textwidth}
      \includegraphics[width=1\textwidth]{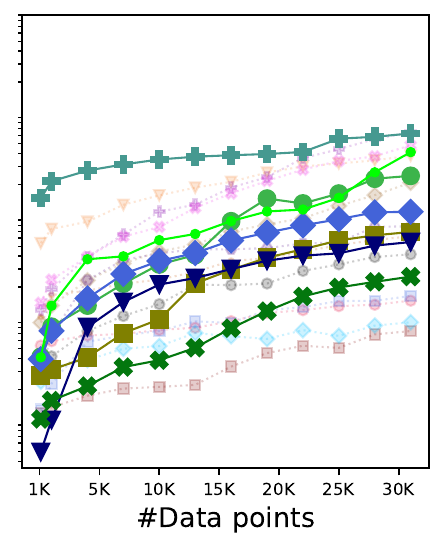}
    \vspace{-6mm}
      \caption{in-processing}
    \end{subfigure} \hspace{-1em}
    \hspace{0.5mm}
    \begin{subfigure}[b]{0.16\textwidth}
      \includegraphics[width=1\textwidth]{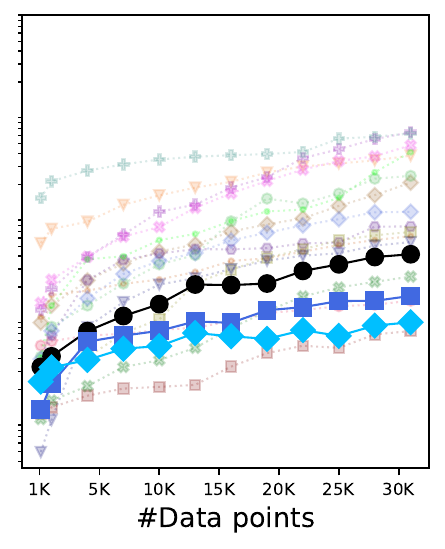}
    \vspace{-6mm}
      \caption{post-processing}
    \end{subfigure}
    \hspace{0.5mm}
    \begin{subfigure}[b]{0.16\textwidth}
      \includegraphics[width=1\textwidth]{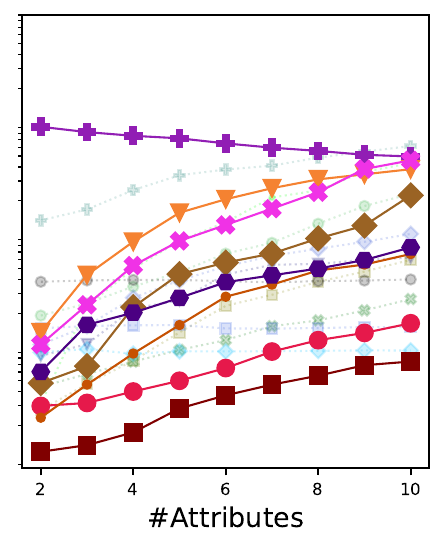}
      \vspace{-6mm}
    \caption{pre-processing}
    \end{subfigure} \hspace{-1em}
    \hspace{0.5mm}
    \begin{subfigure}[b]{0.16\textwidth}
      \includegraphics[width=1\textwidth]{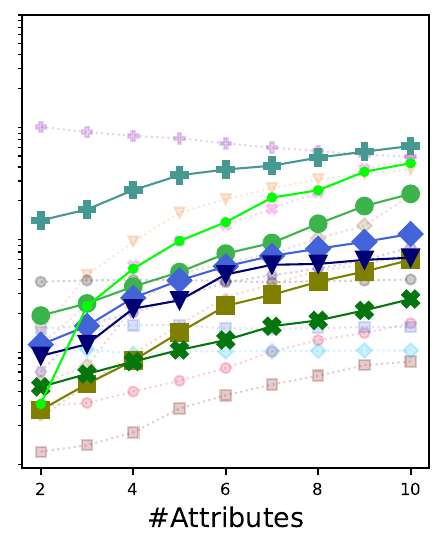}
    \vspace{-6mm}
    \caption{in-processing}
    \end{subfigure} \hspace{-1em}
    \hspace{0.5mm}
    \begin{subfigure}[b]{0.16\textwidth}
      \includegraphics[width=1\textwidth]{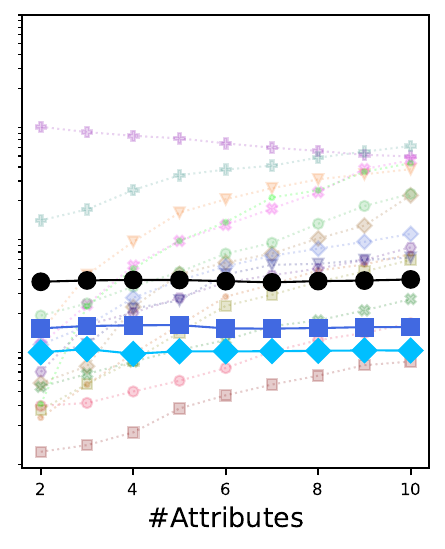}
    \vspace{-6mm}
      \caption{post-processing}
    \end{subfigure}
    \caption{\looseness-1 Complete results of runtime experiments on all
    fair approaches, including the additional ones. (a) -- (c) show runtime
    overhead with varying data size and (d) -- (f) show runtime overhead with
    varying number of attributes in Adult dataset. Note that the y-axis is in log
    scale.}
    \label{fig:runtime_oth_experiment} 
  \end{figure*}

\fi


\end{document}